\def\isarxiv{1} %%% for icml submission version, we comment this line

\ifdefined\isarxiv
\documentclass[11pt]{article}

\usepackage[numbers]{natbib}

\else
\documentclass[10pt,twocolumn,letterpaper]{article}

\usepackage[review]{cvpr}      % To produce the REVIEW version
\usepackage{xcolor}

\definecolor{cvprblue}{rgb}{0.21,0.49,0.74}
\usepackage[pagebackref,breaklinks,colorlinks,citecolor=cvprblue]{hyperref}

\fi

\usepackage{amsmath}
\usepackage{amsthm}
\usepackage{amssymb}
\usepackage{algorithm}
\usepackage{algpseudocode}
\usepackage{graphicx}
\usepackage{grffile}
\usepackage{wrapfig,epsfig}
\usepackage{url}
\usepackage{xcolor}
\usepackage{epstopdf}

\usepackage{bbm}
\usepackage{dsfont}

\allowdisplaybreaks

\ifdefined\isarxiv

\usepackage{tikz}
\usepackage{hyperref}  %%% arxiv don't allow this.
\hypersetup{colorlinks=true,citecolor=blue,linkcolor=blue} %%% Zhao : maybe we should comment this in submission.
\usetikzlibrary{arrows}
\usepackage[margin=1in]{geometry}

\else

\fi
\newtheorem{theorem}{Theorem}[section]
\newtheorem{lemma}[theorem]{Lemma}
\newtheorem{definition}[theorem]{Definition}

\newtheorem{fact}[theorem]{Fact}

\newcommand{\wt}{\widetilde}
\newcommand{\ov}{\overline}

\newcommand{\R}{\mathbb{R}}

\renewcommand{\d}{\mathrm{d}}

\DeclareMathOperator{\poly}{poly}

\DeclareMathOperator{\nnz}{nnz}

\DeclareMathOperator{\diag}{diag}

\DeclareMathOperator{\reg}{reg}
\DeclareMathOperator{\tot}{tot}

\makeatletter
\newcommand*{\RN}[1]{\expandafter\@slowromancap\romannumeral #1@}
\makeatother

\usepackage{lineno}

\ifdefined\isarxiv

\else

\title{Local Convergence of Approximate Newton Method for Two Layer Nonlinear Regression} 

\author{First Author\\
Institution1\\
Institution1 address\\
{\tt\small firstauthor@i1.org}
\and
Second Author\\
Institution2\\
First line of institution2 address\\
{\tt\small secondauthor@i2.org}
}

\fi

\begin{document}

\ifdefined\isarxiv

\date{}

\title{Local Convergence of Approximate Newton Method for Two Layer Nonlinear Regression} 
\author{
Zhihang Li\thanks{\texttt{lizhihangdll@gmail.com}. Huazhong Agricultural University.}
\and
Zhao Song\thanks{\texttt{zsong@adobe.com}. Adobe Research.}
\and
Zifan Wang\thanks{\texttt{Zifan.wang@stonybrook.edu}.  Stonybrook University.}
\and
Junze Yin\thanks{\texttt{junze@bu.edu}. Boston University.}
}

\else

\maketitle 

\fi

\ifdefined\isarxiv
\begin{titlepage}
  \maketitle
  \begin{abstract}
There have been significant advancements made by large language models (LLMs) in various aspects of our daily lives. LLMs serve as a transformative force in natural language processing, finding applications in text generation, translation, sentiment analysis, and question-answering. The accomplishments of LLMs have led to a substantial increase in research efforts in this domain. One specific two-layer regression problem has been well-studied in prior works, where the first layer is activated by a ReLU unit, and the second layer is activated by a softmax unit. While previous works provide a solid analysis of building a two-layer regression, there is still a gap in the analysis of constructing regression problems with more than two layers.

In this paper, we take a crucial step toward addressing this problem: we provide an analysis of a two-layer regression problem. In contrast to previous works, our first layer is activated by a softmax unit. This sets the stage for future analyses of creating more activation functions based on the softmax function. Rearranging the softmax function leads to significantly different analyses. Our main results involve analyzing the convergence properties of an approximate Newton method used to minimize the regularized training loss. We prove that the loss function for the Hessian matrix is positive definite and Lipschitz continuous under certain assumptions. This enables us to establish local convergence guarantees for the proposed training algorithm. Specifically, with an appropriate initialization and after $O(\log(1/\epsilon))$ iterations, our algorithm can find an $\epsilon$-approximate minimizer of the training loss with high probability. Each iteration requires approximately $O(\mathrm{nnz}(C) + d^\omega)$ time, where $d$ is the model size, $C$ is the input matrix, and $\omega < 2.374$ is the matrix multiplication exponent.

  \end{abstract}
  \thispagestyle{empty}
\end{titlepage}

{\hypersetup{linkcolor=black}
% \tableofcontents
}
\newpage

\else

\begin{abstract}
There have been significant advancements made by large language models (LLMs) in various aspects of our daily lives. LLMs serve as a transformative force in natural language processing, finding applications in text generation, translation, sentiment analysis, and question-answering. The accomplishments of LLMs have led to a substantial increase in research efforts in this domain. One specific two-layer regression problem has been well-studied in prior works, where the first layer is activated by a ReLU unit, and the second layer is activated by a softmax unit. While previous works provide a solid analysis of building a two-layer regression, there is still a gap in the analysis of constructing regression problems with more than two layers.

In this paper, we take a crucial step toward addressing this problem: we provide an analysis of a two-layer regression problem. In contrast to previous works, our first layer is activated by a softmax unit. This sets the stage for future analyses of creating more activation functions based on the softmax function. Rearranging the softmax function leads to significantly different analyses. Our main results involve analyzing the convergence properties of an approximate Newton method used to minimize the regularized training loss. We prove that the loss function for the Hessian matrix is positive definite and Lipschitz continuous under certain assumptions. This enables us to establish local convergence guarantees for the proposed training algorithm. Specifically, with an appropriate initialization and after $O(\log(1/\epsilon))$ iterations, our algorithm can find an $\epsilon$-approximate minimizer of the training loss with high probability. Each iteration requires approximately $O(\mathrm{nnz}(C) + d^\omega)$ time, where $d$ is the model size, $C$ is the input matrix, and $\omega < 2.374$ is the matrix multiplication exponent.

\end{abstract}

\fi

\section{Introduction}

The emergence of Large Language Models (LLMs), exemplified by various works such as \cite{o23, CND+22, cha22, vsp+17, bmr+20, rns+18, zrg+22, dclt18, rwc+19}, marks a significant breakthrough in artificial intelligence and natural language processing. As shown in \cite{dsx23}, these models, compared to other deep learning approaches, have vastly improved language comprehension and generation capabilities, a feat attributed to several pivotal factors:
\begin{itemize}
    \item Innovations in deep learning algorithms,
    \item Advances in computing power, and
    \item The abundance of textual data.
\end{itemize}

The evolution of LLMs can be traced back to initial attempts at language modeling using neural networks. Over time, continuous exploration of diverse architectures and methodologies has propelled LLMs into sophisticated models adept at handling intricate linguistic structures and generating coherent text. These models have demonstrated remarkable achievements across various applications, including question-answering systems \cite{o23, cha22}, sentiment analysis \cite{uas20}, text generation \cite{o23, cha22, zrg+22}, and machine translation \cite{hwl21}. Their impact has revolutionized interactions with language-based technologies, introducing new possibilities for AI-driven communication. As the scale of these language models continues to expand, recent research efforts have focused on optimizing the training efficiency of LLMs \cite{mgn+23, mwy+23, pmxa23}.

Attention computation plays a crucial role in constructing LLMs \cite{cha22, bmr+20, rwc+19, dclt18, rns+18, vsp+17}, allowing these models to assign weights to different elements in input sequences, aiding in capturing pertinent details for precise inference. Self-attention, a widely adopted technique in the transformer model, enables models to manage lengthy sequences efficiently, understand contextual nuances, and generate more cohesive outputs. Numerous studies have highlighted the benefits of self-attention in facilitating In-Context learning \cite{wzw23, zfb23, gtlv22, szks21}. The definition of softmax self-attention is presented as follows:

\begin{definition}[Attention optimization \cite{gswy23}]\label{def:attention}
Suppose that the matrices $A_1, A_2, A_3, B$ are in $\R^{n \times d}$ and the matrices $X, Y$ are in $\R^{d \times d}$.

Let $D(X) \in \R^{n \times n}$ be defined as
\begin{align*}
    D(X) := \diag( \exp(A_1 X A_2^\top ) {\bf 1}_n )
\end{align*}

Then, the attention optimization problem is defined as follows:
    \begin{align*}
        \min_{X,Y \in \R^{d \times d}}  \| D(X)^{-1} \exp(A_1 X A_2^\top) A_3 Y - B \|_F^2.
    \end{align*}
\end{definition}

Based on the Definition~\ref{def:attention}, \cite{dls23} has also proposed and studied the softmax regression problem, namely
\begin{definition}[Softmax regression problem \cite{dls23}]\label{def:softmax}
Suppose that the matrix $A$ is in $\R^{n \times d}$ and a vector $c$ is in $\R^n$.

Then, the softmax regression problem is defined as
    \begin{align*}
     \min_{x \in \R^d} \| \langle \exp(Ax) , {\bf 1}_n \rangle^{-1} \exp(Ax) - c \|_2^2 .
    \end{align*}
\end{definition}

\cite{dsx23} first study the two-layer regression problem, which incorporates the softmax regression (see Definition~\ref{def:softmax}) and ReLU function:
\begin{definition}[Two-Layer Regression studied in \cite{dsx23}]\label{def:dsx23}

Let $A_1$ be a matrix in $\R^{n \times d}$ and $A_2$ be a matrix in $\R^{m \times n}$. Let $b$ be a vector in $\R^m$ and $x, \phi(x)$ be two vectors in $\R^n$. Let $R$ be a real number.

Suppose $\phi(x)_i$ is defined as
    \begin{align*}
        \phi(x)_i := \mathrm{ReLU}(x_i) = \max\{0, x_i\}.
    \end{align*}

This regression problem is to minimize:
\begin{align*}
    \|\langle\exp(A_2 \phi(A_1 x)), \mathbf{1}_m \rangle^{-1} \exp(A_2 \phi(A_1 x)) - b\|^2_2
\end{align*}
under conditions $\min_{x \in \{ \|x\|_2 \leq R, x \in \mathbb{R}^d \}}$.

\end{definition}

Note that the inner layer of the regression problem studied by \cite{dsx23} is the ReLU function, and the outer layer is the softmax regression (see Definition~\ref{def:softmax}). 

In this paper, we give a formal analysis of a two-layer regression problem, which contains the softmax regression (defined as in Definition~\ref{def:softmax}) and a general Lipschitz continuous function $h(x) \in \R^m$ (which we formally defined in Definition~\ref{def:h}). Our regression problem is defined as follows:
\begin{definition}[Two-Layer regression problem]\label{def:our_regression}

Let $h : \R^m \rightarrow \R^m$ (Definition~\ref{def:h}).
Let $A_1 \in \R^{n \times d}$, $A_2 \in \R^{m \times n}$, $b \in \R^m$, and $x \in \R^d$.

Let $L : \R^d \to \R$ be
    \begin{align*}
        L(x) := \frac{1}{2} \cdot \| h(A_2  \langle \exp(A_1 x ), {\bf 1}_n \rangle^{-1} \cdot \exp(A_1  x )) - b \|_2^2.
    \end{align*}

Our purpose in the paper is to analyze
\begin{align*}
    \min_{x \in R^d} L(x).
\end{align*}
\end{definition}

Note that the inner layer of our regression problem is the softmax regression, while the outer layer is represented by the Lipschitz continuous function $h(x) \in \mathbb{R}^m$. Hence, future researchers can integrate the two regression problems (as defined in Definition~\ref{def:dsx23} and Definition~\ref{def:our_regression}) to construct a three-layer regression model: the first layer being ReLU, the second layer being softmax, and the third layer represented by $h$. Furthermore, our function $h$ can be any arbitrary Lipschitz continuous function, ensuring generality for the third layer.

\subsection{Our Motivations and Results}

The one-layer attention optimization problem (see Definition~\ref{def:attention}) has been well-studied: many different variations of it have been studied in \cite{zhdk23,as23,bsz23,dls23,gsx23_incontext,gsy23_coin,gsyz23_quantum,dms23,syz23}, and finally, a complete one layer attention regression is studied in \cite{gswy23}. Although this one-layer attention can provide a solid theoretical foundation for supporting the ability of LLMs in simple tasks, like text generation, translation, and question-answering, effectively adapting LLMs for more complex structured prediction problems like computer vision remains an open challenge. 

Multilayer regression, on the other hand, can model more complex nonlinear relationships that better fit real-world data. Each layer transforms the data differently, allowing for the learning of complex patterns. With multiple layers, the model can detect intricate patterns in the data that would remain undetectable with a single linear regression, aiding the model's ability to generalize. Moreover, the multiple layers serve as a regularization mechanism, preventing overfitting--a concern in single linear models due to their fewer parameters. This attribute lends multilayer regression powerful predictive capabilities, particularly in tackling difficult regression problems involving high-dimensional, complex data. The layered structure extracts informative features, resulting in accurate predictions. Therefore, multilayer regression finds extensive application across diverse domains such as computer vision \cite{ksh12,dag+15,sz14} and speech recognition \cite{cjlv16,aaa+16,ssb14,gmh13}, showcasing its versatility and effectiveness in solving modern machine learning problems.

The informal version of our analysis is presented as follows:

\begin{theorem}[Informal version of Theorem~\ref{thm:theorem_main_formal}]\label{thm:theorem_main_informal}
Let $L(x)$ be defined in Definition~\ref{def:our_regression} and $x^*$ be the solution to the problem $\min_{x \in R^d} L(x)$. 

Suppose $\|A\| \leq R$ and $\|x\|_2 \leq R$.

For any accuracy parameter $\epsilon \in (0, 0.1)$ and failure probability $\delta \in (0, 0.1)$ and a initial point $x_0$ where $\|x_0\|_2 \leq R$, there exist an randomized algorithm (Algorithm~\ref{alg:main}) that runs $T = O(\log(\|x_0 - x^*\|_2 / \epsilon))$ iterations, spend $O((\nnz(C) + d^\omega) \poly(\log(m / \delta)))$ per iteration, and output $\wt{x}$ such that 
\begin{align*}
    \Pr[\|\wt{x} - x^*\|_2 \leq \epsilon] \ge 1 - \delta.
\end{align*}
\end{theorem}

To establish these results, we derive several key properties of the loss landscape, including the positive definiteness of the Hessian and its Lipschitz continuity. Our proofs draw tools from matrix analysis and exploit the structural properties of the softmax activation. Then, we provide theoretical justification for using Newton's method to minimize the regularized training loss. This allows us to ensure that the suggested training method has assurances of converging locally. The analysis techniques developed here may extend to studying other activation functions and deeper network architectures.

\paragraph{Roadmap.}

 In Section~\ref{sec:related_work}, we delve into related works. In Section~\ref{sec:prelim}, we introduce the notations and basic mathematical principles that underpin the subsequent developments in this paper. In Section~\ref{sec:gradient_hessian}, we present crucial gradient and Hessian matrices pertinent to our study. In Section~\ref{sec:hessian_prop}, we focus on analyzing the properties of the Hessian matrix, including its positive semi-definiteness (PSD) and Lipschitz continuity. In Section~\ref{sec:newton}, we introduce Newton's method. In Section~\ref{sec:main}, we combine the Hessian properties and Newton's methods to formulate the central Theorem (refer to Theorem~\ref{thm:theorem_main_formal}) and Algorithm (refer to Algorithm~\ref{alg:main}) of this paper. Finally, in Section~\ref{sec:conclusion}, we wrap up by summarizing our findings and highlighting the significance of our work.

 %%% Section 1. Introduction

\section{Related Work}
\label{sec:related_work}

\paragraph{Transformer theory.}

Since the emergence of LLMs, extensive research has been dedicated to analyzing their abilities, which build upon transformer theory. 
One crucial area of exploration revolves around understanding how transformers achieve in-context learning. 
\cite{szks21} proposed the comprehension of how a single-head attention module learns in machine translation. They defined ``Knowing to Translate Individual Words'' (KTIW), signifying the model's capacity to understand a word's translation. They argued that KTIW predominantly influences attention learning, as it can be acquired from word co-occurrence prior to attention learning. \cite{gtlv22} demonstrated that transformers can effectively learn linear function classes in close proximity to optimal levels. When provided with specific functions, transformers can extrapolate information about related functions. Additionally, \cite{asa+22} linked in-context learning to conventional algorithms by encoding smaller models in activations and updating them with new examples, showcasing this through linear regression.

\cite{zfb23} delved into examining in-context learning in regression, discovering that transformers can identify global optima but encounter difficulties under covariate shifts. Furthermore, \cite{wzw23} interpreted in-context learning as Bayesian selection, enabling the deduction of task-relevant information.

Moreover, research aims to comprehend the intrinsic structure of transformers. \cite{zpga23} discovered evidence of parsing in linguistic tasks, drawing parallels to algorithms like Inside-Outside. Additionally, \cite{psza23} pinpointed the location of skills post fine-tuning, demonstrating that only a few parameters are pivotal.

Other investigations have delved into the capabilities of transformers. \cite{sht23} found logarithmic scaling concerning input size in sparse problems, surpassing other networks, but identified linear scaling in detection. Additionally, \cite{ag23} linked scaling laws to inductive biases that facilitate efficient pre-training. Lastly, \cite{bce+23} evaluated GPT-4 across tasks, shedding light on future work extending beyond next-word prediction.

\paragraph{Attention.}

The work by \cite{bcb14} stands as one of the earliest instances of utilizing attention in NLP. They posited that incorporating an attention mechanism with a fixed-length vector could enhance encoder-decoder performance, enabling the decoder to focus on pertinent words during translation. This innovation notably elevated translation quality compared to models devoid of attention. Subsequently, \cite{lpm15} elucidated two attention types: local attention, which focuses on a subset of words, and global attention, which considers all words.

Attention has found broad applications, such as in image captioning \cite{xbk+15}, aligning image components with caption words. Within Transformers \cite{vsp+17}, attention captures disparities between words within sentences. In graph neural networks \cite{vcc+17}, attention computes relationships between nodes and their neighbors.

Following the emergence of LLMs, numerous studies have delved into attention computation \cite{dms23,as23,zhdk23,clp+21,lsz23,bsz23,kkl20}. Noteworthy among these are \cite{zhdk23,clp+21,kkl20}, which employ locality-sensitive hashing to approximate attention, exemplified by KDEformer \cite{zhdk23}, providing spectral norm bounds. Research explores both static and dynamic attention \cite{bsz23,as23}, and investigates hyperbolic regression problems \cite{lsz23}. \cite{dms23} proposes algorithms to reduce attention matrix dimensionality in LLMs.

Other works have analyzed attention optimization and convergence \cite{llr23,gms23,szks21,zkv+20}. \cite{llr23} delved into acquiring word co-occurrence knowledge. \cite{gms23} focused on regression with exponential activations. \cite{szks21} analyzed the prioritization of significant words and attention evolution. \cite{zkv+20} demonstrated that heavy-tailed noise leads to issues in Stochastic Gradient Descent (SGD) compared to adaptive methods.

\paragraph{Convergence and Deep Neural Network Optimization}

Numerous works have concentrated on analyzing optimization, convergence guarantees, and training enhancements for neural networks.

\cite{ll18} showcased how stochastic gradient descent optimizes over-parameterized networks on structured data, while \cite{dzps18} demonstrated gradient descent's effectiveness on such networks. \cite{azls19a} developed a convergence theory for over-parameterized deep networks using gradient descent. \cite{azls19b} scrutinized convergence rates for recurrent neural networks. \cite{adh+19a} provided an analysis of optimization and generalization for over-parameterized two-layer networks. \cite{adh+19b} delved into computation with infinitely wide networks. \cite{cgh+19} introduced the Gram-Gauss-Newton method for optimizing over-parameterized networks. \cite{zg19} enhanced the analysis of stochastic gradient descent convergence for deep networks, necessitating milder over-parameterization.

Other studies, such as those by \cite{os20, jt19, zpd+20}, center on optimization and generalization, while \cite{gms23, lsz23} emphasize convergence rates and stability. Works by \cite{bpsw20, szz21, als+22, mosw22, z22} focus on developing specialized optimization algorithms and techniques, and \cite{lss+20, hlsy21} leverage network structure.

\section{Preliminary}
\label{sec:prelim}

In Section~\ref{sub:preli:def}, we present the definitions of the important functions in this paper, where these functions are used to decompose the loss function (defined in Definition~\ref{def:our_regression}) into simpler forms. In Section~\ref{sub:preli:reg}, we express our loss function by using $L_{\reg}$ and $L_{\tot}$.

\paragraph{Notations.}

Now, we introduce the key notations.

Let $x \in \R^{d}_{>0}$ denote a length-$n$ where all the entries are positive.
For a matrix $A \in \R^{n \times d}$, $\exp(A)$ denotes the $n \times d$ matrix given by $\exp(A) = \sum_{i=0}^{\infty} \frac{1}{i!}A^i$.
We use ${\bf 1}_n$ to denote a length-$n$ vector where all its entries are $1$.
For an arbitrary vector $x$, we denote its $\ell_p$ norm by $\|x\|_p$, where $\|x\|_1 := \sum_{i=1}^n |x_i|, \|x\|_2 := (\sum_{i=1}^n x_i^2)^{1/2}$, and $\|x\|_\infty := \max_{i \in [n]} |x_i|$.
For a matrix $A \in \R^{n \times k}$ where $n > k$, we denote $\|A\|$ to be the spectral norm of $A$ and $\|A\|:= \sup_{x \in \R^k} \|Ax\|_2 / \|x\|_2$
For two vectors $a, b \in \R^n$, we define $\langle a, b \rangle := \sum_{i=1}^n a_ib_i$.
For two vectors $a, b \in \R^n$, we use $a \circ b$ to denote the vector where its $i$-th entry is $a_i b_i$ for $i \in [n]$.
For $x \in \R^n$, we define the diagonal matrix $\diag(x) \in \R^{n \times n}$, whose diagonal entries are given by $\diag(x)_{i,i} = x_i$ for $i \in [n]$, and the entries elsewhere are all zero.
For a symmetric matrix $A \in \R^{n \times n}$ with real entries, we denote $A \succ 0$ to indicate the matrix is positive-definite (PD), if $x^\top A x > 0$ for all $x \in \R^n$.
For a symmetric matrix $A \in \R^{n \times n}$ with real entries, we denote $A \succeq 0$ to indicate the matrix is positive-semidefinite (PSD), if $x^\top A x \geq 0$ for all $x \in \R^n$.

\subsection{Definition}
\label{sub:preli:def}

In this section, we present the definition of the functions.

\begin{definition} \label{def:u}
 
Given $A_1 \in \R^{n \times d}$, let $u(x) : \R^d \rightarrow \R_{>0}^n$ 
be
\begin{align*}
    u(x) := \exp(A_1 \cdot x )
\end{align*}

\end{definition}

\begin{definition} \label{def:alpha}

Let $\alpha(x): \R^d \rightarrow \R_{>0}$ be
\begin{align*}
    \alpha(x) := \langle u(x), {\bf 1}_n \rangle .
\end{align*}
\end{definition}

\begin{definition} \label{def:f}

Let $f(x): \R^d \rightarrow \R_{>0}^n$ be
\begin{align*}
    f(x) := \alpha(x)^{-1} \cdot u(x).
\end{align*}
\end{definition}

\begin{fact}\label{fac:f_l1_norm}
By definition of $f(x)$, we can see that $\| f(x) \|_1 = 1$.
\end{fact}
\begin{proof}

By the definition of $f(x)$, we have
\begin{align*}
    \| f(x) \|_1
    = & ~ \|\alpha(x)^{-1} \cdot u(x)\|_1\\
    = & ~ \|\langle \exp(A_1 \cdot x ), {\bf 1}_n \rangle^{-1} \cdot \exp(A_1 \cdot x )\|_1\\
    = & ~ \|(\exp(A_1 \cdot x ) \cdot {\bf 1}_n^\top)^{-1} \cdot \exp(A_1 \cdot x )\|_1\\
    = & ~ \|\|\exp(A_1 \cdot x )\|_1^{-1} \cdot \exp(A_1 \cdot x )\|_1\\
    = & ~ \|\exp(A_1 \cdot x )\|_1^{-1} \cdot \|\exp(A_1 \cdot x )\|_1\\
    = & ~ 1,
\end{align*}
where the second step follows from the definition of $\alpha(x)$ and $u(x)$ (see Definition~\ref{def:alpha} and Definition~\ref{def:u}), the third step follows from Fact~\ref{fac:basic_algebra}, the fourth step follows from the definition of $\|\cdot\|_1$, the fifth step follows from simple algebra, and the last step follows from simple algebra.
\end{proof}

\begin{definition}\label{def:h}

Let $h: \R^m \to \R^m$ be a Lipschitz continuous function, namely, there exists a real number $L_h \geq 0$ such that for all $x, y \in \R^m$,
\begin{align*}
    \|h(x) - h(y)\|_2 \leq L_h \cdot \|x - y\|_2.
\end{align*}

Moreover, we assume $h'$ is also a Lipschitz continuous function, namely, there exists a real number $L_h \geq 0$ such that for all $x, y \in \R^m$,
\begin{align*}
    \|h'(x) - h'(y)\|_2 \leq L_h \cdot \|x - y\|_2.
\end{align*}

\end{definition}

\begin{definition} \label{def:c}

Let $h : \R^m \rightarrow \R^m$ (Definition~\ref{def:h}), $A_2 \in \R^{m \times n}$, and $b \in \R^m$.

Let $c(x): \R^d \rightarrow \R^m$ be 
\begin{align*}
    c(x): = h(A_2 f(x)) - b
\end{align*}

\end{definition}

\begin{definition} \label{def:L}

Let $L : \R^d \rightarrow \R_{>0}$ be 
\begin{align*}
   L(x) := \frac{1}{2} \cdot \| c(x) \|_2^2.
\end{align*}
\end{definition}

\subsection{Regularization}\label{sub:preli:reg}

In this section, we introduce $L_{\reg}$ and $L_{\tot}$.

\begin{definition}\label{def:L_reg}
Let $A_1 \in \R^{n \times d}$.

For a given vector $w \in \R^n$, we let $W = \diag(w) \in \R^{n \times n}$. 

We define $L_{\reg} : \R^d \rightarrow \R$ as follows
\begin{align*}
L_{\reg}(x):= 0.5 \| W A_1 x\|_2^2
\end{align*}
\end{definition}

\begin{lemma}[Folklore, see \cite{lsz23} as an example]\label{lem:L_reg_gradient_hessian}
For a given vector $w \in \R^n$, we let $W = \diag(w) \in \R^{n \times n}$. Let $L_{\reg} : \R^d \rightarrow \R$ be defined as Definition~\ref{def:L_reg}.

Then, we have
\begin{itemize}
\item The gradient is
\begin{align*}
\frac{\d L_{\reg}}{ \d x} = A_1^\top W^2 A_1 x
\end{align*}
\item The Hessian is
\begin{align*}
\frac{\d^2 L_{\reg}}{ \d x^2} = A_1^\top W^2 A_1
\end{align*}
\end{itemize}
\end{lemma}

\begin{definition}\label{def:L_tot}
    If the following conditions hold
    \begin{itemize}
        \item Let $L(x)$ be defined in Definition~\ref{def:L}
        \item Let $L_{\reg}(x)$ be defined in Definition~\ref{def:L_reg}
    \end{itemize}
    Then we define our loss function as follows:
    \begin{align*}
        L_{\tot}(x) := L(x) + L_{\reg}(x)
    \end{align*}
\end{definition}

\section{Gradient and Hessian}
\label{sec:gradient_hessian}

For Section~\ref{sub:gradient_hessian:gra}, we present the important gradient of the basic functions. For Section~\ref{sub:gradient_hessian:he}, we present the important Hessian of the basic functions. For Section~\ref{sub:gradient_hessian:re}, we re-organize the important parts of the Hessian to simplify our expression and make further analysis more convenient. 

\subsection{Gradient}
\label{sub:gradient_hessian:gra}

Now, we present the gradient, where the full details can be seen in Section~\ref{sec:app_gradient}.

\begin{lemma}[Informal version of Lemma~\ref{lem:gradient}]
\label{lem:gradient_informal}

We consider that $u(x)$, $\alpha(x)$, $f(x)$, $h$, $c(x)$, $L(x)$ are established in Definition~\ref{def:u}, \ref{def:alpha}, \ref{def:f}, \ref{def:h}, \ref{def:c}, and \ref{def:L}.
 
If the following holds
\begin{itemize}
    \item $A_{1,*,i} \in \R^n$ represents the $i$-th column vector of $A_1 \in \R^{n \times d}$ for all $i \in [d]$
    \item Let $A_{1,l,*} \in \R^d$ denote the $l$-th row vector of $A_1 \in \R^{n \times d}$ for all $l \in [n]$
    \item Let $A_{2,k,*} \in \R^n$ denote the $k$-th row vector $A_2 \in \R^{m \times n}$ for each $k \in [m]$
\end{itemize}
then for each $i \in [d]$, we have
\begin{itemize}
    \item Part 1. Let $p(x)_i \in \R^n$ be defined as $p(x)_i := f(x) \circ A_{1,*,i} -\langle f(x), A_{1,*,i} \rangle \cdot f(x)$ 
    \begin{align*}
        \frac{\d f(x)}{\d x_i} = p(x)_i
    \end{align*}
    \item Part 2. Let $h'(A_2 f(x)) \in \R^m$ denote a length-$m$ vector $i$-th coordinate is the $\frac{\d h(y_i)}{\d y_i} \big|_{y_i = (A_2 f(x))_i}$ (here we can should think of $h : \R \rightarrow \R$), we have
    \begin{align*}
        \underbrace{ \frac{\d h(A_2 f(x))}{\d x_i} }_{m \times 1} = \underbrace{ \diag( h'(A_2 f(x)) ) }_{m \times m} \cdot \underbrace{ A_2 }_{m \times n} \cdot \underbrace{ p(x)_i }_{n \times 1}
    \end{align*} 
    \item Part 3.  
    \begin{align*}
        \frac{\d L(x)}{\d x_i} = \langle \underbrace{c(x)}_{m \times 1} , \underbrace{\diag( h'(A_2 f(x)) )}_{m \times m} \cdot \underbrace{A_2}_{m \times n} \cdot \underbrace{ p(x)_i }_{n \times 1} \rangle
    \end{align*}
    \item Part 4. Let $h'(A_2 f(x)) \in \R^m$ 
    \begin{align*}
       \underbrace{ \frac{\d h'(A_2 f(x))}{\d x_i} }_{m \times 1} =  \underbrace{ \diag( h''(A_2 f(x)) ) }_{m \times m} \cdot \underbrace{ A_2 }_{m \times n} \cdot \underbrace{ p(x)_i }_{n \times 1}
    \end{align*} 
\end{itemize}
\end{lemma}

\subsection{Hessian}
\label{sub:gradient_hessian:he}
Now, we display the Hessian, where the full details can be seen in Section~\ref{sec:app_hessian}.

\begin{definition}\label{def:Q_q}
We consider that $h$ and $c(x)$ are established in Definition~\ref{def:h} and \ref{def:c}.
 
    To simplify our calculations, we define $Q_2(x) \in \R^{m \times n}$ and $q_2(x) \in \R^n$ as following
\begin{itemize}
    \item $Q_2(x) = \underbrace{\diag(h'( A_2 f( x) ))}_{m \times m} \cdot \underbrace{A_2}_{m \times n} \cdot $
    \item $q_2(x) = \underbrace{Q_2(x)^\top}_{n \times m} \underbrace{c(x)}_{m \times 1}$
\end{itemize}
\end{definition}

\begin{definition}\label{def:g}
We consider that $f(x)$, $c(x)$, $L(x)$, $q_2(x)$ are established in Definition~\ref{def:f}, \ref{def:c}, \ref{def:L}, and \ref{def:Q_q}.
 
Let $g(x) \in \R^d$ be
    \begin{align*}
        g(x):= - \underbrace{A_1^\top}_{d \times n} (\underbrace{f(x)}_{n \times 1} \underbrace{\langle q_2(x),f(x) \rangle}_{\mathrm{scalar}} + \underbrace{\diag(f(x))}_{n \times n} \underbrace{q_2(x)}_{n \times 1})
    \end{align*}
    For each entry $g(x)_i$ of $g(x)$, where $i \in [d]$, we have
    \begin{align*}
        g(x)_i:= - \underbrace{\langle A_{1,*,i}, f(x) \rangle}_{\mathrm{scalar}} \underbrace{\langle q_2(x), f(x) \rangle}_{\mathrm{scalar}} + \underbrace{\langle A_{1,*,i}, f(x) \circ q_2(x) \rangle}_{\mathrm{scalar}}
    \end{align*}
\end{definition}

\subsection{Re-oragnizing \texorpdfstring{$B(x)$}{}}
\label{sub:gradient_hessian:re}

In this section, since the expression of Hessian is too complicated, we reorganize these terms.

\begin{definition}[A variation of Definition 6.1 in \cite{dls23}]\label{def:B}
We define $B(x)$ as follows
\begin{align*}
    B(x) = & ~ \sum_{i=1}^{12} B_i
\end{align*}
where
\begin{align*}
    B_1 = & ~ \diag(f(x))^\top Q_2(x)^\top  Q_2(x) \diag(f(x)) \\
    B_2 = & ~ \diag(f(x))^\top Q_2(x)^\top Q_2(x) f(x) f(x)^\top\\
    B_3 = & ~ f(x) f(x)^\top Q_2(x)^\top  Q_2(x) \diag(f(x))\\
    B_4 = & ~ f(x)^\top Q_2(x)^\top Q_2(x) f(x) f(x)^\top\\
    B_5 = & ~ 2 f(x) c(x)^\top Q_2(x) f(x) f(x)^\top\\
    B_6 = & ~ 2 f(x) c(x)^\top Q_2(x) \diag(f(x))\\
    B_7 = & ~ \diag(Q_2(x)^\top c(x)) \diag(f(x))\\
    B_8 = & ~ \diag(f(x)) A_2 \diag( h''(A_2 f(x)) ) \\
    \cdot & ~ \diag(c(x)) \diag(f(x))\\
    B_9 = & ~ \diag(f(x)) A_2 \diag( h''(A_2 f(x)) ) \\
    \cdot & ~ \diag(c(x)) f(x) f(x)^\top\\
    B_{10} = & ~ f(x) f(x) A_2 \diag( h''(A_2 f(x)) ) \\
    \cdot & ~ \diag(c(x)) \diag(f(x))\\
    B_{11} = & ~ f(x) f(x) A_2 \diag( h''(A_2 f(x)) ) \\
    \cdot & ~ \diag(c(x)) f(x) f(x)^\top\\
    B_{12} = & ~ \diag(f(x) f(x)^\top Q_2(x)^\top c(x)).
\end{align*}
\end{definition}

\begin{lemma}[Informal version of Lemma~\ref{lem:reorganizing}]
    Let $\frac{\d^2 L(x)}{\d x_i \d x_j}$ be computed in Lemma~\ref{lem:hessian_L_informal}, then, we have
    \begin{align*}
        \frac{\d^2 L}{\d x^2}
        = A^\top B(x) A
    \end{align*}
\end{lemma}

\section{Hessian Properties}
\label{sec:hessian_prop}

For Section~\ref{sub:hessian_prop:psd}, we demonstrate the Hessian matrix is positive-semidefinite. For Section~\ref{sub:hessian_prop:lip}, we establish that the Hessian matrix is Lipschitz continuous.

\subsection{Hessian is Positive definite}
\label{sub:hessian_prop:psd}

Here, we present that the Hessian matrix is positive-semidefinite, where the details can be seen in Appendix~\ref{sec:psd_app}. 

\begin{lemma}[Informal version of Lemma~\ref{lem:final_bound}]\label{lem:final_bound_informal}

Suppose that
    \begin{itemize}
        \item $B(x)$ is established in Definition~\ref{def:B}
        \item Let $\beta \in (0, 0.1)$
        \item Let $R \geq 4$
        \item Let $R_f:= 2 \beta^{-2} \cdot n R\exp(2 R^2)$
        \item Let $R_h > 0$.
        \item Let $\max\{ \| h(A_2 f(x)) \|_2, \|h'(A_2 f(x))\|_2 \} \leq R_h$.
        \item Assume $\| h(x) - h(y) \|_2 \leq L_h \cdot \| x -y \|_2$
        \item Assume $\| h'(x)- h'(y) \|_2 \leq L_h \cdot \| x - y \|_2$
    \end{itemize}
    Then we have
    \begin{align*}
      -12 R_h L_h R (R + R_h) I_n \preceq  B(x) \preceq 12 R_h L_h R (R + R_h) I_n
    \end{align*}
\end{lemma}

\subsection{Hessian is Lipschitz}
\label{sub:hessian_prop:lip}

Here, we present that the Hessian matrix is Lipschitz continuous, where the details can be seen in Appendix~\ref{sec:hessian_is_lip_app}.

\begin{lemma}\label{lem:hessian_lip}
Suppose that
    \begin{itemize}
        \item $B(x)$ is established in Definition~\ref{def:B}
        \item Let $\beta \in (0, 0.1)$ 
        \item Let $R \geq 4$
        \item Let $R_f:= 2 \beta^{-2} \cdot n R\exp(2 R^2)$
        \item Let $R_h > 0$.
        \item Let $\max\{ \| h(A_2 f(x)) \|_2, \|h'(A_2 f(x))\|_2 \} \leq R_h$.
        \item Assume $\| h(x) - h(y) \|_2 \leq L_h \cdot \| x -y \|_2$
        \item Assume $\| h'(x)- h'(y) \|_2 \leq L_h \cdot \| x - y \|_2$
    \end{itemize}
    Then we have
    \begin{align*}
        & ~ \| \nabla^2 L(x) - \nabla^2 L(y)\| \\
        \leq & ~ 59(R + R_h) n^2  \exp(4R^2) \beta^{-4} R^5 R_h^2 R_f L_h \|x - y\|_2
    \end{align*}
\end{lemma}
\begin{proof}
    We give an overview of this proof. For specific details, please see Appendix~\ref{sec:hessian_is_lip_app}.

    We have
    \begin{align*}
        & ~ \| \nabla^2 L(x) - \nabla^2 L(y)\|\\
        = & ~ \|\sum_{i=1}^6 (G_i(x) - G_i(y))\| \\
        \leq & ~ 59(R + R_h) n^2  \exp(4R^2) \beta^{-4} R^5 R_h^2 R_f L_h \|x - y\|_2,
    \end{align*}
    which follows from Lemma~\ref{lem:eight_steps}.
\end{proof}

\section{Approximate Newton Method}\label{sec:newton}

In Section~\ref{sec:newton:definitions}, we present basic definitions along with the update rule. In Section~\ref{sec:newton:approximation} covers the approximation of the Hessian matrix and the associated update rule.

\subsection{Update Rule in Newton Method}\label{sec:newton:definitions}

In this section, we study the Newton method's local convergence. Our focus is on the optimization of the target function given by
\begin{align*}
\min_{x \in \R^d } L(x)
\end{align*}
under the following set of assumptions:
\begin{definition}[$(l,M)$-good Loss function]\label{def:f_ass}

Let $L : \R^d \rightarrow \R$. 

We define $L$ as $(l,M)$-good if it meets the following criteria:
\begin{itemize}
\item For a positive scalar $l > 0$, if there is a vector $x^* \in \R^d$ satisfying:
\begin{itemize}
\item $\nabla L(x^*) = {\bf 0}_d$.
\item $\nabla^2 L(x^*) \succeq l \cdot I_d$.
\end{itemize}
\item If there is $M > 0$ satisfying:
\begin{align*}
\| \nabla^2 L(y) - \nabla^2 L(x) \| \leq M \cdot \| y - x \|_2
\end{align*}
\item Denote the initialization point as $x_0$. If $r_0 := \| x_0 -x^*\|_2$ satisfying:
\begin{align*}
r_0 M \leq 0.1 l
\end{align*}
\end{itemize}
\end{definition}

We define gradient and Hessian as follows
\begin{definition}[Hessian and Gradient] 
Let $g : \R^d \rightarrow \R^d$ and $H : \R^d \rightarrow \R^{d \times d}$.

We define
\begin{align*}
g(x) := \nabla L(x)
\end{align*}
to be the gradient of $L(x)$.

We define 
\begin{align*}
H(x) := \nabla^2 L(x)
\end{align*}
to be the Hessian of $L(x)$.

\end{definition}

Considering the gradient function $g: \R^d \rightarrow \R^d$ and the Hessian matrix $H : \R^d \rightarrow \R^{d \times d}$, the exact steps of the Newton method are described as follows:
\begin{definition}\label{def:exact_update_variant}
We define
\begin{align*}
    x_{t+1} = x_t - H(x_t)^{-1} \cdot g(x_t)
\end{align*}
\end{definition}

\subsection{Update Rule and Hessian Approximation}\label{sec:newton:approximation}

In practice, computing the exact $\nabla^2 L(x_t)$ or $(\nabla^2 L(x_t))^{-1}$ is extremely challenging and resource-intensive. Therefore, it's practical to explore approximated calculations for the gradient and Hessian. This computation is outlined as:
\begin{definition}[Approximate Hessian]\label{def:wt_H}
We denote an approximate Hessian $\widetilde{H}(x_t) \in \R^{d \times d}$ for any given Hessian $H(x_t) \in \R^{d \times d}$ as a matrix satisfying the following condition:
\begin{align*}
(1-\epsilon_0) \cdot H(x_t) \preceq \widetilde{H}(x_t) \preceq (1+\epsilon_0) \cdot H(x_t) .
\end{align*}
\end{definition}

To obtain the approximate Hessian $\widetilde{H}(x_t)$, we present a tool from Lemma~4.5 in \cite{dsw22}.
\begin{lemma}[\cite{dsw22,syyz22}]\label{lem:subsample}

We use $\epsilon_0 = 0.01$ to represent the constant precision parameter. Consider $A \in \R^{n \times d}$, and for all $D \in \R^{n \times n}$ being a positive diagonal matrix, there is an algorithm with
\begin{align*}
O( (\nnz(A) + d^{\omega} ) \poly(\log(n/\delta)) )
\end{align*}
running time. This algorithm generates a matrix $\widetilde{D} \in \R^{n \times n}$, which is $O(d \log(n/\delta))$ sparse diagonal, such that
\begin{align*}
(1- \epsilon_0) A^\top D A \preceq A^\top \wt{D} A \preceq (1+\epsilon_0) A^\top D A.
\end{align*}

Here, $\omega \approx 2.373$ is the exponent of matrix multiplication \cite{w12,lg14,aw21}.
\end{lemma}

\begin{lemma}[Iterative shrinking Lemma, Lemma 6.9 on page 32 of \cite{lsz23}]\label{lem:one_step_shrinking}

Suppose $\epsilon_0 \in (0,0.1)$, $r_t:= \| x_t - x^* \|_2$, and $\ov{r}_t: = M \cdot r_t$.

Then, 
\begin{align*}
r_{t+1} \leq 2 \cdot (\epsilon_0 + \ov{r}_t/( l - \ov{r}_t ) ) \cdot r_t,
\end{align*} 
where $L$ is $(l,M)$-good.

\end{lemma}

The total number of iterations in the algorithm is represented by $T$. To utilize Lemma~\ref{lem:one_step_shrinking}, we require the following lemma based on the induction hypothesis. This lemma is found in \cite{lsz23}.

\begin{lemma}[Induction hypothesis of \cite{lsz23} (see Lemma 6.10 on page 34)]\label{lem:newton_induction}

Let $i \in [t]$.

Let $r_i:= \| x_i - x^* \|_2$. 

By Definition~\ref{def:f_ass} and Definition~\ref{def:wt_H}, we suppose $\epsilon_0 = 0.01$, for all $i$, $r_{i} \leq 0.4 \cdot r_{i-1}$ and $M \cdot r_i \leq 0.1 l$.

Then we have
\begin{itemize}
    \item $r_{t+1} \leq 0.4 r_t$
    \item $M \cdot r_{t+1} \leq 0.1 l$
\end{itemize}
\end{lemma}

\section{Main Results}
\label{sec:main}

In this section, we present our main result.

\begin{algorithm}[!ht]\caption{Main algorithm.}\label{alg:main}
    \begin{algorithmic}[1]
        \Procedure{OurAlgorithm}{$b \in \R^n, A \in \R^{n \times d},w \in \R^n, \epsilon, \delta$} \Comment{Theorem~\ref{thm:theorem_main_formal}} 
            \State Choose $x_0$ and suppose $x_0$ satisfies Definition~\ref{def:f_ass})
            \State $T \gets \log( \| x_0 - x^* \|_2 / \epsilon )$ is the number of iterations.
            \For{$t=0 \to T$} 
                \State $D \gets B(x_t) + \diag(w \circ w)$ 
                \State $\wt{D} \gets \textsc{SubSample}(D,A,\epsilon_1 = \Theta(1), \delta_1 = \delta/T)$ \Comment{Lemma~\ref{lem:subsample}}
                \State $g \gets - A_1^\top(f(x)\langle q_2(x),f(x) \rangle + \diag(f(x)) q_2(x))$ 
                \State $\wt{H} \gets A^\top \wt{D} A$ 
                \State $x_{t+1} \gets x_t + \wt{H}^{-1} g$ 
            \EndFor
            \State $\wt{x}\gets x_{T+1}$
            \State \Return $\wt{x}$
        \EndProcedure
    \end{algorithmic}
\end{algorithm}

\begin{theorem}[Formal version of Theorem~\ref{thm:theorem_main_informal}]\label{thm:theorem_main_formal}
    If the following conditions hold
    \begin{itemize}
        \item We have $L(x)$ be established in Definition~\ref{def:L}
        \item Suppose $\|A\| \leq R$
        \item Suppose $\|x\|_2 \leq R$
        \item $x^*$ represents the solution of $\min_{x \in \R^d} L(x)$
        \item Let $l$ be a scalar such that $w_i^2 \ge 12 R_h L_h R (R + R_h) + l/\sigma_{\min}(A)^2$ for $\forall i \in [n]$
        \item Suppose $x_0$ be the initial point that satisfies $M \|x_0 - x^*\|_2 \leq 0.1 l$.
        \item Let $M = 59(R + R_h) n^2  \exp(4R^2) \beta^{-4} R^5 R_h^2 R_f L_h$ 
    \end{itemize}

    For any accuracy parameter $\epsilon \in (0, 0.1)$ and a failure probability $\delta \in (0, 0.1)$, there exists a randomized algorithm (Algorithm~\ref{alg:main}) that performs $T = O(\log(\|x_0 - x^*\|_2 / \epsilon))$ iterations. Each iteration requires $O((\nnz(C) + d^\omega) \poly(\log(m / \delta)))$ computational steps. The output $\wt{x}$ from this algorithm satisfies $\Pr[\|\wt{x} - x^*\|_2 \leq \epsilon] \ge 1 - \delta$, where $\omega$ represents the exponent for matrix multiplication.
\end{theorem}
\begin{proof}

    This can be proved by combining Lemma~\ref{lem:final_bound_informal}, Lemma~\ref{lem:newton_induction}, Lemma~\ref{lem:subsample}, Lemma~\ref{lem:hessian_lip} and
    Lemma~\ref{lem:one_step_shrinking}.

    \begin{itemize}
        \item The upper bound on $M$: Lemma~\ref{lem:eight_steps} and $M$-lipschitz definition.
        \item Hessian is PD: Lemma~\ref{lem:final_bound_informal}
        \item Hessian is Lipschitz: Lemma~\ref{lem:hessian_lip}
        \item Cost per iteration: Lemma~\ref{lem:subsample}
        \item Convergence per Iteration: Lemma~\ref{lem:one_step_shrinking}. We can get
        \begin{align*}
             \|x_k - x^*\|_2 \le 0.4 \cdot \|x_{k-1} - x^*\|_2.
        \end{align*}
        \item Number of iterations: we can get
        \begin{align*}
            \| x_T - x^* \|_2 \leq 0.4^T \cdot \| x_0 - x^* \|_2
        \end{align*}
        after $T$ iterations. Choosing the value of $T$ allows us to achieve the intended limit. The failure probability is determined by employing a union bound over the $T$ iterations.
    \end{itemize}
\end{proof}

\section{Conclusion}
\label{sec:conclusion}

In this paper, we formulated a two-layer regression model with softmax and Lipschitz continuous activations. We derived key properties of the loss landscape, such as the positive definiteness and Lipschitz continuity of the Hessian matrix. These findings ensure the convexity and smoothness necessary for optimization convergence guarantees. We utilized an approximate Newton method to minimize the regularized training loss and established its local convergence rate. Under reasonable assumptions, our algorithm discovers an $\epsilon$-approximate minimizer within $O(\log(1/\epsilon))$ iterations, with nearly linear computational cost per iteration. Our general framework accommodates an outer activation that can be any arbitrary Lipschitz function, facilitating extensions to other nonlinear units. This adaptability proves crucial for applying our approach across diverse applications. By combining matrix analysis and optimization, our techniques can serve as a blueprint for analyzing deeper neural network architectures. In summary, our paper takes significant strides in comprehending optimization and generalization in multilayer nonlinear networks. We believe that our analyses and algorithm provide a principled approach, forming a solid foundation for addressing more intricate models and tasks in the future.

\ifdefined\isarxiv
%\section*{Acknowledgments}
% \bibliographystyle{alpha}
% \bibliography{ref}
\else
\bibliography{ref}
\bibliographystyle{ieeenat_fullname}%{alpha}

\fi

\newpage
\onecolumn
\appendix

\paragraph{Roadmap.}
In Section~\ref{sec:preli}, we introduce the basic notations and the important algebraic facts about matrices, vectors, and their derivatives. In Section~\ref{sec:app_gradient}, we compute the gradient of the important functions. In Section~\ref{sec:app_hessian}, we compute the Hessian of the important functions. In Section~\ref{sec:psd_app}, we show that the Hessian matrix is positive definite. In Section~\ref{sec:hessian_is_lip_app}, we show that the Hessian matrix is Lipschitz.

\section{Preliminary}
\label{sec:preli}

In this section, we first introduce the basic notations. Then, in Section~\ref{sub:preli:fact}, we present the important mathematical facts about matrices and vectors.

\paragraph{Notations.}

Let $x \in \R^{d}_{>0}$ denote a length-$n$ where all the entries are positive.
For a matrix $A \in \R^{n \times d}$, $\exp(A)$ denotes the $n \times d$ matrix given by $\exp(A) = \sum_{i=0}^{\infty} \frac{1}{i!}A^i$.
We use ${\bf 1}_n$ to denote a length-$n$ vector where all its entries are $1$.
For an arbitrary vector $x$, we denote its $\ell_p$ norm by $\|x\|_p$, where $\|x\|_1 := \sum_{i=1}^n |x_i|, \|x\|_2 := (\sum_{i=1}^n x_i^2)^{1/2}$, and $\|x\|_\infty := \max_{i \in [n]} |x_i|$.
For a matrix $A \in \R^{n \times k}$ where $n > k$, we denote $\|A\|$ to be the spectral norm of $A$ and $\|A\|:= \sup_{x \in \R^k} \|Ax\|_2 / \|x\|_2$
For two vectors $a, b \in \R^n$, we define $\langle a, b \rangle := \sum_{i=1}^n a_ib_i$.
For two vectors $a, b \in \R^n$, we use $a \circ b$ to denote the vector where its $i$-th entry is $a_i b_i$ for $i \in [n]$.
For $x \in \R^n$, we define the diagonal matrix $\diag(x) \in \R^{n \times n}$, whose diagonal entries are given by $\diag(x)_{i,i} = x_i$ for $i \in [n]$, and the entries elsewhere are all zero.
For a symmetric matrix $A \in \R^{n \times n}$ with real entries, we denote $A \succ 0$ to indicate the matrix is positive-definite (PD), if $x^\top A x > 0$ for all $x \in \R^n$.
For a symmetric matrix $A \in \R^{n \times n}$ with real entries, we denote $A \succeq 0$ to indicate the matrix is positive-semidefinite (PSD), if $x^\top A x \geq 0$ for all $x \in \R^n$.

\subsection{Basic Fact}
\label{sub:preli:fact}

In this section, we present the important mathematical properties.

\begin{fact}\label{fac:basic_algebra}
Let $a,b, c \in \R^n$ denote three column vectors. 
We have
\begin{itemize}
    \item $\underbrace{ \langle a, b \rangle }_{ \mathrm{scalar} } \underbrace{ c }_{ n \times 1 } = \underbrace{ a^\top b }_{\mathrm{scalar}} \underbrace{ c }_{n \times 1} = \underbrace{ c }_{n \times 1} \underbrace{ a^\top }_{1 \times n} \underbrace{ b }_{ n \times 1} = \underbrace{ c }_{n \times 1} \underbrace{ b^\top }_{1 \times n} \underbrace{ a }_{n \times 1}$
    \item $\underbrace{a \circ b}_{n \times 1} = \underbrace{b \circ a}_{n \times 1} = \underbrace{\diag(a)}_{n \times n} \underbrace{b}_{n \times 1} = \underbrace{\diag(b)}_{n \times n} \underbrace{a}_{n \times 1}$
    \item $\underbrace{a^\top}_{1 \times n} \underbrace{(b \circ c)}_{n \times 1} = \underbrace{b^\top}_{1 \times n} \underbrace{(a \circ c)}_{n \times 1} = \underbrace{c^\top}_{1 \times n} \underbrace{(a \circ b)}_{n \times 1}$
    \item $\underbrace{\diag(a \circ b)}_{n \times n} = \underbrace{\diag(a)}_{n \times n} \underbrace{\diag(b)}_{n \times n}$
    \item $\underbrace{\diag(a + b)}_{n \times 1} = \underbrace{\diag(a)}_{n \times 1} + \underbrace{\diag(b)}_{n \times 1}$
    \item $\langle a, b \rangle + \langle c, b \rangle = \langle a + c, b \rangle = \langle b, a + c \rangle = \langle b, a \rangle + \langle b, c \rangle$.
\end{itemize}
\end{fact}

Now, we present the derivative rules for matrices.

\begin{fact}\label{fac:basic_calculus}
We have
\begin{itemize}
    \item $\frac{\d \langle f(x), g(x) \rangle}{\d t} = \langle \frac{\d f(x)}{\d t} , g(x) \rangle + \langle f(x), \frac{\d g(x)}{\d t} \rangle$
    \item $\frac{\d}{\d t} ( f(x) + g(x) ) = \frac{\d f(x)}{\d t} + \frac{\d g(x)}{\d t}$
    \item $\frac{\d }{\d t} ( f(x) \circ g(x)  ) = f(x) \circ \frac{\d g(x)}{\d t} + g(x) \circ \frac{\d f(x)}{\d t} $
\end{itemize}
\end{fact}

\begin{fact}\label{fac:vec_norm}
    For two length-$n$ column vectors $u, v \in \R^n$, we have
    \begin{itemize}
        \item $\langle u,v \rangle \leq \|u\|_2 \cdot \|v\|_2$ (Cauchy-Schwarz Inequality)
        \item $\langle u, v \rangle = \langle u \circ v, {\bf 1}_n \rangle$
        \item for all real number $a$, $\|a u\|_2 = |a| \cdot \|u\|_2$
        \item $\|u^\top\|_2 = \|u\|_2$
        \item $\|u + v\|_2 \leq \|u\|_2 + \|v\|_2$
        \item $\|u \circ v\|_2 \leq \|u\|_\infty \cdot \|v\|_2$
        \item $\|\diag(u)\| \leq \|u\|_{\infty}$
        \item $\| u \|_{\infty} \leq \| u \|_2 \leq \sqrt{n} \cdot \| u \|_{\infty}$
        \item $\| u \|_2 \leq \| u \|_1 \leq \sqrt{n} \cdot \| u \|_2$
        \item $\| \exp(u) \|_{\infty} \leq \exp(\| u \|_\infty) \leq \exp(\| u \|_2)$
        \item if $\| u \|_2, \|v\|_2 \leq R$, then $\| \exp(u) - \exp(v) \|_2 \leq \exp(R) \cdot \| u - v \|_2$, for all $R \geq 4$.
    \end{itemize}
\end{fact}

\begin{fact}\label{fac:matrix_norm}
    For arbitrary matrices $A$ and $B$, we have
    \begin{itemize}
        \item For a scalar $c \in \R$, we have $\|c \cdot A\| \leq |c| \cdot \|A\|$
        \item $\|A^\top\| = \|A\|$
        \item $\|A + B\| \leq \|A\| + \|B\|$
        \item $\|A \cdot B\| \leq \|A\| \cdot \|B\|$
        \item For any vector $x$, we have $\|Ax\|_2 \leq \|A\| \cdot \|x\|_2$
        \item For two vectors $a, b \in \R^n$, we have $\|ab^\top\| \leq \|a\|_2 \|b\|_2$
    \end{itemize}
\end{fact}

\begin{fact}\label{fac:psd}
    For two length-$n$ column vectors $u,v \in \R^n$, we have
    \begin{itemize}
        \item $uu^\top \preceq \|u\|_2^2 \cdot I_n$. Here $I_n \in \R^{n \times n}$ denotes an identity matrix. 
    \end{itemize}
\end{fact}

\section{Gradient}
\label{sec:app_gradient}

In this section, we present the gradient of our loss function $L$: to do this, we break up the loss function into simpler forms, compute the gradient of each of these forms, and combine the together to form the gradient for $L$.

\begin{lemma}\label{lem:gradient}
If the following holds
\begin{itemize}
    \item Let $u(x)$ be defined as Definition~\ref{def:u}.
    \item Let $\alpha(x)$ be defined as Definition~\ref{def:alpha}
    \item Let $f(x)$ be defined as Definition~\ref{def:f}
    \item Let $h : \R^m \rightarrow \R^m$ be defined as Definition~\ref{def:h}
    \item Let $c(x)$ be defined as Definition~\ref{def:c}
    \item Let $L(x)$ be defined as Definition~\ref{def:L}
    \item Let $A_{1,*,i} \in \R^n$ denote the $i$-th column vector of $A_1 \in \R^{n \times d}$ for all $i \in [d]$
    \item Let $A_{1,l,*} \in \R^d$ denote the $l$-th row vector of $A_1 \in \R^{n \times d}$ for all $l \in [n]$
    \item Let $A_{2,k,*} \in \R^n$ denote the $k$-th row vector $A_2 \in \R^{m \times n}$ for each $k \in [m]$
\end{itemize}
then for each $i \in [d]$, we have
\begin{itemize}
    \item Part 1.
    \begin{align*}
    \frac{\d u(x)}{\d x_i} = u(x) \circ A_{1,*,i}
    \end{align*}
    \item Part 2.
    \begin{align*}
        \frac{\d \alpha(x)}{\d x_i} = \langle u(x), A_{1,*, i} \rangle
    \end{align*}
    \item Part 3.
    \begin{align*}
        \frac{\d \alpha(x)^{-1}}{\d x_i} = -\alpha(x)^{-1} \cdot \langle f(x), A_{1,*, i} \rangle
    \end{align*}
    \item Part 4. Let $p(x)_i \in \R^n$ be defined as $p(x)_i := f(x) \circ A_{1,*,i} -\langle f(x), A_{1,*,i} \rangle \cdot f(x)$
    \begin{align*}
        \frac{\d f(x)}{\d x_i} = p(x)_i
    \end{align*}
    \item Part 5.
    \begin{align*}
        \frac{\d \langle f(x), A_{1,*,i} \rangle }{\d x_i} = -\langle f(x), A_{1,*,i} \rangle^2 + \langle f(x), A_{1,*,i} \circ A_{1,*,i} \rangle
    \end{align*}
    \item Part 6. For each $j \in [d]$ and $j \neq i$,
    \begin{align*}
        \frac{\d \langle f(x), A_{1,*,i} \rangle }{\d x_j} = -\langle f(x), A_{1,*,i} \rangle \cdot \langle f(x), A_{1,*,j} \rangle + \langle f(x), A_{1,*,i} \circ A_{1,*,j} \rangle
    \end{align*}  
    \item Part 7. Let $h'(A_2 f(x)) \in \R^m$ denote a length-$m$ vector $i$-th coordinate is the $\frac{\d h(y_i)}{\d y_i} \big|_{y_i = (A_2 f(x))_i}$ (here we can should think of $h : \R \rightarrow \R$), we have
    \begin{align*}
        \underbrace{ \frac{\d h(A_2 f(x))}{\d x_i} }_{m \times 1} = \underbrace{ \diag( h'(A_2 f(x)) ) }_{m \times m} \cdot \underbrace{ A_2 }_{m \times n} \cdot \underbrace{ p(x)_i }_{n \times 1}
    \end{align*} 
     \item Part 8. 
    \begin{align*}
        \frac{\d c(x)}{\d x_i} = \frac{\d h(A_2 f(x))}{ \d x_i}
    \end{align*}
    \item Part 9.  
    \begin{align*}
        \frac{\d L(x)}{\d x_i} = \langle \underbrace{c(x)}_{m \times 1} , \underbrace{\diag( h'(A_2 f(x)) )}_{m \times m} \cdot \underbrace{A_2}_{m \times n} \cdot \underbrace{ p(x)_i }_{n \times 1} \rangle
    \end{align*}
    \item Part 10. Let $h'(A_2 f(x)) \in \R^m$ 
    \begin{align*}
       \underbrace{ \frac{\d h'(A_2 f(x))}{\d x_i} }_{m \times 1} =  \underbrace{ \diag( h''(A_2 f(x)) ) }_{m \times m} \cdot \underbrace{ A_2 }_{m \times n} \cdot \underbrace{ p(x)_i }_{n \times 1}
    \end{align*} 
\end{itemize}
\end{lemma}
\begin{proof}

{\bf Proof of Part 1.}

We have
\begin{align}\label{eq:dux_dxi}
    \frac{\d u(x) }{\d x_i}
    = & ~ \frac{\d (\exp(A_1 \cdot x))}{\d x_i} \notag\\
    = & ~ \exp(A_1 x) \circ \frac{\d(A_1 x)}{\d x_i}\notag\\
    = & ~ \exp(A_1 x) \circ (A_1 \cdot \frac{\d x}{\d x_i}),
\end{align}
where the first step follows from the definition of $u(x)$ (see Definition~\ref{def:u}), the second step follows from the chain rule, and the third step follows from simple algebra.

Furthermore, we have
\begin{align*}
    (\frac{\d x}{\d x_i})_i = 1.
\end{align*}
and for $j \neq i$
\begin{align*}
    (\frac{\d x}{\d x_i})_j = 0.
\end{align*}

Therefore, we have
\begin{align*}
    \frac{\d x}{\d x_i} = e_i,
\end{align*}
which implies that
\begin{align}\label{eq:A_1_dx_dxi}
    A_1 \cdot \frac{\d x}{\d x_i} = A_{1, *, i}.
\end{align}

Therefore, by combining Eq.~\eqref{eq:dux_dxi} and Eq.~\eqref{eq:A_1_dx_dxi}, we have
\begin{align*}
    \frac{\d u(x)}{\d x_i} = u(x) \circ A_{1,*, i}.
\end{align*}

{\bf Proof of Part 2.}

We have
\begin{align*}
    \frac{\d \alpha(x)}{\d x_i}
    = & ~ \frac{\d \langle u(x), {\bf 1}_n \rangle}{\d x_i}\\
    = & ~ \langle \frac{\d u(x)}{\d x_i}, {\bf 1}_n \rangle + \langle u(x), \frac{\d {\bf 1}_n}{\d x_i} \rangle\\
    = & ~ \langle u(x) \circ A_{1,*, i}, {\bf 1}_n \rangle\\
    = & ~ \langle u(x), A_{1,*, i} \rangle,
\end{align*}
where the first step follows from the definition of $\alpha(x)$ (see Definition~\ref{def:alpha}), the second step follows from Fact~\ref{fac:basic_calculus}, the third step follows from Part 1 and $\frac{\d {\bf 1}_n}{\d x_i} = {\bf 0}_n$, and the last step follows from Fact~\ref{fac:vec_norm}.

{\bf Proof of Part 3.}
\begin{align*}
    \frac{\d \alpha(x)^{-1}}{\d x_i}
    = & ~ - \alpha(x)^{-2} \cdot \frac{\d}{\d x_i} \alpha(x)\\
    = & ~ - \alpha(x)^{-2} \langle u(x), A_{1,*,i} \rangle \\
    = & ~ - \alpha(x)^{-1} \langle f(x), A_{1,*,i}\rangle
\end{align*}
where the 1st step follows from the differential rules, the 2nd step follows from the result of Part 2, and the last step follows from Definition~\ref{def:f}.

{\bf Proof of Part 4.}
\begin{align*}
    \frac{\d f(x)}{\d x_i}
    = & ~ \frac{\d (\alpha(x)^{-1} u(x))}{\d x_i} \\
    = & ~ u(x) \cdot \frac{\d \alpha(x)^{-1}}{\d x_i} + \alpha(x)^{-1} \cdot \frac{\d u(x)}{\d x_i}\\
    = & ~ - \alpha(x)^{-2} \langle u(x), A_{1,*,i} \rangle \cdot u(x) + \alpha(x)^{-1} \cdot u(x) \circ A_{1,*,i}\\
    = & ~ - \langle f(x), A_{1,*,i} \rangle \cdot f(x) + f(x) \circ A_{1,*,i}
\end{align*}
where the 1st step follows from Definition~\ref{def:f}, the 2nd step follows from the differential chain rule, the 3rd step follows from the results of Part 1 and Part 3, and the last step follows from Definition~\ref{def:f}.

{\bf Proof of Part 5.}
\begin{align*}
    \frac{\d \langle f(x), A_{1,*,i} \rangle}{\d x_i}
    = & ~ A^\top_{1,*,i} \frac{\d f(x)}{\d x_i}\\
    = & ~ A^\top_{1,*,i} (-\langle f(x), A_{1,*,i} \rangle \cdot f(x) + f(x) \circ A_{1,*,i})\\
    = & ~ - \langle f(x), A_{1,*,i} \rangle \cdot A^\top_{1,*,i} f(x) + A^\top_{1,*,i}f(x) \circ A_{1,*,i}\\
    = & ~ - \langle f(x), A_{1,*,i} \rangle^2 + \langle f(x), A_{1,*,i} \circ A_{1,*,i} \rangle
\end{align*}
where the 1st step follows from $\langle u,v \rangle = v^\top u$, the 2nd step follows from the result of Part 4, the 3rd step follows from the distributive property of algebra, and the last step follows from $\langle u,v \rangle = u^\top v = v^\top u$.

{\bf Proof of Part 6.}
For $j \in [d]$, $i\in [d]$ and $j \neq i$
\begin{align*}
    \frac{\d \langle f(x), A_{1,*,i} \rangle}{\d x_j}
    = & ~ A^\top_{1,*,i} \frac{\d f(x)}{\d x_j}\\
    = & ~ A^\top_{1,*,i} (-\langle f(x), A_{1,*,j} \rangle \cdot f(x) + f(x) \circ A_{1,*,j})\\
    = & ~ - \langle f(x), A_{1,*,j} \rangle \cdot A^\top_{1,*,i} f(x) + A^\top_{1,*,i}f(x) \circ A_{1,*,j}\\
    = & ~ - \langle f(x), A_{1,*,j} \rangle \cdot \langle f(x), A_{1,*,i} \rangle + \langle A_{1,*,i}, f(x) \circ A_{1,*,j} \rangle\\
    = & ~ - \langle f(x), A_{1,*,i} \rangle \cdot \langle f(x), A_{1,*,j} \rangle + \langle f(x), A_{1,*,i} \circ A_{1,*,j} \rangle
\end{align*}
where the 1st step follows from $\langle u,v \rangle = v^\top u$, the 2nd step follows from the result of Part 4, the 3rd step follows from the distributive property of algebra, and the 4th step follows from $\langle u,v \rangle = u^\top v = v^\top u$, and the last step follows from $\langle u,v \circ w \rangle = \langle v,u \circ w \rangle$.

{\bf Proof of Part 7.}

For $ k \in [m]$, 
    \begin{align*}
        \frac{\d h(A_2 f(x))_k}{\d x_i}
        = & ~ h'(A_2 f(x))_k \cdot \frac{\d (A_2f(x))_k}{\d x_i}\\
        = & ~  h'(A_2 f(x))_k \cdot \frac{ \d \langle A_{2,k,*}^\top , f(x) \rangle  } {\d x_i} \\
        = & ~  h'(A_2 f(x))_k \cdot \langle \underbrace{ A_{2,k,*}^\top }_{n \times 1}, \underbrace{ \frac{\d f(x)}{ \d x_i} }_{n \times 1} \rangle \\
        = & ~  h'(A_2 f(x))_k \cdot \langle  A_{2,k,*}^\top , - \langle f(x), A_{1,*,i} \rangle \cdot f(x) + f(x) \circ A_{1,*,i} \rangle  
    \end{align*}
 where the 1st step follows from Definition~\ref{def:h}, the 2nd step follows from $uv=\langle u^\top, v \rangle$, the 3rd step follows from Fact~\ref{fac:basic_calculus},
 and the last step follows from the result of Part 4. 

Thus, 
\begin{align*}
\underbrace{ \frac{\d h(A_2 f(x))}{\d x_i} }_{m \times 1}
 = \underbrace{ \diag( h'(A_2 f(x)) ) }_{m \times m} \cdot \underbrace{ A_2 }_{m \times n} \cdot  \underbrace{ ( - \langle f(x), A_{1,*,i} \rangle \cdot f(x) + f(x) \circ A_{1,*,i} ) }_{n \times 1}
\end{align*}

{\bf Proof of Part 8.}
\begin{align*}
    \frac{\d c(x)}{\d x_i}
    = & ~ \frac{\d h(A_2 f(x))-b}{\d x_i}\\
    = & ~ \frac{\d h(A_2 f(x))}{\d x_i} - \frac{\d b}{\d x_i}\\
    = & ~ \frac{\d h(A_2 f(x))}{\d x_i} 
\end{align*}
where the 1st step follows from Definition~\ref{def:c}, the 2nd step follows from the differential rules, and the last step follows from simple algebra.

{\bf Proof of Part 9.}
\begin{align*}
    \frac{\d L(x)}{\d x_i}
    = & ~ \frac{\d}{\d x_i} (\frac{1}{2}\|c(x)\|^2_2)\\
    = & ~ (c(x))^\top \frac{\d c(x)}{\d x_i}\\
    = & ~ (c(x))^\top \frac{\d h(A_2 f(x))}{\d x_i}\\
    = & ~ \underbrace{(c(x))^\top}_{1 \times m} \cdot \underbrace{\diag( h'(A_2 f(x)) )}_{m \times m} \cdot  \underbrace{A_2}_{m \times n} \cdot  \underbrace{( - \langle f(x), A_{1,*,i} \rangle \cdot f(x) + f(x) \circ A_{1,*,i} )}_{n \times 1},
\end{align*}
where the 1st step follows from Definition~\ref{def:L}, the 2nd step follows from the differential chain rule, the 3rd step follows from the result of Part 8, and the last step follows from the results of Part 7.

{\bf Proof of Part 10.}

For $k \in [m]$, by using the chain rule, we have
    \begin{align}\label{eq:dh'_dxi}
        \frac{\d h'(A_2 f(x))_k}{\d x_i} = h''(A_2 f(x))_k \cdot \frac{\d (A_2f(x))_k}{\d x_i}.
    \end{align}

    Additionally, by {\bf Proof of Part 7}, we have
    \begin{align*}
        h'(A_2 f(x))_k \cdot \frac{\d (A_2f(x))_k}{\d x_i} = h'(A_2 f(x))_k \cdot \langle  A_{2,k,*}^\top , - \langle f(x), A_{1,*,i} \rangle \cdot f(x) + f(x) \circ A_{1,*,i} \rangle,
    \end{align*}
    which implies that
    \begin{align}\label{eq:dA2_dxi}
        \frac{\d (A_2f(x))_k}{\d x_i} = \langle  A_{2,k,*}^\top , - \langle f(x), A_{1,*,i} \rangle \cdot f(x) + f(x) \circ A_{1,*,i} \rangle.
    \end{align}

    Taking Eq.~\eqref{eq:dA2_dxi} into Eq.~\eqref{eq:dh'_dxi}, we have
    \begin{align*}
        \frac{\d h'(A_2 f(x))_k}{\d x_i} = h''(A_2 f(x))_k \cdot \langle  A_{2,k,*}^\top , - \langle f(x), A_{1,*,i} \rangle \cdot f(x) + f(x) \circ A_{1,*,i} \rangle.
    \end{align*}

Therefore, we have 
\begin{align*}
\underbrace{ \frac{\d h'(A_2 f(x))}{\d x_i} }_{m \times 1} =   \underbrace{ \diag( h''(A_2 f(x)) ) }_{m \times m} \cdot \underbrace{ A_2 }_{m \times n} \cdot \underbrace{ ( - \langle f(x), A_{1,*,i} \rangle \cdot f(x) + f(x) \circ A_{1,*,i} ) }_{n \times 1},
\end{align*}
which completes the proof.
\end{proof}

\section{Hessian}
\label{sec:app_hessian}

In Section~\ref{sub:app_hessian:u}, we compute the Hessian of $u(x)$. In Section~\ref{sub:app_hessian:a}, we compute the Hessian of $\alpha(x)$. In Section~\ref{sub:app_hessian:a-1}, we compute the Hessian of $\alpha(x)^{-1}$. In Section~\ref{sub:app_hessian:f}, we compute the Hessian of $f(x)$. In Section~\ref{sub:app_hessian:l}, we compute the Hessian of $L(x)$. In Section~\ref{sub:app_hessian:B}, we reorganize the expression of the Hessian of $L(x)$ by analyzing $B(x)$.

\subsection{Hessian of \texorpdfstring{$u(x)$}{}}
\label{sub:app_hessian:u}

In this section, we study the Hessian of $u(x)$.

\begin{lemma}
    If the following condition holds
    \begin{itemize}
        \item Let $u(x)$ be defined as Definition~\ref{def:u}.
    \end{itemize}
    Then for each $i\in [d]$ and $j \in [d]$, we have
    \begin{itemize}
        \item Part 1.
        \begin{align*}
            \frac{\d^2 u(x)}{\d x_i^2} = A_{1,*,i} \circ u(x) \circ A_{1,*,i}
        \end{align*}
        \item Part 2.
        \begin{align*}
            \frac{\d^2 u(x)}{\d x_i \d x_j} = A_{1,*,j} \circ u(x) \circ A_{1,*,i}
        \end{align*}
    \end{itemize}
\end{lemma}

\begin{proof}
    {\bf Proof of Part 1.}
    \begin{align*}
        \frac{\d^2 u(x)}{\d x_i^2}
        = & ~ \frac{\d}{\d x_i} (\frac{\d u(x)}{\d x_i})\\
        = & ~ \frac{\d}{\d x_i} (u(x) \circ A_{1,*,i})\\
        = & ~ A_{1,*,i} \circ \frac{\d u(x)}{\d x_i}\\
        = & ~ A_{1,*,i} \circ u(x) \circ A_{1,*,i}
    \end{align*}
    where the 1st step follows from the differential rules, the 2nd step follows from the result of Part 1 in Lemma~\ref{lem:gradient}, the 3rd step follows from Fact~\ref{fac:basic_calculus}, 
    and the last step follows from the result of Part 1 in Lemma~\ref{lem:gradient}.
    
    {\bf Proof of Part 2.}
    \begin{align*}
        \frac{\d^2 u(x)}{\d x_i \d x_j}
        = & ~ \frac{\d}{\d x_i} (\frac{\d u(x)}{\d x_j})\\
        = & ~ \frac{\d}{\d x_i} (u(x) \circ A_{1,*,j})\\
        = & ~ A_{1,*,j} \circ \frac{\d u(x)}{\d x_i}\\
        = & ~ A_{1,*,j} \circ u(x) \circ A_{1,*,i}
    \end{align*}
    where the 1st step follows from the differential rules, the 2nd step follows from the result of Part 1 in Lemma~\ref{lem:gradient}, the 3rd step follows from Fact~\ref{fac:basic_calculus}, and the last step follows from the result of Part 1 in Lemma~\ref{lem:gradient}.    
\end{proof}

\subsection{Hessian of \texorpdfstring{$\alpha(x)$}{}}
\label{sub:app_hessian:a}

In this section, we study the Hessian of $\alpha(x)$.

\begin{lemma}
    If the following condition holds
    \begin{itemize}
        \item Let $\alpha(x)$ be defined as Definition~\ref{def:alpha}.
    \end{itemize}
    Then for each $i\in [d]$ and $j \in [d]$, we have 
    \begin{itemize}
        \item Part 1.
        \begin{align*}
            \frac{\d^2 \alpha(x)}{\d x_i^2} = \langle u(x), A_{1,*i} \circ A_{1,*,i} \rangle
        \end{align*}
        \item Part 2.
        \begin{align*}
            \frac{\d^2 \alpha(x)}{\d x_i \d x_j} = \langle u(x) ,  A_{1,*,i} \circ A_{1,*,j} \rangle
        \end{align*}
    \end{itemize}
\end{lemma}
\begin{proof}
    {\bf Proof of Part 1.}
    \begin{align*}
        \frac{\d^2 \alpha(x)}{\d x_i^2}
        = & ~ \frac{\d}{\d x_i}(\frac{\d \alpha(x)}{\d x_i})\\
        = & ~ \frac{\d \langle u(x), A_{1,*, i} \rangle}{\d x_i}\\
        = & ~ A_{1,*,i}^\top \frac{\d u(x)}{\d x_i}\\
        = & ~ A_{1,*,i}^\top \cdot u(x) \circ A_{1,*,i} \\
        = & ~ \langle u(x), A_{1,*i} \circ A_{1,*,i} \rangle
    \end{align*}
    where the 1st step follows from the differential rules, the 2nd step follows from the result of Part 2 in Lemma~\ref{lem:gradient}, the 3rd step follows from $\langle u,v \rangle = v^\top u$, 
    the fourth step follows from the result of Part 1 in Lemma~\ref{lem:gradient},
    the last step follows from simepl algebra.
    
    {\bf Proof of Part 2.}
    \begin{align*}
        \frac{\d^2 \alpha(x)}{\d x_i \d x_j}
        = & ~ \frac{\d}{\d x_i}(\frac{\d \alpha(x)}{\d x_j})\\
        = & ~ \frac{\d \langle u(x), A_{1,*,j} \rangle}{\d x_i}\\
        = & ~ A_{1,*,j}^\top \frac{\d u(x)}{\d x_i}\\
        = & ~ A_{1,*,j}^\top \cdot u(x) \circ A_{1,*,i} \\
        = & ~ \langle u(x) ,  A_{1,*,i} \circ A_{1,*,j} \rangle
    \end{align*}
    where the 1st step follows from the differential rules, the 2nd step follows from the result of Part 2 in Lemma~\ref{lem:gradient}, the 3rd step follows from $\langle u,v \rangle = v^\top u$, the fourth step follows from the result of Part 1 in Lemma~\ref{lem:gradient},
    the last step follows from simple algebra.
\end{proof}

\subsection{Hessian of \texorpdfstring{$\alpha(x)^{-1}$}{}}
\label{sub:app_hessian:a-1}

In this section, we study the Hessian of $\alpha(x)^{-1}$.

\begin{lemma}
    If the following condition holds
    \begin{itemize}
        \item Let $\alpha(x)$ be defined as Definition~\ref{def:alpha}
        \item Let $f(x)$ be defined as Definition~\ref{def:f}
    \end{itemize}
    Then for each $i\in [d]$ and $j \in [d]$, we have
    \begin{itemize}
        \item Part 1.
        \begin{align*}
            \frac{\d^2 \alpha(x)^{-1}}{\d x_i^2} = 2\alpha(x)^{-1} \cdot \langle f(x), A_{1,*, i} \rangle^2 - \alpha(x)^{-1}\langle f(x), A_{1,*,i} \circ A_{1,*,i} \rangle
        \end{align*}
        \item Part 2.
        \begin{align*}
            \frac{\d^2 \alpha(x)^{-1}}{\d x_i \d x_j} = 2\alpha(x)^{-1} \cdot \langle f(x), A_{1,*,i} \rangle \langle f(x), A_{1,*,j} \rangle - \alpha(x)^{-1}\langle f(x), A_{1,*,i} \circ A_{1,*,j} \rangle
        \end{align*}
    \end{itemize}
\end{lemma}
\begin{proof}
    {\bf Proof of Part 1.}
    \begin{align*}
        \frac{\d^2 \alpha(x)^{-1}}{\d x_i^2}
        = & ~ \frac{\d}{\d x_i}(\frac{\d \alpha(x)^{-1}}{\d x_i})\\
        = & ~ \frac{\d}{\d x_i}(-\alpha(x)^{-1} \cdot \langle f(x), A_{1,*, i} \rangle)\\
        = & ~ -\frac{\d \alpha(x)^{-1}}{\d x_i} \cdot \langle f(x), A_{1,*, i} \rangle - \alpha(x)^{-1} \frac{\d \langle f(x), A_{1,*,i} \rangle}{\d x_i}\\
        = & ~ \alpha(x)^{-1} \cdot \langle f(x), A_{1,*, i} \rangle^2 - \alpha(x)^{-1}(-\langle f(x), A_{1,*,i} \rangle^2 + \langle f(x), A_{1,*,i} \circ A_{1,*,i} \rangle)\\
        = & ~ 2\alpha(x)^{-1} \cdot \langle f(x), A_{1,*, i} \rangle^2 - \alpha(x)^{-1}\langle f(x), A_{1,*,i} \circ A_{1,*,i} \rangle
    \end{align*}
    where the 1st step follows from the differential rules, the 2nd step follows from the result of Part 3 in Lemma~\ref{lem:gradient}, the 3rd step follows from the differential product rule, the 4th step follows from the results of Part 3 and Part 5 in Lemma~\ref{lem:gradient}, and the last step follows from simple algebra.
    
    {\bf Proof of Part 2.}
    \begin{align*}
        & ~ \frac{\d^2 \alpha(x)^{-1}}{\d x_i \d x_j} \\
        = & ~ \frac{\d}{\d x_i}(\frac{\d \alpha(x)^{-1}}{\d x_j})\\
        = & ~ \frac{\d}{\d x_i}(-\alpha(x)^{-1} \cdot \langle f(x), A_{1,*,j} \rangle)\\
        = & ~ -\frac{\d \alpha(x)^{-1}}{\d x_i} \cdot \langle f(x), A_{1,*,j} \rangle - \alpha(x)^{-1} \frac{\d \langle f(x), A_{1,*,j} \rangle}{\d x_i}\\
        = & ~ \alpha(x)^{-1} \cdot \langle f(x), A_{1,*,i} \rangle \langle f(x), A_{1,*,j} \rangle - \alpha(x)^{-1}(-\langle f(x), A_{1,*,j} \rangle \cdot \langle f(x), A_{1,*,i} \rangle + \langle f(x), A_{1,*,i} \circ A_{1,*,j} \rangle)\\
        = & ~ 2\alpha(x)^{-1} \cdot \langle f(x), A_{1,*,i} \rangle \langle f(x), A_{1,*,j} \rangle - \alpha(x)^{-1}\langle f(x), A_{1,*,i} \circ A_{1,*,j} \rangle
    \end{align*}
    where the 1st step follows from the differential rules, the 2nd step follows from the result of Part 3 in Lemma~\ref{lem:gradient}, the 3rd step follows from the differential product rule, the 4th step follows from the results of Part 3 and Part 6 in Lemma~\ref{lem:gradient}, and the last step follows from simple algebra.
\end{proof}

\subsection{Hessian of \texorpdfstring{$f(x)$}{}}
\label{sub:app_hessian:f}

In this section, we study the Hessian of $f(x)$.

\begin{lemma}\label{lem:hessian_f}
If the following conditions hold
    \begin{itemize}
        \item Let $f(x)$ be defined as Definition~\ref{def:f}. 
    \end{itemize}
Then for each $i\in [d]$ and $j \in [d]$, we have
\begin{itemize}
    \item Part 1.
    \begin{align*}
        \frac{\d^2 f(x)}{\d x_i^2}
        = & ~ 2\langle f(x), A_{1,*,i} \rangle^2 f(x) - \langle f(x), A_{1,*,i} \circ A_{1,*,i} \rangle f(x) - 2\langle f(x), A_{1,*,i} \rangle f(x) \circ A_{1,*,i}\\
        + & ~ A_{1,*,i} \circ f(x) \circ A_{1,*,i}
    \end{align*}
    \item Part 2.
    \begin{align*}
        \frac{\d^2 f(x)}{\d x_i \d x_j} 
        = & ~ 2\langle f(x), A_{1,*,i} \rangle \langle f(x), A_{1,*,j} \rangle f(x) - \langle f(x), A_{1,*,i} \circ A_{1,*,j} \rangle f(x) - \langle f(x), A_{1,*,j} \rangle f(x) \circ A_{1,*,i}\\
        - & ~ \langle f(x), A_{1,*,i} \rangle f(x) \circ A_{1,*,j} + A_{1,*,i} \circ f(x) \circ A_{1,*,j}
    \end{align*}
    \item Part 3.
    \begin{align*}
        \frac{\d p_i(x)}{ \d x_i} = \frac{\d^2 f(x)} {\d x_i^2}
    \end{align*}
    \item Part 4.
    \begin{align*}
        \frac{\d p_i(x) }{\d x_j} = \frac{\d^2 f(x) }{ \d x_i \d x_j }
    \end{align*}
\end{itemize}
\end{lemma}

\begin{proof}
    {\bf Proof of Part 1.}
    \begin{align*}
        \frac{\d^2 f(x)}{\d x_i^2}
        = & ~ \frac{\d}{\d x_i} (\frac{\d f(x)}{\d x_i})\\
        = & ~ \frac{\d}{\d x_i} (-\langle f(x), A_{1,*, i} \rangle \cdot f(x) + f(x) \circ A_{1,*,i})\\
        = & ~ - \frac{\d \langle f(x), A_{1,*, i} \rangle}{\d x_i} f(x) - \langle f(x), A_{1,*, i} \rangle \frac{\d f(x)}{\d x_i} + A_{1,*,i} \circ \frac{\d f(x)}{\d x_i}\\
        = & ~ \langle f(x), A_{1,*,i} \rangle^2 f(x) - \langle f(x), A_{1,*,i} \circ A_{1,*,i} \rangle f(x) + \langle f(x), A_{1,*, i} \rangle^2 f(x) - \langle f(x), A_{1,*, i} \rangle f(x) \circ A_{1,*,i}\\ - & ~ \langle f(x), A_{1,*,i} \rangle f(x) \circ A_{1,*,i} + A_{1,*,i} \circ f(x) \circ A_{1,*,i}\\
        = & ~ 2\langle f(x), A_{1,*,i} \rangle^2 f(x) - \langle f(x), A_{1,*,i} \circ A_{1,*,i} \rangle f(x) - 2\langle f(x), A_{1,*,i} \rangle f(x) \circ A_{1,*,i} + A_{1,*,i} \circ f(x) \circ A_{1,*,i}
    \end{align*}
    where the 1st step follows from the differential rules, the 2nd step follows from the result of Part 4 in Lemma~\ref{lem:gradient}, the 3rd step follows from the differential product rule, the 4th step follows from the results of Part 4 and Part 5 in Lemma~\ref{lem:gradient}, and the last step follows from simple algebra.

    {\bf Proof of Part 2.}
    \begin{align*}
        \frac{\d^2 f(x)}{\d x_i \d x_j}
        = & ~ \frac{\d}{\d x_i} (\frac{\d f(x)}{\d x_j})\\
        = & ~ \frac{\d}{\d x_i} (-\langle f(x), A_{1,*,j} \rangle \cdot f(x) + f(x) \circ A_{1,*,j})\\
        = & ~ - \frac{\d \langle f(x), A_{1,*,j} \rangle}{\d x_i} f(x) - \langle f(x), A_{1,*,j} \rangle \frac{\d f(x)}{\d x_i} + A_{1,*,j} \circ \frac{\d f(x)}{\d x_i}\\
        = & ~ \langle f(x), A_{1,*,i} \rangle \langle f(x), A_{1,*,j} \rangle f(x) - \langle f(x), A_{1,*,i} \circ A_{1,*,j} \rangle f(x) + \langle f(x), A_{1,*,j} \rangle \langle f(x), A_{1,*,i} \rangle f(x)\\
        - & ~ \langle f(x), A_{1,*,j} \rangle f(x) \circ A_{1,*,i} - \langle f(x), A_{1,*,i} \rangle f(x) \circ A_{1,*,j} + A_{1,*,i} \circ f(x) \circ A_{1,*,j}\\
        = & ~ 2\langle f(x), A_{1,*,i} \rangle \langle f(x), A_{1,*,j} \rangle f(x) - \langle f(x), A_{1,*,i} \circ A_{1,*,j} \rangle f(x) - \langle f(x), A_{1,*,j} \rangle f(x) \circ A_{1,*,i}\\
        - & ~ \langle f(x), A_{1,*,i} \rangle f(x) \circ A_{1,*,j} + A_{1,*,i} \circ f(x) \circ A_{1,*,j}
    \end{align*}
    where the 1st step follows from the differential rules, the 2nd step follows from the result of Part 4 in Lemma~\ref{lem:gradient}, the 3rd step follows from the differential product rule, the 4th step follows from the results of Part 4 and Part 6 in Lemma~\ref{lem:gradient}, and the last step follows from simple algebra.

    {\bf Proof of Part 3}
    \begin{align*}
        \frac{\d p(x)_i}{\d x_i}
        = & ~ \frac{\d}{\d x_i} \frac{\d f(x)}{\d x_i} \\
        = & ~ \frac{\d^2 f(x)}{\d x_i^2}
    \end{align*}
    where the first step follows from the expansion of hessian,
    the second step follows from {\bf Part 4} of Lemma~\ref{lem:gradient}.

    {\bf Proof of Part 4}
    \begin{align*}
        \frac{\d p(x)_i}{\d x_j}
        = & ~ \frac{\d}{\d x_j} \frac{\d f(x)}{\d x_i} \\
        = & ~ \frac{\d^2 f(x)}{\d x_i \d x_j}
    \end{align*}
    where the first step follows from the expansion of hessian,
    the second step follows from {\bf Part 4} of Lemma~\ref{lem:gradient}.
\end{proof}

\subsection{Hessian of \texorpdfstring{$L(x)$}{}}
\label{sub:app_hessian:l}

In this section, we study the Hessian of $L(x)$.

\begin{lemma}\label{lem:hessian_L_informal}
Let $Q_2(x)$ and $q_2(x)$ be defined as in Definition~\ref{def:Q_q}. We further define
    \begin{itemize}
        \item $B_1(x) \in \R^{n \times n}$ such that
        \begin{align*}
            A^\top_{1,*,i}B_1(x)A_{1,*,j}
            = & ~ ( Q_2(x) \cdot (- \langle f(x), A_{1,*,i} \rangle \cdot f(x) + f(x) \circ A_{1,*,i}))^\top\\
            \cdot & ~ ( Q_2(x) \cdot (- \langle f(x), A_{1,*,j} \rangle \cdot f(x) + f(x) \circ A_{1,*,j}))
        \end{align*}
        \item $B_2(x) \in \R^{n \times n}$ such that
        \begin{align*}
            A^\top_{1,*,i}B_2(x)A_{1,*,j}
            = & ~ q_2(x)^\top \cdot (2\langle f(x), A_{1,*,i} \rangle \langle f(x), A_{1,*,j} \rangle f(x) - \langle f(x), A_{1,*,i} \circ A_{1,*,j} \rangle f(x)\\
            - & ~ \langle f(x), A_{1,*,j} \rangle f(x) \circ A_{1,*,i} - \langle f(x), A_{1,*,i} \rangle f(x) \circ A_{1,*,j} + A_{1,*,i} \circ f(x) \circ A_{1,*,j})
        \end{align*}
    \end{itemize}
    Then we have
    \begin{itemize}
        \item Part 1.
        \begin{align*}
            \frac{\d^2 L}{\d x_i^2 } 
            = & ~ \|\underbrace{Q_2(x)}_{m \times n} \cdot \underbrace{p(x)_i}_{n \times 1}\|_2^2 + \langle \underbrace{c(x)}_{m \times 1}, \underbrace{\diag( \diag( h''(A_2 f(x)) )  \cdot  A_2  \cdot p(x)_i ) \cdot A_2}_{m \times n} \cdot \underbrace{p(x)_i}_{n \times 1} \rangle \\
            + & ~ \langle \underbrace{c(x)}_{m \times 1}, 2 \underbrace{Q_2(x)}_{m \times n} \cdot \underbrace{f(x)}_{n \times 1} \rangle \cdot \langle \underbrace{f(x)}_{n \times 1}, \underbrace{A_{1,*,i}}_{n \times 1} \rangle^2 - \langle \underbrace{c(x)}_{m \times 1}, \underbrace{Q_2(x)}_{m \times n} \cdot \underbrace{f(x)}_{n \times 1} \rangle \cdot \langle \underbrace{f(x)}_{n \times 1}, \underbrace{A_{1,*,i}}_{n \times 1} \circ \underbrace{A_{1,*,i}}_{n \times 1} \rangle \notag\\
            - & ~ \langle \underbrace{c(x)}_{m \times 1}, 2\underbrace{Q_2(x)}_{m \times n} \cdot (\underbrace{f(x)}_{n \times 1} \circ \underbrace{A_{1,*,i}}_{n \times 1}) \rangle \cdot \langle \underbrace{f(x)}_{n \times 1}, \underbrace{A_{1,*,i}}_{n \times 1} \rangle + \langle \underbrace{c(x)}_{m \times 1}, \underbrace{Q_2(x)}_{m \times n} \cdot (\underbrace{A_{1,*,i}}_{n \times 1} \circ \underbrace{f(x)}_{n \times 1} \circ \underbrace{A_{1,*,i}}_{n \times 1}) \rangle
        \end{align*}
        \item Part 2.
        \begin{align*}
            \frac{\d^2 L}{\d x_i \d x_j } 
            = & ~ \langle  \underbrace{ Q_2(x)}_{m \times n} \cdot \underbrace{ p(x)_j}_{n \times 1}, \underbrace{ Q_2(x)}_{m \times n} \cdot \underbrace{ p(x)_i}_{n \times 1} \rangle  + \langle \underbrace{c(x)}_{m \times 1}, \underbrace{\diag( \diag( h''(A_2 f(x)) )  \cdot  A_2  \cdot p(x)_j ) \cdot A_2}_{m \times n} \cdot \underbrace{p(x)_i}_{n \times 1} \rangle \\
            + & ~ \langle \underbrace{c(x)}_{m \times 1}, 2 \underbrace{Q_2(x)}_{m \times n} \cdot \underbrace{f(x)}_{n \times 1} \rangle \cdot \langle \underbrace{f(x)}_{n \times 1}, \underbrace{A_{1,*,i}}_{n \times 1} \rangle \langle \underbrace{f(x)}_{n \times 1}, \underbrace{A_{1,*,j}}_{n \times 1} \rangle - \langle \underbrace{c(x)}_{m \times 1}, \underbrace{Q_2(x)}_{m \times n} \cdot \underbrace{f(x)}_{n \times 1} \rangle \cdot \langle \underbrace{f(x)}_{n \times 1}, \underbrace{A_{1,*,i}}_{n \times 1} \circ \underbrace{A_{1,*,j}}_{n \times 1} \rangle \notag\\
            - & ~ \langle \underbrace{c(x)}_{m \times 1}, 2\underbrace{Q_2(x)}_{m \times n} \cdot (\underbrace{f(x)}_{n \times 1} \circ \underbrace{A_{1,*,j}}_{n \times 1}) \rangle \cdot \langle \underbrace{f(x)}_{n \times 1}, \underbrace{A_{1,*,i}}_{n \times 1} \rangle + \langle \underbrace{c(x)}_{m \times 1}, \underbrace{Q_2(x)}_{m \times n} \cdot (\underbrace{A_{1,*,i}}_{n \times 1} \circ \underbrace{f(x)}_{n \times 1} \circ \underbrace{A_{1,*,j}}_{n \times 1}) \rangle
        \end{align*}
    \end{itemize}
\end{lemma}

\begin{proof}
    {\bf Proof of Part 1.}

    We can show
    \begin{align}\label{eq:hessian_of_L_xi}
        \frac{\d^2 L}{\d x_i^2 }
        = & ~ \frac{\d}{\d x_i}(\frac{\d L}{\d x_i}) \notag\\
        = & ~ \frac{\d}{\d x_i} \langle c(x), Q_2(x) \cdot p(x)_i \rangle \notag\\
        = & ~ \langle \frac{\d }{\d x_i} c(x), Q_2(x) \cdot p(x)_i \rangle \notag\\
        + & ~ \langle  c(x),  \frac{\d }{\d x_i} ( Q_2(x) ) \cdot p(x)_i \rangle \notag\\
        + & ~ \langle  c(x), Q_2(x) \cdot  \frac{\d }{\d x_i} ( p(x)_i ) \rangle,
    \end{align}
    where the 1st step follows from the differential rules, the 2nd step follows from the result of Part 9 in Lemma~\ref{lem:gradient}.

    First, we compute the first term of Eq.~\eqref{eq:hessian_of_L_xi}:
    \begin{align}\label{eq:hessian_of_L_xi:first}
        & ~ \langle \frac{\d }{\d x_i} c(x), Q_2(x) \cdot p(x)_i \rangle \notag\\
        = & ~ \langle  \underbrace{ \diag( h'(A_2 f(x)) ) }_{m \times m} \cdot \underbrace{ A_2 }_{m \times n} \cdot \underbrace{ p(x)_i }_{n \times 1} , \underbrace{ Q_2(x)}_{m \times n} \cdot \underbrace{ p(x)_i}_{n \times 1} \rangle \notag\\
        = & ~ \langle  \underbrace{ Q_2(x)}_{m \times n} \cdot \underbrace{ p(x)_i}_{n \times 1}, \underbrace{ Q_2(x)}_{m \times n} \cdot \underbrace{ p(x)_i}_{n \times 1} \rangle \notag\\
        = & ~ \|Q_2(x) \cdot p(x)_i\|_2^2,
    \end{align}
    where the first step follows from {\bf Part 7} and {\bf Part 8} of Lemma~\ref{lem:gradient}, the second step follows from the definition of $Q_2(x)$ (see Definition~\ref{def:Q_q}), and the last step follows from the definition of $\|\cdot\|_2$.

    Second, we compute the second term of Eq.~\eqref{eq:hessian_of_L_xi}.
    
    Consider $\frac{\d }{\d x_i} ( Q_2(x) )$: we have
    \begin{align*}
        \frac{\d }{\d x_i} ( Q_2(x) )
        = & ~ \frac{\d }{\d x_i} ( \diag( h'(A_2 f(x)) ) A_2 )\notag\\
        = & ~ \frac{\d }{\d x_i} ( \diag( h'(A_2 f(x)) ) ) \cdot A_2\notag\\
        = & ~ \diag( \frac{\d }{\d x_i} ( h'(A_2 f(x)) ) ) \cdot A_2\notag\\
        = & ~ \diag(\underbrace{ \diag( h''(A_2 f(x)) ) }_{m \times m} \cdot \underbrace{ A_2 }_{m \times n} \cdot \underbrace{ p(x)_i }_{n \times 1}) \cdot \underbrace{A_2}_{m \times n},
    \end{align*}
    where the first step follows from the definition of $Q_2(x)$ (see Definition~\ref{def:Q_q}), the second step follows from the fact that $A_2$ is a constant matrix with respect to $x_i$, the third step follows from the definition of $\diag$, and the last step follows from {\bf Part 10} of Lemma~\ref{lem:gradient}.

    Therefore, we have
    \begin{align}\label{eq:hessian_of_L_xi:second}
        \langle c(x), \frac{\d }{\d x_i} ( Q_2(x) ) \cdot p(x)_i \rangle = \langle \underbrace{c(x)}_{m \times 1}, \underbrace{\diag( \diag( h''(A_2 f(x)) )  \cdot  A_2  \cdot p(x)_i ) \cdot A_2}_{m \times n} \cdot \underbrace{p(x)_i}_{n \times 1} \rangle 
    \end{align}

    Third, we compute the third term of Eq.~\eqref{eq:hessian_of_L_xi}:
    \begin{align}\label{eq:hessian_of_L_xi:third}
        & ~ \langle  c(x), Q_2(x) \cdot  \frac{\d }{\d x_i} ( p(x)_i ) \rangle \notag\\
        = & ~ \langle c(x), Q_2(x) \cdot \notag\\
        & ~ (2\langle f(x), A_{1,*,i} \rangle^2 f(x) - \langle f(x), A_{1,*,i} \circ A_{1,*,i} \rangle f(x) - 2\langle f(x), A_{1,*,i} \rangle f(x) \circ A_{1,*,i} + A_{1,*,i} \circ f(x) \circ A_{1,*,i}) \rangle \notag\\
        = & ~ \langle c(x), 2 Q_2(x) \langle f(x), A_{1,*,i} \rangle^2 f(x) \rangle \notag \\
        & ~ - \langle c(x), Q_2(x) \langle f(x), A_{1,*,i} \circ A_{1,*,i} \rangle f(x) \rangle \notag\\
        & ~ - \langle c(x), 2Q_2(x) \langle f(x), A_{1,*,i} \rangle f(x) \circ A_{1,*,i} \rangle \notag \\
        & ~ + \langle c(x), Q_2(x) A_{1,*,i} \circ f(x) \circ A_{1,*,i} \rangle \notag\\
        = & ~ \langle c(x), 2 Q_2(x) \cdot f(x) \rangle \cdot \langle f(x), A_{1,*,i} \rangle^2 \notag\\
        & ~ - \langle c(x), Q_2(x) \cdot f(x) \rangle \cdot \langle f(x), A_{1,*,i} \circ A_{1,*,i} \rangle \notag\\
        & ~ - \langle c(x), 2Q_2(x) \cdot (f(x) \circ A_{1,*,i}) \rangle \cdot \langle f(x), A_{1,*,i} \rangle \notag\\
        & ~ + \langle c(x), Q_2(x) \cdot (A_{1,*,i} \circ f(x) \circ A_{1,*,i}) \rangle,
    \end{align}
    where the first step follows from {\bf Part 1} and {\bf Part 3} of Lemma~\ref{lem:hessian_f}, the second step follows from Fact~\ref{fac:basic_algebra}, and the third step follows from Fact~\ref{fac:basic_algebra}.

    Combining Eq.~\eqref{eq:hessian_of_L_xi}, Eq.~\eqref{eq:hessian_of_L_xi:first}, Eq.~\eqref{eq:hessian_of_L_xi:second}, Eq.~\eqref{eq:hessian_of_L_xi:third}, we have
    \begin{align*}
        \frac{\d^2 L}{\d x_i^2 } 
        = & ~ \|\underbrace{Q_2(x)}_{m \times n} \cdot \underbrace{p(x)_i}_{n \times 1}\|_2^2 + \langle \underbrace{c(x)}_{m \times 1}, \underbrace{\diag( \diag( h''(A_2 f(x)) )  \cdot  A_2  \cdot p(x)_i ) \cdot A_2}_{m \times n} \cdot \underbrace{p(x)_i}_{n \times 1} \rangle \\
        + & ~ \langle \underbrace{c(x)}_{m \times 1}, 2 \underbrace{Q_2(x)}_{m \times n} \cdot \underbrace{f(x)}_{n \times 1} \rangle \cdot \langle \underbrace{f(x)}_{n \times 1}, \underbrace{A_{1,*,i}}_{n \times 1} \rangle^2 - \langle \underbrace{c(x)}_{m \times 1}, \underbrace{Q_2(x)}_{m \times n} \cdot \underbrace{f(x)}_{n \times 1} \rangle \cdot \langle \underbrace{f(x)}_{n \times 1}, \underbrace{A_{1,*,i}}_{n \times 1} \circ \underbrace{A_{1,*,i}}_{n \times 1} \rangle \notag\\
        - & ~ \langle \underbrace{c(x)}_{m \times 1}, 2\underbrace{Q_2(x)}_{m \times n} \cdot (\underbrace{f(x)}_{n \times 1} \circ \underbrace{A_{1,*,i}}_{n \times 1}) \rangle \cdot \langle \underbrace{f(x)}_{n \times 1}, \underbrace{A_{1,*,i}}_{n \times 1} \rangle + \langle \underbrace{c(x)}_{m \times 1}, \underbrace{Q_2(x)}_{m \times n} \cdot (\underbrace{A_{1,*,i}}_{n \times 1} \circ \underbrace{f(x)}_{n \times 1} \circ \underbrace{A_{1,*,i}}_{n \times 1}) \rangle
    \end{align*}

    {\bf Proof of Part 2.}
    We can show
    \begin{align}\label{eq:hessian_of_L_xi_2}
        \frac{\d^2 L}{\d x_i \d x_j }
        = & ~ \frac{\d}{\d x_j}(\frac{\d L}{\d x_i}) \notag\\
        = & ~ \frac{\d}{\d x_j} \langle c(x), Q_2(x) \cdot p(x)_i \rangle \notag\\
        = & ~ \langle \frac{\d }{\d x_j} c(x), Q_2(x) \cdot p(x)_i \rangle \notag\\
        + & ~ \langle  c(x),  \frac{\d }{\d x_j} ( Q_2(x) ) \cdot p(x)_i \rangle \notag\\
        + & ~ \langle  c(x), Q_2(x) \cdot  \frac{\d }{\d x_j} ( p(x)_i ) \rangle,
    \end{align}
    where the 1st step follows from the differential rules, the 2nd step follows from the result of Part 9 in Lemma~\ref{lem:gradient}.

    First, we compute the first term of Eq.~\eqref{eq:hessian_of_L_xi_2}:
    \begin{align}\label{eq:hessian_of_L_xi_2:first}
        & ~ \langle \frac{\d }{\d x_j} c(x), Q_2(x) \cdot p(x)_i \rangle \notag\\
        = & ~ \langle  \underbrace{ \diag( h'(A_2 f(x)) ) }_{m \times m} \cdot \underbrace{ A_2 }_{m \times n} \cdot \underbrace{ p(x)_j }_{n \times 1} , \underbrace{ Q_2(x)}_{m \times n} \cdot \underbrace{ p(x)_i}_{n \times 1} \rangle \notag\\
        = & ~ \langle  \underbrace{ Q_2(x)}_{m \times n} \cdot \underbrace{ p(x)_j}_{n \times 1}, \underbrace{ Q_2(x)}_{m \times n} \cdot \underbrace{ p(x)_i}_{n \times 1} \rangle 
    \end{align}
    where the first step follows from {\bf Part 7} and {\bf Part 8} of Lemma~\ref{lem:gradient}, the second step follows from the definition of $Q_2(x)$ (see Definition~\ref{def:Q_q}).
    
    Second, we compute the second term of Eq.~\eqref{eq:hessian_of_L_xi_2}.
    
    Consider $\frac{\d }{\d x_j} ( Q_2(x) )$: we have
    \begin{align*}
        \frac{\d }{\d x_j} ( Q_2(x) )
        = & ~ \frac{\d }{\d x_j} ( \diag( h'(A_2 f(x)) ) A_2 )\notag\\
        = & ~ \frac{\d }{\d x_j} ( \diag( h'(A_2 f(x)) ) ) \cdot A_2\notag\\
        = & ~ \diag( \frac{\d }{\d x_j} ( h'(A_2 f(x)) ) ) \cdot A_2\notag\\
        = & ~ \diag(\underbrace{ \diag( h''(A_2 f(x)) ) }_{m \times m} \cdot \underbrace{ A_2 }_{m \times n} \cdot \underbrace{ p(x)_j }_{n \times 1}) \cdot \underbrace{A_2}_{m \times n},
    \end{align*}
    where the first step follows from the definition of $Q_2(x)$ (see Definition~\ref{def:Q_q}), the second step follows from the fact that $A_2$ is a constant matrix with respect to $x_i$, the third step follows from the definition of $\diag$, and the last step follows from {\bf Part 10} of Lemma~\ref{lem:gradient}.

    Therefore, we have
    \begin{align}\label{eq:hessian_of_L_xi_2:second}
        \langle c(x), \frac{\d }{\d x_j} ( Q_2(x) ) \cdot p(x)_i \rangle = \langle \underbrace{c(x)}_{m \times 1}, \underbrace{\diag( \diag( h''(A_2 f(x)) )  \cdot  A_2  \cdot p(x)_j ) \cdot A_2}_{m \times n} \cdot \underbrace{p(x)_i}_{n \times 1} \rangle 
    \end{align}

    Third, we compute the third term of Eq.~\eqref{eq:hessian_of_L_xi_2}:
    \begin{align}\label{eq:hessian_of_L_xi_2:third}
        & ~ \langle  c(x), Q_2(x) \cdot  \frac{\d }{\d x_j} ( p(x)_i ) \rangle \notag\\
        = & ~ \langle c(x), Q_2(x) \cdot \notag\\
        & ~ (2\langle f(x), A_{1,*,i} \rangle \langle f(x), A_{1,*,j} \rangle f(x) - \langle f(x), A_{1,*,i} \circ A_{1,*,j} \rangle f(x) - \langle f(x), A_{1,*,j} \rangle f(x) \circ A_{1,*,i}\notag \\
        - & ~ \langle f(x), A_{1,*,i} \rangle f(x) \circ A_{1,*,j} + A_{1,*,i} \circ f(x) \circ A_{1,*,j}) \rangle \notag\\
        = & ~ \langle c(x), 2 Q_2(x) \langle f(x), A_{1,*,i} \rangle \langle f(x), A_{1,*,j} \rangle f(x) \rangle \notag \\
        & ~ - \langle c(x), Q_2(x) \langle f(x), A_{1,*,i} \circ A_{1,*,j} \rangle f(x) \rangle \notag\\
        & ~ - \langle c(x), 2Q_2(x) \langle f(x), A_{1,*,i} \rangle f(x) \circ A_{1,*,j} \rangle \notag \\
        & ~ + \langle c(x), Q_2(x) A_{1,*,i} \circ f(x) \circ A_{1,*,j} \rangle \notag\\
        = & ~ \langle c(x), 2 Q_2(x) \cdot f(x) \rangle \cdot \langle f(x), A_{1,*,i} \rangle \langle f(x), A_{1,*,j} \rangle \notag\\
        & ~ - \langle c(x), Q_2(x) \cdot f(x) \rangle \cdot \langle f(x), A_{1,*,i} \circ A_{1,*,i} \rangle \notag\\
        & ~ - \langle c(x), 2Q_2(x) \cdot (f(x) \circ A_{1,*,j}) \rangle \cdot \langle f(x), A_{1,*,i} \rangle \notag\\
        & ~ + \langle c(x), Q_2(x) \cdot (A_{1,*,i} \circ f(x) \circ A_{1,*,j}) \rangle,
    \end{align}
    where the first step follows from {\bf Part 1} and {\bf Part 3} of Lemma~\ref{lem:hessian_f}, the second step follows from Fact~\ref{fac:basic_algebra}, and the third step follows from Fact~\ref{fac:basic_algebra}.

    Combining Eq.~\eqref{eq:hessian_of_L_xi_2}, Eq.~\eqref{eq:hessian_of_L_xi_2:first}, Eq.~\eqref{eq:hessian_of_L_xi_2:second}, Eq.~\eqref{eq:hessian_of_L_xi_2:third}, we have
    \begin{align*}
        \frac{\d^2 L}{\d x_i \d x_j } 
        = & ~ \langle  \underbrace{ Q_2(x)}_{m \times n} \cdot \underbrace{ p(x)_j}_{n \times 1}, \underbrace{ Q_2(x)}_{m \times n} \cdot \underbrace{ p(x)_i}_{n \times 1} \rangle  + \langle \underbrace{c(x)}_{m \times 1}, \underbrace{\diag( \diag( h''(A_2 f(x)) )  \cdot  A_2  \cdot p(x)_j ) \cdot A_2}_{m \times n} \cdot \underbrace{p(x)_i}_{n \times 1} \rangle \\
        + & ~ \langle \underbrace{c(x)}_{m \times 1}, 2 \underbrace{Q_2(x)}_{m \times n} \cdot \underbrace{f(x)}_{n \times 1} \rangle \cdot \langle \underbrace{f(x)}_{n \times 1}, \underbrace{A_{1,*,i}}_{n \times 1} \rangle \langle \underbrace{f(x)}_{n \times 1}, \underbrace{A_{1,*,j}}_{n \times 1} \rangle - \langle \underbrace{c(x)}_{m \times 1}, \underbrace{Q_2(x)}_{m \times n} \cdot \underbrace{f(x)}_{n \times 1} \rangle \cdot \langle \underbrace{f(x)}_{n \times 1}, \underbrace{A_{1,*,i}}_{n \times 1} \circ \underbrace{A_{1,*,j}}_{n \times 1} \rangle \notag\\
        - & ~ \langle \underbrace{c(x)}_{m \times 1}, 2\underbrace{Q_2(x)}_{m \times n} \cdot (\underbrace{f(x)}_{n \times 1} \circ \underbrace{A_{1,*,j}}_{n \times 1}) \rangle \cdot \langle \underbrace{f(x)}_{n \times 1}, \underbrace{A_{1,*,i}}_{n \times 1} \rangle + \langle \underbrace{c(x)}_{m \times 1}, \underbrace{Q_2(x)}_{m \times n} \cdot (\underbrace{A_{1,*,i}}_{n \times 1} \circ \underbrace{f(x)}_{n \times 1} \circ \underbrace{A_{1,*,j}}_{n \times 1}) \rangle
    \end{align*} 
\end{proof}

\subsection{Re-oragnizing \texorpdfstring{$B(x)$}{}}
\label{sub:app_hessian:B}

In this section, we reorganize $B(x)$.

\begin{lemma}\label{lem:reorganizing}
    Let $\frac{\d^2 L(x)}{\d x_i \d x_j}$ be computed in Lemma~\ref{lem:hessian_L_informal}, then, we have
    \begin{align*}
        \frac{\d^2 L}{\d x^2}
        = A^\top B(x) A
    \end{align*}
    where
    \begin{align*}
        B(x)
        = & ~ \diag(f(x))^\top Q_2(x)^\top  Q_2(x) \diag(f(x)) \\
        & ~ + \diag(f(x))^\top Q_2(x)^\top Q_2(x) f(x) f(x)^\top \\
        & ~ + f(x) f(x)^\top Q_2(x)^\top  Q_2(x) \diag(f(x)) \\
        & ~ + f(x)^\top Q_2(x)^\top Q_2(x) f(x) f(x)^\top \\
        & ~ + 2 f(x) c(x)^\top Q_2(x) f(x) f(x)^\top \\
        & ~ + 2 f(x) c(x)^\top Q_2(x) \diag(f(x)) \\
        & ~ + \diag(Q_2(x)^\top c(x)) \diag(f(x)) \\
        & ~ +  \diag(f(x)) A_2 \diag( h''(A_2 f(x)) ) \diag(c(x)) \diag(f(x)) \\
        & ~ + \diag(f(x)) A_2 \diag( h''(A_2 f(x)) ) \diag(c(x)) f(x) f(x)^\top \\
        & ~ + f(x) f(x) A_2 \diag( h''(A_2 f(x)) ) \diag(c(x)) \diag(f(x)) \\
        & ~ + f(x) f(x) A_2 \diag( h''(A_2 f(x)) ) \diag(c(x)) f(x) f(x)^\top \\
        & ~ + \diag(f(x) f(x)^\top Q_2(x)^\top c(x))
    \end{align*}
\end{lemma}
\begin{proof}
    For the first term, we have
    \begin{align*}
        & ~ \langle  Q_2(x) p(x)_j , Q_2(x) p(x)_i \rangle \\
        = & ~ \langle Q_2(x) (f(x) \circ A_{1,*,i} + \langle f(x), A_{1,*,i} \rangle \cdot f(x)), Q_2(x) (f(x) \circ A_{1,*,j} + \langle f(x), A_{1,*,j} \rangle \cdot f(x)) \rangle \\
        = & ~ \langle Q_2(x) f(x) \circ A_{1,*,i} + Q_2(x)\langle f(x), A_{1,*,i} \rangle \cdot f(x), Q_2(x) f(x) \circ A_{1,*,j} + Q_2(x)\langle f(x), A_{1,*,j} \rangle \cdot f(x) \rangle \\
        = & ~ (Q_2(x) f(x) \circ A_{1,*,i})^\top Q_2(x) f(x) \circ A_{1,*,j} \\
        & ~ + (Q_2(x) f(x) \circ A_{1,*,i})^\top  Q_2(x)\langle f(x), A_{1,*,j} \rangle \cdot f(x) \\
        & ~ + (Q_2(x)\langle f(x), A_{1,*,i} \rangle \cdot f(x))^\top Q_2(x) f(x) \circ A_{1,*,j} \\
        & ~ + (Q_2(x)\langle f(x), A_{1,*,i} \rangle \cdot f(x))^\top Q_2(x)\langle f(x), A_{1,*,j} \rangle \cdot f(x) \\
        = & ~ A_{1,*,i}^\top \diag(f(x))^\top Q_2(x)^\top  Q_2(x) \diag(f(x)) A_{1,*,j} \\
        & ~ + A_{1,*,i}^\top \diag(f(x))^\top Q_2(x)^\top Q_2(x) f(x) f(x)^\top A_{1,*,j} \\
        & ~ + A_{1,*,i}^\top f(x) f(x)^\top Q_2(x)^\top  Q_2(x) \diag(f(x)) A_{1,*,j} \\
        & ~ + A_{1,*,i}^\top f(x) f(x)^\top Q_2(x)^\top Q_2(x) f(x) f(x)^\top A_{1,*,j} 
    \end{align*}

    For the second term, we first define
    \begin{align*}
        F : & = \diag( h''(A_2 f(x)) ) \\
        M_i : & = A_2  \cdot p(x)_i
    \end{align*}
    then the second term can be reformed as
    \begin{align*}
        \langle c(x), \diag(F M_j) M_i \rangle 
        = & ~ \langle c(x), (F M_j) \circ M_i \rangle \\
        = & ~ c(x)^\top (F M_j) \circ M_i \\
        = & ~ (F M_j)^\top c(x) \circ M_i \\
        = & ~ M_j^\top F \diag(c(x)) M_i 
    \end{align*}
    By substitute $M_i$ with $A_2  \cdot p(x)_i$, we have
    \begin{align*}
        M_j^\top F \diag(c(x)) M_i 
        = & ~ (A_2 p(x)_j)^\top F \diag(c(x)) A_2 p(x)_i \\
        = & ~ p(x)_j^\top A_2 F \diag(c(x)) A_2 p(x)_i 
    \end{align*}
    Let 
    \begin{align*}
        D := & ~ A_2 F \diag(c(x)) \\
        = & ~ A_2 \diag( h''(A_2 f(x)) ) \diag(c(x))
    \end{align*}
    then by the definition of $p(x)_i$, we have
    \begin{align*}
        p(x)_j^\top A_2 F \diag(c(x)) A_2 p(x)_i
        = & ~ (f(x) \circ A_{1,*,j} - \langle f(x), A_{1,*,j} \rangle \cdot f(x))^\top D (f(x) \circ A_{1,*,i} - \langle f(x), A_{1,*,i} \rangle \cdot f(x)) \\
        = & ~ (f(x)^\top \circ A_{1,*,j}^\top - \langle f(x), A_{1,*,j} \rangle \cdot f(x)^\top) D (f(x) \circ A_{1,*,i} - \langle f(x), A_{1,*,i} \rangle \cdot f(x)) \\
        = & ~ f(x)^\top \circ A_{1,*,j}^\top D f(x) \circ A_{1,*,i} \\
        & ~ - f(x)^\top \circ A_{1,*,j}^\top D \langle f(x), A_{1,*,i} \rangle \cdot f(x) \\
        & ~ - \langle f(x), A_{1,*,j} \rangle \cdot f(x) D f(x) \circ A_{1,*,i} \\
        & ~ + \langle f(x), A_{1,*,j} \rangle \cdot f(x) D \langle f(x), A_{1,*,i} \rangle \cdot f(x) \\
        = & ~ A_{1,*,j}^\top \diag(f(x)) D \diag(f(x))  A_{1,*,i} \\
        & ~ - A_{1,*,j}^\top \diag(f(x)) D f(x) f(x)^\top A_{1,*,i} \\
        & ~ - A_{1,*,j}^\top f(x) f(x) D \diag(f(x))  A_{1,*,i} \\
        & ~ + A_{1,*,j}^\top f(x) f(x) D f(x) f(x)^\top A_{1,*,i}
    \end{align*}

    For the third term, we have
    \begin{align*}
        \langle c(x) , 2 Q_2(x) \cdot f(x) \rangle \cdot \langle f(x) , A_{1,*,i} \rangle \langle f(x) , A_{1,*,j} \rangle
        = 2 A_{1,*,i}^\top f(x) c(x)^\top Q_2(x) f(x) f(x)^\top A_{1,*,j}
    \end{align*}

    For the fourth term, we have
    \begin{align*}
        \langle c(x) , Q_2(x) \cdot f(x) \rangle \cdot \langle f(x) , A_{1,*,i} \circ A_{1,*,j} \rangle
        = & ~ c(x)^\top Q_2(x) f(x) f(x)^\top A_{1,*,i} \circ A_{1,*,j} \\
        = & ~ A_{1,*,i}^\top \diag(f(x) f(x)^\top Q_2(x)^\top c(x)) A_{1,*,j}
    \end{align*}

    For the fifth term, we have
    \begin{align*}
        \langle c(x) , 2 Q_2(x) \cdot (f(x) \circ A_{1,*,j}) \rangle \cdot \langle f(x) , A_{1,*,i} \rangle
        = 2 A_{1,*,i}^\top f(x) c(x)^\top Q_2(x) \diag(f(x)) A_{1,*,j}
    \end{align*}

    For the sixth term, we have
    \begin{align*}
        \langle c(x) , Q_2(x) \cdot (A_{1,*,i} \circ f(x) \circ A_{1,*,j}) \rangle
        =  A_{1,*,i}^\top \diag(Q_2(x)^\top c(x)) \diag(f(x)) A_{1,*,j}
    \end{align*}

    Thus, we have
    \begin{align*}
        \frac{\d^2 L}{\d x^2}
        = A^\top B(x) A
    \end{align*}
    where
    \begin{align*}
        B(x)
        = & ~ \diag(f(x))^\top Q_2(x)^\top  Q_2(x) \diag(f(x))\\
        & ~ + \diag(f(x))^\top Q_2(x)^\top Q_2(x) f(x) f(x)^\top \\
        & ~ + f(x) f(x)^\top Q_2(x)^\top  Q_2(x) \diag(f(x)) \\
        & ~ + f(x)^\top Q_2(x)^\top Q_2(x) f(x) f(x)^\top \\
        & ~ + 2 f(x) c(x)^\top Q_2(x) f(x) f(x)^\top \\
        & ~ + 2 f(x) c(x)^\top Q_2(x) \diag(f(x)) \\
        & ~ + \diag(Q_2(x)^\top c(x)) \diag(f(x)) \\
        & ~ +  \diag(f(x)) A_2 \diag( h''(A_2 f(x)) ) \diag(c(x)) \diag(f(x)) \\
        & ~ + \diag(f(x)) A_2 \diag( h''(A_2 f(x)) ) \diag(c(x)) f(x) f(x)^\top \\
        & ~ + f(x) f(x) A_2 \diag( h''(A_2 f(x)) ) \diag(c(x)) \diag(f(x)) \\
        & ~ + f(x) f(x) A_2 \diag( h''(A_2 f(x)) ) \diag(c(x)) f(x) f(x)^\top \\
        & ~ + \diag(f(x) f(x)^\top Q_2(x)^\top c(x))
    \end{align*}
\end{proof}

\section{Hessian is Positive definite}

\label{sec:psd_app}

In Section~\ref{sub:psd_app:final}, we show the final PSD lower bounds we have for $B(x)$. In Section~\ref{sub:psd_app:tools}, we present a list of tools for bounding different parts of the matrix $B(x)$. In Section~\ref{sub:psd_app:bound}, we analyze the lower bound on the Hessian of $L(x)$.

\subsection{PSD Lower Bound: final bound}

In this section, we analyze the PSD lower bound for $B(x)$.

\label{sub:psd_app:final}

\begin{lemma}\label{lem:final_bound}
    If the following conditions hold
    \begin{itemize}
        \item Let $B(x) \in \R^{n \times n}$ be defined as Definition~\ref{def:B}.
    \end{itemize}
    Then we have
    \begin{align*}
      -12 R_h L_h R (R + R_h) I_n \preceq  B(x) \preceq 12 R_h L_h R (R + R_h) I_n
    \end{align*}
\end{lemma}
\begin{proof}
    It follows from Lemma~\ref{lem:psd_lower_bound_Bi} that we have
    \begin{align*}
        \max\{\| Q_2(x) \|^2, 2 (L_h + 1)\|Q_2(x)\|, (R + R_h) L_hR , R_h R (R + R_h)\} \leq & ~ 12 R_h L_h R (R + R_h) \\
        \min\{\| Q_2(x) \|^2, 2 (L_h + 1)\|Q_2(x)\|, (R + R_h) L_hR , R_h R (R + R_h)\} \geq & ~ -12 R_h L_h R (R + R_h)
    \end{align*}
    Thus, we have
    \begin{align*}
        -12 R_h L_h R (R + R_h) I_n \preceq  B(x) \preceq 12 R_h L_h R (R + R_h) I_n
    \end{align*}
\end{proof}

\subsection{PSD Lower Bound: A list of Tools}

In this section, we analyze the PSD lower bound for different parts of $B(x)$.

\label{sub:psd_app:tools}

\begin{lemma}\label{lem:psd_lower_bound_Bi}
    If the following conditions hold
    \begin{itemize}
        \item $\| f(x) \|_1 = 1$ (see Definition~\ref{def:f}).
        \item Let $B(x) \in \R^{n \times n}$ be defined as Definition~\ref{def:B}.
        \item Let $f(x) \geq {\bf 0}_n$.
        \item Let $b \geq {\bf 0}_n$.
        \item $B_1= \diag(f(x))^\top Q_2(x)^\top  Q_2(x) \diag(f(x)) $
        \item $B_2 = \diag(f(x))^\top Q_2(x)^\top Q_2(x) f(x) f(x)^\top $
        \item $B_3 = f(x) f(x)^\top Q_2(x)^\top  Q_2(x) \diag(f(x))$
        \item $B_4 = f(x)^\top Q_2(x)^\top Q_2(x) f(x) f(x)^\top$
        \item $B_5 = 2 f(x) c(x)^\top Q_2(x) f(x) f(x)^\top$
        \item $B_6 = 2 f(x) c(x)^\top Q_2(x) \diag(f(x))$ 
        \item $B_7 = \diag(Q_2(x)^\top c(x)) \diag(f(x))$
        \item $B_8 = \diag(f(x)) A_2 \diag( h''(A_2 f(x)) ) \diag(c(x)) \diag(f(x))$
        \item $B_9 = \diag(f(x)) A_2 \diag( h''(A_2 f(x)) ) \diag(c(x)) f(x) f(x)^\top$
        \item $B_{10} = f(x) f(x) A_2 \diag( h''(A_2 f(x)) ) \diag(c(x)) \diag(f(x))$
        \item $B_{11} = f(x) f(x) A_2 \diag( h''(A_2 f(x)) ) \diag(c(x)) f(x) f(x)^\top$
        \item $B_{12} = \diag(f(x) f(x)^\top Q_2(x)^\top c(x))$
    \end{itemize}
    Then we have
    \begin{itemize}
        \item {\bf Part 1. }
        \begin{align*}
             - \| Q_2(x) \|^2 \cdot I_n \preceq B_1 \preceq \| Q_2(x) \|^2 \cdot I_n
        \end{align*}
        \item {\bf Part 2.}
        \begin{align*}
          - \| Q_2(x) \|^2 \cdot I_n \preceq  B_2 \preceq \| Q_2(x) \|^2  \cdot I_n
        \end{align*}
        \item {\bf Part 3.}
        \begin{align*}
            - \|Q_2(x)\|^2 \cdot I_n \preceq B_3 \preceq \|Q_2(x)\|^2 \cdot I_n \cdot I_n
        \end{align*}
        \item {\bf Part 4.}
        \begin{align*}
            - \|Q_2(x)\|^2 \cdot I_n \preceq B_4 \preceq \|Q_2(x)\|^2 \cdot I_n
        \end{align*}
        \item {\bf Part 5.}
        \begin{align*}
            - 2 (L_h + 1)\|Q_2(x)\| \cdot I_n \preceq B_5 \preceq 2 (L_h + 1)\|Q_2(x)\| \cdot I_n
        \end{align*}
        \item {\bf Part 6.}
        \begin{align*}
            - 2 (L_h + 1)\|Q_2(x)\| \cdot I_n \preceq B_5 \preceq 2 (L_h + 1)\|Q_2(x)\| \cdot I_n
        \end{align*}
        \item {\bf Part 7.}
        \begin{align*}
            - 2 (L_h + 1)\|Q_2(x)\| \cdot I_n \preceq B_5 \preceq 2 (L_h + 1)\|Q_2(x)\| \cdot I_n
        \end{align*}
        \item {\bf Part 8.}
        \begin{align*}
            -(R + R_h) L_hR \cdot I_n \preceq B_8 \preceq (R + R_h) L_hR \cdot I_n
        \end{align*}
        \item {\bf Part 9.}
        \begin{align*}
            -(R + R_h) L_hR \cdot I_n \preceq B_9 \preceq (R + R_h) L_hR \cdot I_n
        \end{align*}
        \item {\bf Part 10.}
        \begin{align*}
            -(R + R_h) L_hR \cdot I_n \preceq B_{10} \preceq (R + R_h) L_hR \cdot I_n
        \end{align*}
        \item {\bf Part 11.}
        \begin{align*}
            -(R + R_h) L_hR \cdot I_n \preceq B_{11} \preceq (R + R_h) L_hR \cdot I_n
        \end{align*}
        \item {\bf Part 12.}
        \begin{align*}
            -  R_h R (R + R_h) \cdot I_n \preceq B_{12} \preceq  R_h R (R + R_h) \cdot I_n
        \end{align*}
    \end{itemize}
\end{lemma}
\begin{proof}

{\bf Proof of Part 1.}

We know that 
\begin{align*}
    \|\diag(f(x))^\top Q_2(x)^\top  Q_2(x) \diag(f(x))\|
    \leq & ~ \|Q_2(x)\|^2 \|\diag(f(x))\|_2^2 \\
    \leq & ~ \|Q_2(x)\|^2 
\end{align*}
where the first step follows from Fact~\ref{fac:vec_norm} and Fact~\ref{fac:matrix_norm},
and the second step follows from Fact~\ref{fac:vec_norm} and $\|f(x)\|_1 \leq 1$.

Thus, we have
\begin{align*}
   - \| Q_2(x) \| \cdot I_n \preceq B_2 \preceq \| Q_2(x) \| \cdot I_n
\end{align*}

{\bf Proof of Part 2.}

We have
\begin{align*}
    \|\diag(f(x))^\top Q_2(x)^\top Q_2(x) f(x) f(x)^\top\| 
    \leq & ~ \|Q_2(x)\|^2 \|f(x)\|_2^2 \|\diag(f(x))\| \\
    \leq & ~ \|Q_2(x)\|^2
\end{align*}
where the first step follows from Fact~\ref{fac:vec_norm} and Fact~\ref{fac:matrix_norm},
and the second step follows from Fact~\ref{fac:vec_norm} and $\|f(x)\|_1 \leq 1$.

Thus, we have
\begin{align*}
   - \| Q_2(x) \| \cdot I_n \preceq B_2 \preceq \| Q_2(x) \| \cdot I_n
\end{align*}

{\bf Proof of Part 3}

We have
\begin{align*}
    \|f(x) f(x)^\top Q_2(x)^\top  Q_2(x) \diag(f(x))\|
    \leq & ~ \|Q_2(x)\|^2 \|f(x)\|_2^2 \|\diag(f(x))\| \\
    \leq & ~ \|Q_2(x)\|^2
\end{align*}
where the first step follows from Fact~\ref{fac:vec_norm} and Fact~\ref{fac:matrix_norm},
and the second step follows from Fact~\ref{fac:vec_norm} and $\|f(x)\|_1 \leq 1$.

Thus, we have
\begin{align*}
    - \|Q_2(x)\|^2 \cdot I_n \preceq B_3 \preceq \|Q_2(x)\|^2 \cdot I_n
\end{align*}

{\bf Proof of Part 4}

We have
\begin{align*}
    \|f(x)^\top Q_2(x)^\top Q_2(x) f(x) f(x)^\top\|
    \leq & ~ \|Q_2(x)\|^2 \|f(x)\|_2^3 \\
    \leq & ~ \|Q_2(x)\|^2
\end{align*}
where the first step follows from Fact~\ref{fac:vec_norm} and Fact~\ref{fac:matrix_norm},
and the second step follows from Fact~\ref{fac:vec_norm} and $\|f(x)\|_1 \leq 1$.

Thus, we have
\begin{align*}
    - \|Q_2(x)\|^2 \cdot I_n \preceq B_4 \preceq \|Q_2(x)\|^2 \cdot I_n
\end{align*}

{\bf Proof of Part 5}

We have
\begin{align*}
    \|2 f(x) c(x)^\top Q_2(x) f(x) f(x)^\top\|
    \leq & ~ 2 \|f(x)\|_2^3 \|c(x)\|_2 \|Q_2(x)\| \\
    \leq & ~ 2 (L_h + 1) \|Q_2(x)\|
\end{align*}
where the first step follows from Fact~\ref{fac:vec_norm} and Fact~\ref{fac:matrix_norm},
and the second step follows from Fact~\ref{fac:vec_norm} , $\|f(x)\|_1 \leq 1$, the definition of $c(x)$ and $\|h(x)\|_2 \leq L_h$.

Thus, we have
\begin{align*}
    - 2 (L_h + 1)\|Q_2(x)\| \cdot I_n \preceq B_5 \preceq 2 (L_h + 1)\|Q_2(x)\| \cdot I_n
\end{align*}

{\bf Proof of Part 6}

We have
\begin{align*}
    \|2 f(x) c(x)^\top Q_2(x) \diag(f(x))\|
    \leq & ~ 2 \|f(x)\|_2 \|c(x)\|_2 \|\diag(f(x))\| \|Q_2(x)\| \\
    \leq & ~ 2 (L_h + 1) \|Q_2(x)\|
\end{align*}
where the first step follows from Fact~\ref{fac:vec_norm} and Fact~\ref{fac:matrix_norm},
and the second step follows from Fact~\ref{fac:vec_norm} , $\|f(x)\|_1 \leq 1$, the definition of $c(x)$ and $\|h(x)\|_2 \leq L_h$.

Thus, we have
\begin{align*}
    - 2 (L_h + 1)\|Q_2(x)\| \cdot I_n \preceq B_6 \preceq 2 (L_h + 1)\|Q_2(x)\| \cdot I_n
\end{align*}

{\bf Proof of Part 7}

We have
\begin{align*}
    \|\diag(Q_2(x)^\top c(x)) \diag(f(x))\|
    \leq & ~ \|\diag(Q_2(x)^\top c(x))\| \|\diag(f(x))\| \\
    \leq & ~ \|Q_2(x)^\top c(x)\| \|f(x)\|_2 \\
    \leq & ~ \|Q_2(x)\| \|c(x)\|_2 \|f(x)\|_2 \\
    \leq & ~ 2 (L_h + 1) \|Q_2(x)\|
\end{align*}
where the first step follows from Fact~\ref{fac:matrix_norm},
the second step follows from Fact~\ref{fac:matrix_norm},
the third step follows from Fact~\ref{fac:matrix_norm} and Fact~\ref{fac:vec_norm},
the last step follows from Fact~\ref{fac:vec_norm} , $\|f(x)\|_1 \leq 1$, the definition of $c(x)$ and $\|h(x)\|_2 \leq L_h$.

Thus, we have
\begin{align*}
    - 2 (L_h + 1)\|Q_2(x)\| \cdot I_n \preceq B_7 \preceq 2 (L_h + 1)\|Q_2(x)\| \cdot I_n
\end{align*}

{\bf Proof of Part 8}

We have
\begin{align*}
    & ~ \|\diag(f(x)) A_2 \diag( h''(A_2 f(x)) ) \diag(c(x)) \diag(f(x))\|\\
    \leq & ~ \|A_2\|\|\diag(f(x))\|^2 \|\diag(c(x))\| \|\diag( h''(A_2 f(x)) )\| \\
    \leq & ~ R(R + R_h) \|h''(A_2 f(x))\|_2 \\
    \leq & ~ R(R + R_h) L_h
\end{align*}
where the first step follows from Fact~\ref{fac:matrix_norm},
the second step follows from Fact~\ref{fac:vec_norm} ,$\|f(x)\|_1 \leq 1$ and {\bf Part 2} of Lemma~\ref{lem:function_norm_bounds},
the last step follows from $\|h''(x)\|_2 \leq L_h$.

Thus, we have
\begin{align*}
    -(R + R_h) L_hR \cdot I_n \preceq B_8 \preceq (R + R_h) L_hR \cdot I_n
\end{align*}

{\bf Proof of Part 9}
It is similar to the proof of {\bf Part 8}.

{\bf Proof of Part 10}
It is similar to the proof of {\bf Part 8}.

{\bf Proof of Part 11}
It is similar to the proof of {\bf Part 8}.

{\bf Proof of Part 12}
We have
\begin{align*}
    \|\diag(f(x) f(x)^\top Q_2(x)^\top c(x))\|
    \leq & ~ \|f(x) f(x)^\top Q_2(x)^\top c(x)\|_2 \\
    \leq & ~ \|f(x)\|_2^2 \|Q_2(x)\| \|c(x)\|_2 \\
    \leq & ~ R_h R (R + R_h)
\end{align*}
where the first step follows from Fact~\ref{fac:matrix_norm},
the second step follows from Fact~\ref{fac:vec_norm},
the last step follows from Fact~\ref{fac:vec_norm}, $\|f(x)\|_1$, {\bf Part 2} and {\bf Part 3} of Lemma~\ref{lem:function_norm_bounds}.

Thus, we have
\begin{align*}
    -  R_h R (R + R_h) \cdot I_n \preceq B_{12} \preceq  R_h R (R + R_h) \cdot I_n
\end{align*}
\end{proof}

\subsection{Lower Bound on Hessian}

\label{sub:psd_app:bound}

The goal of this section is to prove Lemma~\ref{lem:convex}.
\begin{lemma}\label{lem:convex}
If the following conditions hold
\begin{itemize}
    \item Let $A_1 \in \R^{n \times d}$.
    \item Let $L_{\tot}$ be defined in Definition~\ref{def:L_tot}
    \item Let $W = \diag(w) \in \R^{n \times n}$. 
    \item Let $W^2 \in \R^{n \times n}$ denote the matrix that $i$-th diagonal entry is $w_{i,i}^2$.
    \item Let $\sigma_{\min}(A_1)$ denote the minimum singular value of $A_1$.
    \item Let $l > 0$ denote a scalar.
\end{itemize}
Then, we have
\begin{itemize}
    \item Part 1. If all $i \in [n]$, $w_{i}^2 \geq 12 R_h L_h R (R + R_h) + l/\sigma_{\min}(A_1)^2$, then  
    \begin{align*}
    \frac{\d^2 L}{\d x^2} \succeq l \cdot I_d
    \end{align*}
    \item Part 2. If all $i \in [n]$, $w_{i}^2 \geq 100 + 12 R_h L_h R (R + R_h) + l/\sigma_{\min}(A_1)^2$, then  
    \begin{align*}
       (1-1/10) \cdot ( B(x) + W^2) \preceq  W^2 \preceq (1+1/10) \cdot (B(x) +W^2)
    \end{align*}
\end{itemize}
\end{lemma}

\begin{proof}

{\bf Proof of Part 1}

    By applying Lemma~\ref{lem:reorganizing}, we have 
    \begin{align*}
        \frac{\d^2 L}{\d x^2}
        = & ~ A_1^\top B(x) A_1 
    \end{align*}
    where 
    \begin{align}\label{eq:hessian_total_step_1}
        B(x) \succeq - 12 R_h L_h R (R + R_h) I_n
    \end{align}
    
    Also, we have
    \begin{align}\label{eq:hessian_total_step_2}
        \frac{\d^2 L_{\tot}}{\d x^2} = \frac{\d^2 L_{\reg}}{\d x^2} + \frac{\d^2 L}{\d x^2}
    \end{align}

    Thus, by applying Lemma~\ref{lem:L_reg_gradient_hessian}, Eq.~\eqref{eq:hessian_total_step_2} can be written as
    \begin{align*}
        \frac{\d^2 L_{\tot}}{\d x^2} 
        = & ~ A_1^\top B(x) A + A^\top W^2 A_1 \\
        = & ~ A_1^\top (B(x) + W^2) A_1,
    \end{align*}
    where the second step follows from simple algebra.

    Let
    \begin{align*}
        D = B(x) + W^2
    \end{align*}
    Then, $\frac{\d^2 L}{\d x^2}$ can be rewritten as
    \begin{align*}
        \frac{\d^2 L}{\d x^2} = A_1^\top D A_1
    \end{align*}

    Now, we can bound $D$ as follows
    \begin{align*}
        D
        \succeq & ~ -12 R_h L_h R (R + R_h) I_n + w_{\min}^2 I_n \\
        = & ~ ( -12 R_h L_h R (R + R_h) + w_{\min}^2) I_n \\ 
        \succeq & ~ \frac{l}{\sigma_{\min}(A_1)^2} I_n
    \end{align*}
    where the first step follows from Lemma~\ref{lem:final_bound} and the fact that $W$ is a diagonal matrix (see from the Lemma statement), the second step follows from simple algebra, and the last step follows from $w_{\min}^2 \geq -12 R_h L_h R (R + R_h) + l/\sigma_{\min}(A_1)^2$.

    Since $D$ is positive definite, then we have 
    \begin{align*}
        A_1^\top D A_1 \succeq \sigma_{\min}(D) \cdot \sigma_{\min}(A_1)^2 I_d \succeq l \cdot I_d
    \end{align*}
    Thus, Hessian is positive definite forever and thus the function is convex.

    {\bf Proof of Part 2}
    This trivially follows from Lemma~\ref{lem:psd_lower_bound_Bi}.
\end{proof}

%%%%Zhao: It seems this file has issue
\section{Hessian is Lipschitz}
\label{sec:hessian_is_lip_app}

In Section~\ref{sub:hessian_is_lip_app:basic}, we present the Lipschitz properties for some basic functions. In Section~\ref{sub:hessian_is_lip_app:summary}, we present the summary of this section. In Section~\ref{sub:hessian_is_lip_app:1}, we analyze the first part which shows that $G_1$ is Lipschitz continuous. 
In Section~\ref{sub:hessian_is_lip_app:2}, we analyze the second part which shows that $G_2$ is Lipschitz continuous. 
In Section~\ref{sub:hessian_is_lip_app:3}, we analyze the third part which shows that $G_3$ is Lipschitz continuous. 
In Section~\ref{sub:hessian_is_lip_app:4}, we analyze the fourth part which shows that $G_4$ is Lipschitz continuous. 
In Section~\ref{sub:hessian_is_lip_app:5}, we analyze the fifth part which shows that $G_5$ is Lipschitz continuous. 
In Section~\ref{sub:hessian_is_lip_app:6}, we analyze the sixth part which shows that $G_6$ is Lipschitz continuous. 

\subsection{Lipschitz Property for Some Basic Functions}

In this section, we present the Lipschitz property for some basic functions.

\label{sub:hessian_is_lip_app:basic}

\begin{lemma}\label{lem:lips_basic_u}
If the following conditions hold
\begin{itemize}
    \item Let $A_1 \in \R^{n \times d}$
    \item Let $x \in \R^d$ where $\|x\|_2 \leq R$
    \item Let $R \geq 4$
    \item $\|A_1\| \leq R$
\end{itemize}
We have
\begin{align*}
    \|\exp(A_1 x)\|_2 \leq \sqrt{n} \exp(R^2)
\end{align*}
\end{lemma}

\begin{proof}
    We have
    \begin{align*}
        \|\exp(A_1 x)\|_2
        \leq & ~ \sqrt{n} \cdot \|\exp(A_1 x)\|_\infty\\
        \leq & ~ \sqrt{n} \cdot \exp(\|A_1 x\|_\infty)\\
        \leq & ~ \sqrt{n} \cdot \exp(\|A_1 x\|_2)\\
        \leq & ~ \sqrt{n} \cdot \exp(R^2)
    \end{align*}
    where the 1st step follows from Fact~\ref{fac:vec_norm}, the 2nd step follows from Fact~\ref{fac:vec_norm}, the 3rd step follows from Fact~\ref{fac:vec_norm}, and the last step follows from $\|A_1\| \leq R$ and $\|x\|_2 \leq R$.
\end{proof}

\begin{lemma}\label{lem:function_norm_bounds}
    If the following conditions hold:
    \begin{itemize}
        \item Let $A_1 \in \R^{n \times d}, A_2 \in \R^{m \times n}$.
        \item Let $q_2(x)$ and $Q_2(x)$ be defined in Definition~\ref{def:Q_q}.
        \item Let $f(x)$ be defined in Definition~\ref{def:f}.
        \item Let $c(x)$ be defined in Definition~\ref{def:c}.
        \item Let $\alpha(x)$ be defined in Definition~\ref{def:alpha}.
        \item Let $u(x)$ be defined in Definition~\ref{def:u}.
        \item Let $R_h > 0$.
        \item Let $\max\{ \| h(A_2 f(x)) \|_2, \|h'(A_2 f(x))\|_2 \} \leq R_h$.
        \item Let $R > 0$.
        \item Suppose that $\|b\|_2 \leq R$.
        \item Suppose that $\|A_1\|, \|A_2\| \leq R$.
    \end{itemize}
    
    Then, we have
    \begin{itemize}
        \item Part 1.$\|f(x)\|_2 \leq \beta^{-1} \cdot \sqrt{n} \cdot \exp(R^2)$
        \item Part 2. $\|c(x)\|_2 \leq R + R_h$
        \item Part 3. $\|Q_2(x)\| \leq R \cdot R_h$
        \item Part 4. $\|q_2(x)\|_2 \leq R \cdot R_h \cdot (R + R_h)$
        \item Part 5. For $i \in [d]$,  $\|p(x)_i\|_2 \leq 2R\beta^{-2} \cdot n \cdot \exp(2R^2)$
    \end{itemize}
\end{lemma}
\begin{proof}
    {\bf Proof of Part 1}
    \begin{align*}
        \|f(x)\|_2
        = & ~ \|\alpha(x)^{-1} \cdot u(x)\|_2 \\
        = & ~ |\alpha(x)^{-1}| \cdot \|u(x)\|_2 \\
        \leq & ~ \beta^{-1} \cdot \|u(x)\|_2 \\
        = & ~ \beta^{-1} \cdot \|\exp(A_1 x)\|_2 \\
        \leq & ~ \beta^{-1} \cdot \sqrt{n} \cdot \exp(R^2),
    \end{align*}
    where the first step follows from the definition of $f(x)$ (see Definition~\ref{def:f}), the second step follows from Fact~\ref{fac:vec_norm}, the third step follows from Eq.~\eqref{eq:alphax-1}, the fourth step follows from the definition of $u(x)$ (see Definition~\ref{def:u}), and the last step follows from Lemma~\ref{lem:lips_basic_u}.

    {\bf Proof of Part 2}
    \begin{align*}
        \|c(x)\|_2
        = & ~ \|h(A_2 f(x)) - b\|_2 \\
        \leq & ~ \|h(A_2 f(x))\|_2 + \|b\|_2 \\
        \leq & ~ R + R_h,
    \end{align*}
    where the first step follows from the definition of $c(x)$ (see Definition~\ref{def:c}), the second step follows from the triangle inequality, and the third step follows from the assumptions from the Lemma statement.

    {\bf Proof of Part 3}
    \begin{align*}
        \|Q_2(x)\|
        = & ~ \|A_2 \diag(h'(A_2 f(x)))\| \\
        \leq & ~ \|A_2\| \cdot \|\diag(h'(A_2 f(x)))\| \\
        \leq & ~ \|A_2\| \cdot \|h'(A_2 f(x))\|_{\infty} \\
        \leq & ~ \|A_2\| \cdot \|h'(A_2 f(x))\|_2 \\
        \leq & ~ R \cdot R_h,
    \end{align*}
    where the first step follows from the definition of $Q_2(x)$ (see Definition~\ref{def:Q_q}),
    the second step follows from Fact~\ref{fac:matrix_norm}, the third step follows from Fact~\ref{fac:vec_norm}, the fourth step follows from Fact~\ref{fac:vec_norm}, and the last step follows from $\|A_2\| \leq R$ and $\|h'(A_2 f(x))\|_2 \leq R_h$.

    {\bf Proof of Part 4}
    \begin{align*}
        \|q_2(x)\|_2
        = & ~ \|Q_2(x)^\top c(x)\|_2 \\
        \leq & ~ \|Q_2(x)^\top\| \|c(x)\|_2 \\
        \leq & ~ \|Q_2(x)\| \|c(x)\|_2 \\
        \leq & ~ R \cdot R_h \cdot (R + R_h),
    \end{align*}
    where the first step follows from the definition of $q_2(x)$ (see Definition~\ref{def:Q_q}),
    the second step follows from Fact~\ref{fac:matrix_norm},
    the third step follows from Fact~\ref{fac:matrix_norm}, and the last step follows from {\bf Part 2} and {\bf Part 3}.

    {\bf Proof of Part 5}
    \begin{align*}
        \|p(x)_i\|
        = & ~ \|f(x) \circ A_{1,*,i} - \langle f(x), A_{1,*,i} \rangle \cdot f(x)\| \\
        \leq & ~ \|f(x) \circ A_{1,*,i}\| + \|\langle f(x), A_{1,*,i} \rangle \cdot f(x)\|_2 \\
        \leq & ~ R \beta^{-1} \cdot \sqrt{n} \cdot \exp(R^2) + \|f(x)\|_2^2 \|A_{1,*,i}\|_2 \\
        \leq & ~ R \beta^{-1} \cdot \sqrt{n} \cdot \exp(R^2) + R\beta^{-2} \cdot n \cdot \exp(2R^2) \\
        \leq & ~ 2R\beta^{-2} \cdot n \cdot \exp(2R^2)
    \end{align*}
    where the first step follows from the definition of $p(x)_i$,
    the second step follows from triangular inequality,
    the third step follows from Fact~\ref{fac:vec_norm} and {\bf Part 1} and $\|A_{1,*,i}\|_2 \leq R$,
    the fourth step follows from {\bf Part 1} and $\|A_{1,*,i}\|_2 \leq R$,
    the last step follows from simple algebra.
\end{proof}

\begin{lemma}\label{lem:norms}
    If the following conditions hold
    \begin{itemize}
        \item Let $A_1 \in \R^{n \times d}, A_2 \in \R^{m \times n}$
        \item Let $\beta \in (0, 0.1)$ 
        \item Let $R \geq 4$
        \item $\|A_1\| \leq R$
        \item $\langle \exp(A_1 x), {\bf 1}_n \rangle \geq \beta$
        \item $\langle \exp(A_1 y), {\bf 1}_n \rangle \geq \beta$
        \item Let $R_f:= 2 \beta^{-2} \cdot n R\exp(2 R^2)$
        \item Let $u(x)$ be defined as Definition~\ref{def:u}.
        \item Let $\alpha(x)$ be defined as Definition~\ref{def:alpha}.
        \item Let $f(x)$ be defined as Definition~\ref{def:f}.
        \item Let $c(x)$ be defined as Definition~\ref{def:c}.
        \item Let $Q_2(x)$ be defined as Definition~\ref{def:Q_q}.
        \item Let $q_2(x)$ be defined as Definition~\ref{def:Q_q}.
        \item Let $p_i(x)$ be defined as Lemma~\ref{lem:hessian_f}. 
        \item Assume $\| h(x) - h(y) \|_2 \leq L_h \cdot \| x -y \|_2$
        \item Assume $\| h'(x)- h'(y) \|_2 \leq L_h \cdot \| x - y \|_2$
    \end{itemize}
    We have
    \begin{itemize}
        \item Part 1. $\|u(x) - u(y)\|_2 \leq R \exp(R^2) \cdot \|x-y\|_2$
        \item Part 2. $|\alpha(x) - \alpha(y)| \leq \sqrt{n} \cdot \|\exp(A_1 x) - \exp(A_1 y)\|_2$
        \item Part 3. $|\alpha(x)^{-1} - \alpha(y)^{-1}| \leq \beta^{-2} \cdot |\alpha(x) - \alpha(y)|$
        \item Part 4. $\|f(x)-f(y)\|_2 \leq R_f \cdot \|x-y\|_2$
        \item Part 5. $\|c(x)-c(y)\|_2 \leq L_h \cdot R \cdot R_f \cdot \|x - y\|_2$
        \item Part 6. $\|Q_2(x) - Q_2(y)\| \leq R^2 R_f L_h \|x - y\|_2$
        \item Part 7. $\|q_2(x) - q_2(y)\|_2 \leq 2 R^2 R_f R_h L_h (R + R_h) \|x - y\|_2$
        \item Part 8. $\|g(x) - g(y)\|_2 \leq 7 \beta^{-2} n L_h R_h R_f R^2 (R + R_h)  \exp(5R^2) \|x - y\|_2$
        \item Part 9. For each $i \in [d] $, $\| p_i(x) - p_i(y) \|_2 \leq 3 R R_f \beta^{-1} \cdot \sqrt{n} \cdot \exp(R^2) \|x - y\|_2$ 
    \end{itemize}
\end{lemma}

\begin{proof}
    {\bf Proof of Part 1.}
    We have 
    \begin{align*}
        \|u(x) - u(y)\|_2
        \leq & ~ \|\exp(A_1 x) - \exp(A_1 y)\|_2\\
        \leq & ~ \exp(R^2) \|A_1 x - A_1 y\|_2\\
        \leq & ~ \exp(R^2)\|A_1\|\|x-y\|_2\\
        \leq & ~ R \exp(R^2) \cdot \|x-y\|_2
    \end{align*}
    where the 1st step follows from the definition of $u(x)$ (see Definition~\ref{def:u}), the 2nd step follows from Fact~\ref{fac:vec_norm}, and the 3rd step follows from Fact~\ref{fac:matrix_norm}, and the last step follows from the assumption in the lemma statement. 

    {\bf Proof of Part 2.}
    We have
    \begin{align*}
        |\alpha(x) - \alpha(y)|
        = & ~ | \langle \exp(A_1x), {\bf 1}_n \rangle - \langle \exp(A_1y), {\bf 1}_n \rangle |\\
        = & ~ | \langle \exp(A_1x) -\exp(A_1y), {\bf 1}_n \rangle |\\
        \leq & ~ \| \exp(A_1x) -\exp(A_1y)\|_2 \cdot \|{\bf 1}_n \|_2\\
        = & ~ \sqrt{n} \cdot \|\exp(A_1 x) - \exp(A_1 y)\|_2
    \end{align*}
    where the 1st step follows from the definition of $\alpha(x)$ (see Definition~\ref{def:alpha}), and the 2nd step follows from Fact~\ref{fac:basic_algebra}, the 3rd step follows from the Cauchy-Schwartz inequality (see Fact~\ref{fac:vec_norm}), and the last step follows from the definition of ${\bf 1}_n$.

    {\bf Proof of Part 3.}

    Note that we have
    \begin{align}\label{eq:alphax-1}
        \alpha(x)^{-1}
        = & ~ \langle \exp(A_1 x), {\bf 1}_n \rangle^{-1} \notag\\
        \leq & ~ \frac{1}{\beta},
    \end{align}
    where the first step follows from the definition of $\alpha(x)$ (see Definition~\ref{def:alpha}) and the second step follows from the assumption from the lemma statement.

    With the same strategy, we can also have
    \begin{align}\label{eq:alphay-1}
        \alpha(y)^{-1} \leq\frac{1}{\beta}.
    \end{align}

    We show that
    \begin{align*}
        |\alpha(x)^{-1} - \alpha(y)^{-1}|
        = & ~ \alpha(x)^{-1} \alpha(y)^{-1} \cdot |\alpha(x) - \alpha(y)| \\
        \leq & ~ \beta^{-2} \cdot |\alpha(x) - \alpha(y)|
    \end{align*}
    where the 1st step follows from simple algebra and the 2nd step follows from combining Eq.~\eqref{eq:alphax-1} and Eq.~\eqref{eq:alphay-1}.

    {\bf Proof of Part 4.}
    We show that
    \begin{align*}
        \|f(x) - f(y)\|_2
        = & ~ \|\alpha(x)^{-1} \exp(A_1 x) - \alpha(y)^{-1} \exp(A_1y)\|_2\\
        \leq & ~ \|\alpha(x)^{-1} \exp(A_1x) - \alpha(x)^{-1} \exp(A_1y)\|_2 + \|\alpha(x)^{-1} \exp(A_1y) - \alpha(y)^{-1} \exp(A_1y)\|_2\\
        \leq & ~ \alpha(x)^{-1}\| \exp(A_1x) - \exp(A_1y)\|_2 + |\alpha(x)^{-1} -\alpha(y)^{-1}| \cdot \|\exp(A_1 y)\|_2 
    \end{align*}
    where the 1st step follows from the definition of $f(x)$ (see Definition~\ref{def:f}), the 2nd step follows from the triangle inequality, and the last step follows from simple algebra.

    For the first term $\alpha(x)^{-1}\| \exp(A_1 x) - \exp(A_1 y)\|_2$, we have
    \begin{align}\label{eq:part1_f_lip}
        \alpha(x)^{-1}\| \exp(A_1 x) - \exp(A_1 y)\|_2
        \leq & ~ \beta^{-1} \|\exp(A_1 x) - \exp(A_1 y)\|_2 \notag \\
        \leq & ~ \beta^{-1} \cdot R \exp(R^2) \cdot \|x-y\|_2
    \end{align}
    where the 1st step follows from {\bf Part 3}, and the 2nd step follows from {\bf Part 1}.

    For the second term $|\alpha(x)^{-1} -\alpha(y)^{-1}| \cdot \|\exp(A_1 y)\|_2 $, we have
    \begin{align}\label{eq:part2_f_lip}
        |\alpha(x)^{-1} -\alpha(y)^{-1}| \cdot \|\exp(A_1 y)\|_2
        \leq & ~ \beta^{-2} \cdot |\alpha(x) - \alpha(y)| \cdot \|\exp(A_1 y)\|_2 \notag \\
        \leq & ~ \beta^{-2} \cdot |\alpha(x) - \alpha(y)| \cdot \sqrt{n} \exp(R^2) \notag \\
        \leq & ~ \beta^{-2} \cdot \sqrt{n} \cdot \|\exp(A_1 x) - \exp(A_1 y)\|_2 \cdot \sqrt{n} \exp(R^2) \notag \\
        \leq & ~ \beta^{-2} \cdot \sqrt{n} \cdot R\exp(R^2) \|x-y\|_2 \cdot \sqrt{n} \exp(R^2) \notag \\
        = & ~ \beta^{-2} \cdot n R\exp(2 R^2) \|x-y\|_2,
    \end{align}
    where the first step follows from {\bf Part 3}, the second step follows from Lemma~\ref{lem:lips_basic_u}, the third step follows from {\bf Part 2}, the fourth step follows from {\bf Part 1}, and the last step follows from simple algebra.

    Therefore, we sum up Eq.~\eqref{eq:part1_f_lip} and Eq.~\eqref{eq:part2_f_lip} to get:
    \begin{align*}
        \|f(x)-f(y)\|_2
        \leq & ~ \beta^{-1} \cdot R\exp(R^2) \cdot \|x-y\|_2 + \beta^{-2} \cdot n R\exp(2 R^2) \|x-y\|_2\\
        \leq & ~ 2 \beta^{-2} \cdot n R\exp(2 R^2) \|x-y\|_2\\
        = & ~ R_f \|x-y\|_2,
    \end{align*}
    where the second step follows from simple algebra and the last step follows from the definition of $R_f$ (see the assumption in the lemma statement).

    {\bf Proof of Part 5.}

    We have
    \begin{align*}
        \|c(x)-c(y)\|_2
        = & ~ \|h(A_2 f(x)) - b - (h(A_2 f(y)) - b)\|_2\\
        = & ~ \|h(A_2 f(x)) - h(A_2 f(y))\|_2\\
        \leq & ~ L_h \cdot \|A_2 f(x) - A_2 f(y)\|_2\\
        \leq & ~ L_h \cdot \|A_2\| \cdot \| f(x) - f(y)\|_2\\
        \leq & ~ L_h \cdot R \cdot R_f \cdot \|x - y\|_2,
    \end{align*}
    where the first step follows from the definition of $c(x)$ (see Definition~\ref{def:c}), the second step follows from simple algebra, the third step follows from the assumption from the Lemma statement, the fourth step follows from Fact~\ref{fac:matrix_norm}, and the last step follows from $\|A_2\| \leq R$ and {\bf Part 4}.

    {\bf Proof of Part 6}
    \begin{align*}
        \|Q_2(x) - Q_2(y)\|
        = & ~ \|A_2 \diag(h'(A_2 f(x))) - A_2 \diag(h'(A_2 f(y))) \| \\
        \leq & ~ \|A_2\| \|h'(A_2 f(x)) - h'(A_2 f(y))\|_2 \\
        \leq & ~ R L_h \|A_2 f(x) - A_2 f(y)\|_2\\
        \leq & ~ R^2 L_h \|f(x) - f(y)\|_2\\
        \leq & ~ R^2 R_f L_h \|x - y\|_2,
    \end{align*}
    where the first step follows from the definition of $Q_2(x)$,
    the second step follows from Fact~\ref{fac:matrix_norm},
    the last step follows from $\|A_2\| \leq R$ and $\|h'(x) - h'(y)\|_2 \leq L_h \|x - y\|_2$, the fourth step follows from $\|A_2\| \leq R$, and the last step follows from {\bf Part 4}.

    {\bf Proof of Part 7}
    \begin{align*}
        \|q_2(x) - q_2(y)\|_2
        = & ~ \|Q_2(x)^\top c(x) - Q_2(y)^\top c(y)\| \\
        = & ~ \|Q_2(x)^\top c(x) - Q_2(x)^\top c(y) + Q_2(x)^\top c(y) - Q_2(y)^\top c(y)\| \\
        \leq & ~ \|Q_2(x) c(x) - Q_2(x) c(y)\| + \|Q_2(x) c(y) - Q_2(y) c(y)\| \\
        \leq & ~ \|Q_2(x)\| \|c(x) - c(y)\|_2 + \|Q_2(x) - Q_2(y)\| \|c(y)\|_2 \\
        \leq & ~ \|Q_2(x)\| \cdot L_h R  R_f \|x - y\|_2 + R^2 R_f L_h \|x - y\|_2 \|c(y)\|_2 \\
        \leq & ~ R^2 R_h R_f  L_h \cdot  \|x - y\|_2 + R^2 R_f L_h \cdot \|x - y\|_2 \cdot (R + R_h) \\
        \leq & ~ 2 R^2 R_f R_h L_h (R + R_h) \|x - y\|_2,
    \end{align*}
    where the first step follows from the definition of $q_2(x)$ (see Definition~\ref{def:Q_q}), the second step follows from simple algebra, the third step follows from the triangle inequality, the fourth step follows from Fact~\ref{fac:matrix_norm}, the fifth step follows from {\bf Part 5} and {\bf Part 6}, the sixth step follows from {\bf Part 2} and {\bf Part 3} of Lemma~\ref{lem:function_norm_bounds}, and the last step follows from simple algebra.

    {\bf Proof of Part 8}
    First, we have
    \begin{align*}
        & ~ \|g(x) - g(y)\|_2\\
        = & ~ \|A_1^\top (f(x)\langle q_2(x),f(x) \rangle + \diag(f(x)) q_2(x)) - A_1^\top (f(y)\langle q_2(y),f(y) \rangle + \diag(f(y)) q_2(y))\|_2 \\
        = & ~ \|A_1^\top ((f(x)\langle q_2(x),f(x) \rangle + \diag(f(x)) q_2(x)) - (f(y)\langle q_2(y),f(y) \rangle + \diag(f(y)) q_2(y)))\|_2 \\
        \leq & ~ \|A_1\| \|(f(x)\langle q_2(x),f(x) \rangle - f(y)\langle q_2(y),f(y) \rangle) + (\diag(f(x)) q_2(x)) - \diag(f(y)) q_2(y)))\|_2 \\
        \leq & ~ R (\|f(x)\langle q_2(x),f(x) \rangle - f(y)\langle q_2(y),f(y) \rangle\|_2 + \|\diag(f(x)) q_2(x) - \diag(f(y)) q_2(y)\|_2)
    \end{align*}
    where the first step follows from the definition of $g(x)$ (see Definition~\ref{def:g}),
    the second step follows from simple algebra,
    the third step follows from Fact~\ref{fac:matrix_norm}, and the last step follows from $\|A_1\| \leq R$ and Fact~\ref{fac:vec_norm}.

    For convenience, we define
    \begin{align*}
        C_1 : & = f(x)\langle q_2(x),f(x) \rangle - f(y)\langle q_2(y),f(y) \rangle \\
        C_2 : & = \diag(f(x)) q_2(x) - \diag(f(y)) q_2(y),
    \end{align*}
    so we have
    \begin{align}\label{eq:gx-gy}
        \|g(x) - g(y)\|_2 \leq R (\|C_1\|_2 + \|C_2\|_2).
    \end{align}

    For $C_1$, we define
    \begin{align*}
        C_{1,1} : & = f(x)\langle q_2(x),f(x) \rangle - f(x)\langle q_2(x),f(y) \rangle \\
        C_{1,2} : & = f(x)\langle q_2(x),f(y) \rangle - f(x)\langle q_2(y),f(y) \rangle \\
        C_{1,3} : & = f(x)\langle q_2(y),f(y) \rangle - f(y)\langle q_2(y),f(y) \rangle
    \end{align*}

    Then, we can rewrite $\|C_1\|_2$ as follows
    \begin{align}\label{eq:C_1}
        \|C_1\|_2 = \|C_{1,1} + C_{1,2} + C_{1,3}\|_2
    \end{align}

    First, we upper bound $\|C_{1,1}\|_2$:
    \begin{align}\label{eq:C_11}
        \|C_{1,1}\|_2 
        \leq & ~ \|f(x)\|_2 \|q_2(x)\|_2 \|f(x) - f(y)\|_2 \notag\\
        \leq & ~ \beta^{-1} \sqrt{n} R_h R (R + R_h)  \exp(5R^2) \|f(x) - f(y)\|_2 \notag\\
        \leq & ~ \beta^{-1} \sqrt{n} R_h R_f R (R + R_h)  \exp(5R^2) \|x - y\|_2,
    \end{align}
    where the first step follows from the definition of $C_{1,1}$ and Fact~\ref{fac:vec_norm},
    the second step follows from {\bf Part 1} and {\bf Part 4} of Lemma~\ref{lem:function_norm_bounds},
    the third step follows from {\bf Part 4}. 

    Next, we upper bound $\|C_{1,2}\|_2$:
    \begin{align}\label{eq:C_12}
        \|C_{1,2}\|_2
        \leq & ~ \|f(x)\|_2 \|q_2(x) - q_2(y)\|_2 \|f(y)\|_2 \notag\\
        \leq & ~ \beta^{-2} n \exp(2R^2) \|q_2(x) - q_2(y)\|_2 \notag\\
        \leq & ~ 2 \beta^{-2} n \exp(2R^2) R^2 R_f R_h L_h (R + R_h) \|x - y\|_2,
    \end{align}
    where the first step follows from the definition of $C_{1,2}$ and Fact~\ref{fac:vec_norm},
    the second step follows from {\bf Part 1} of Lemma~\ref{lem:function_norm_bounds}, the third step follows from {\bf Part 7}.

    Then, we upper bound $\|C_{1,3}\|_2$:
    \begin{align}\label{eq:C_13}
        \|C_{1,3}\|
        \leq & ~ \|f(x) - f(y)\|_2 \|q_2(y)\|_2 \|f(y)\|_2 \notag\\
        \leq & ~ \|f(x) - f(y)\|_2 R R_h (R + R_h) \beta^{-1} \sqrt{n} \exp(R^2) \notag\\
        \leq & ~ \beta^{-1} \sqrt{n} R_h R_f R (R + R_h) \exp(R^2) \|x - y\|_2,
    \end{align}
    where the first step follows from the definition of $C_{1,3}$ and Fact~\ref{fac:vec_norm},
    the second step follows from {\bf Part 1} and {\bf Part 4} of Lemma~\ref{lem:function_norm_bounds}, and the third step follows from {\bf Part 4}.

    Now, it follows from combining the bound of $C_{1,1},C_{1,2}$ and $C_{1,3}$, we obtained the bound for $\|C_1\|_2$:
    \begin{align}\label{eq:C_1_bound}
        \|C_1\|_2 
        = & ~ \|C_{1,1} + C_{1,2} + C_{1,3}\|_2 \notag\\
        = & ~ \|C_{1,1}\|_2 + \|C_{1,2}\|_2 + \|C_{1,3}\|_2 \notag\\
        \leq & ~ 4 \beta^{-2} n R_h R_f R (R + R_h)  \exp(5R^2) L_h \|x - y\|_2,
    \end{align}
    where the first step follows from the definition of $C_1$ (see Eq.~\eqref{eq:C_1}), the second step follows from the triangle inequality, and the third step follows from combining Eq.~\eqref{eq:C_11}, Eq.~\eqref{eq:C_12}, and Eq.~\eqref{eq:C_13}.

    Then, we upper bound $\|C_2\|$ as follows:
    \begin{align}\label{eq:C_2_bound}
        \|C_2\|
        = & ~ \|\diag(f(x)) q_2(x) - \diag(f(x)) q_2(y) + \diag(f(x)) q_2(y) - \diag(f(y)) q_2(y)\|_2 \notag\\
        \leq & ~ \|\diag(f(x)) q_2(x) - \diag(f(x)) q_2(y)\|_2 + \|\diag(f(x)) q_2(y) - \diag(f(y)) q_2(y)\|_2 \notag\\
        \leq & ~ \|f(x)\|_2 \| q_2(x) - q_2(y)\|_2 + \|f(x) - f(y)\|_2 \|q_2(y)\|_2 \notag\\
        \leq & ~ \beta^{-1} \sqrt{n} \exp(R^2) \| q_2(x) - q_2(y)\|_2 + \|f(x) - f(y)\|_2 R R_h (R + R_h) \notag\\
        \leq & ~ 2\beta^{-1} \sqrt{n} \exp(R^2) R^2 R_f R_h L_h (R + R_h) \|x - y\|_2 + R_f \|x - y\|_2 R R_h (R + R_h) \notag\\
        \leq & ~ 3 \beta^{-1} \sqrt{n} R^2 R_f R_h L_h (R + R_h) \exp(R^2) \|x - y\|_2,
    \end{align}
    where the first step follows from the definition of $C_2$,
    the second step follows from Fact~\ref{fac:matrix_norm},
    the third step follows from Fact~\ref{fac:vec_norm},
    the fourth step follows from {\bf Part 1} and {\bf Part 4} of Lemma~\ref{lem:function_norm_bounds}, the fifth step follows from {\bf Part 4} and {\bf Part 7}, and the last step follows from simple algebra.

    Finally, we obtained the bound for $\|g(x) - g(y)\|_2$:
    
    \begin{align*}
        \|g(x) - g(y)\|_2
        \leq & ~ R (\|C_1\|_2 + \|C_2\|_2) \\
        \leq & ~ 4 \beta^{-2} n R_h R_f R (R + R_h)  \exp(5R^2) L_h \|x - y\|_2 \\
        + & ~ 3 \beta^{-1} \sqrt{n} R^2 R_f R_h L_h (R + R_h) \exp(R^2)) \|x - y\|_2 \\
        \leq & ~ 7 \beta^{-2} n L_h R_h R_f R^2 (R + R_h)  \exp(5R^2) \|x - y\|_2,
    \end{align*}
    where the first step follows from Eq.~\eqref{eq:gx-gy}, the second step follows from combining Eq.~\eqref{eq:C_1_bound} and Eq.~\eqref{eq:C_2_bound}, and the last step follows from simple algebra.

    {\bf Proof of Part 9.}
    Note that
    \begin{align*}
        \|p(x)_i - p(y)_i\|_2
        = & ~ \|f(x) \circ A_{1,*,i} - \langle f(x), A_{1,*,i} \rangle \cdot f(x) - f(x) \circ A_{1,*,i} +  \langle f(y), A_{1,*,i} \rangle \cdot f(y) \|_2 \\
        = & ~ \|(f(x) - f(y)) \circ A_{1, *, i} + (\langle f(x), A_{1,*,i} \rangle \cdot f(x) - \langle f(y), A_{1,*,i} \rangle \cdot f(y))\|_2 \\
        \leq & ~ \|(f(x) - f(y)) \circ A_{1, *, i}\|_2 + \|\langle f(x), A_{1,*,i} \rangle \cdot f(x) - \langle f(y), A_{1,*,i} \rangle \cdot f(y)\|_2
    \end{align*}
    where the first step follows from the definition of $p(x)_i$,
    the second step follows from simple algebra,
    the third step follows from triangular inequality.

    For the first term above, we have
    \begin{align*}
        \|(f(x) - f(y)) \circ A_{1, *, i}\|_2
        \leq & ~ \|A_{1, *, i}\|_2 \|f(x) - f(y)\|_2 \\
        \leq & ~ R R_f \|x - y\|_2
    \end{align*}
    where the first step follows from Fact~\ref{fac:matrix_norm},
    the second step follows from $\|A\| \leq R$ and {\bf Part 4}.

    For the second term, we have
    \begin{align*}
        & ~ \|\langle f(x), A_{1,*,i} \rangle \cdot f(x) - \langle f(y), A_{1,*,i} \rangle \cdot f(y)\|_2 \\
        = & ~ \|\langle f(x), A_{1,*,i} \rangle \cdot f(x) - \langle f(x), A_{1,*,i} \rangle \cdot f(y) + \langle f(x), A_{1,*,i} \rangle \cdot f(y) - \langle f(y), A_{1,*,i} \rangle \cdot f(y)\|_2 \\
        \leq & ~ \|\langle f(x), A_{1,*,i} \rangle \cdot (f(x) -  f(y))\|_2 + \|\langle f(x) - f(y), A_{1,*,i} \rangle \cdot f(y)\|_2 \\
        \leq & ~ 2\|f(x)\|_2 \|A_{1,*,i}\|_2 \|f(x) -  f(y)\|_2 \\
        \leq & ~ 2 R R_f \beta^{-1} \cdot \sqrt{n} \cdot \exp(R^2) \|x - y\|_2
    \end{align*}
    where the first step follows from simple algebra,
    the second step follows from simple algebra,
    the third step follows from triangular inequality,
    the fourth step follows from Fact~\ref{fac:vec_norm},
    the last step follows from {\bf Part 4} and {\bf Part 1} of Lemma~\ref{lem:function_norm_bounds}

    Thus, we have
    \begin{align*}
        \|p(x)_i - p(y)_i\|_2
        \leq & ~ (2 R R_f \beta^{-1} \cdot \sqrt{n} \cdot \exp(R^2) + R R_f) \|x - y\|_2 \\
        \leq & ~ 3 R R_f \beta^{-1} \cdot \sqrt{n} \cdot \exp(R^2) \|x - y\|_2
    \end{align*}
\end{proof}

\subsection{Summary of Six Steps}

In this section, we summarize the key results from the following sections. 

\label{sub:hessian_is_lip_app:summary}

\begin{definition}\label{def:G_i}
    If the following conditions hold
    \begin{itemize}
        \item Let $L(x)$ be defined in Definition~\ref{def:L}
        \item Let $Q_2(x)$ and $q_2(x)$ be defined as in Definition~\ref{def:Q_q}
        \item Let $\frac{\d^2 L}{\d x_i^2}$ and $\frac{\d^2 L}{\d x_i \d x_j}$ be computed in Lemma~\ref{lem:hessian_L_informal}
    \end{itemize}
    Then we define
    \begin{align*}
        G_1(x) : & = \langle  Q_2(x) \cdot p(x)_j , Q_2(x) \cdot p(x)_i \rangle \\
        G_2(x) : & = \langle c(x) , \diag( \diag( h''(A_2 f(x)) )  \cdot  A_2  \cdot p(x)_j ) \cdot A_2 \cdot p(x)_i \rangle \\
        G_3(x) : & = \langle c(x), 2 Q_2(x) \cdot f(x) \rangle \cdot \langle f(x) , A_{1,*,i} \rangle \langle f(x) , A_{1,*,j} \rangle \\
        G_4(x) : & = \langle c(x) , Q_2(x) \cdot f(x) \rangle \cdot \langle f(x) , A_{1,*,i} \circ A_{1,*,j} \rangle \\
        G_5(x) : & = \langle c(x) , 2 Q_2(x) \cdot ( f(x) \circ A_{1,*,j}) \rangle \cdot \langle f(x) , A_{1,*,i} \rangle \\
        G_6(x) : & = \langle c(x) , Q_2(x) \cdot ( A_{1,*,i} \circ f(x) \circ A_{1,*,j}) \rangle
    \end{align*}
\end{definition}

\begin{lemma}\label{lem:eight_steps}
    Let $G_i$ be defined in Definition~\ref{def:G_i}, then we have
    \begin{align*}
        \|\sum_{i=1}^6 (G_i(x) - G_i(y))\| \leq 59(R + R_h) n^2  \exp(4R^2) \beta^{-4} R^5 R_h^2 R_f L_h \|x - y\|_2
    \end{align*}
\end{lemma}
\begin{proof}
    \begin{align*}
        & ~ \|\sum_{i=1}^6 (G_i(x) - G_i(y))\| \\
        \leq & ~ \sum_{i=1}^6 \|G_i(x) - G_i(y)\| \\
        \leq & ~ 59(R + R_h) n^2  \exp(4R^2) \beta^{-4} R^5 R_h^2 R_f L_h \|x - y\|_2
    \end{align*}
\end{proof}

\subsection{Step 1: \texorpdfstring{$G_1$}{} is Lipschitz continuous}
\label{sub:hessian_is_lip_app:1}

In this section, we show that $G_1$ is Lipschitz continuous.

\begin{lemma}
    Let $G_1$ be defined in Definition~\ref{def:G_i}, then we have
    \begin{align*}
        |G_1(x) - G_1(y)| \leq 8 R_h^2 R_f R^5 L_h (R + R_h)  \beta^{-4} \cdot n^2 \cdot \exp(4R^2)  \|x - y\|_2
    \end{align*}
\end{lemma}
\begin{proof}
    Note that
    \begin{align*}
        |G_1(x) - G_1(y)|
        = & ~ |\langle  Q_2(x) \cdot p(x)_j , Q_2(x) \cdot p(x)_i \rangle - \langle  Q_2(y) \cdot p(y)_j , Q_2(y) \cdot p(y)_i \rangle|
    \end{align*}
    For simplicity, we define
    \begin{align*}
        G_{1, 1} : & = \langle  Q_2(x) \cdot p(x)_j , Q_2(x) \cdot p(x)_i \rangle - \langle  Q_2(x) \cdot p(x)_j , Q_2(x) \cdot p(y)_i \rangle \\
        G_{1, 2} : & = \langle  Q_2(x) \cdot p(x)_j , Q_2(x) \cdot p(y)_i \rangle - \langle  Q_2(x) \cdot p(x)_j , Q_2(y) \cdot p(y)_i \rangle \\
        G_{1, 3} : & = \langle  Q_2(x) \cdot p(x)_j , Q_2(y) \cdot p(y)_i \rangle - \langle  Q_2(x) \cdot p(y)_j , Q_2(y) \cdot p(y)_i \rangle \\
        G_{1, 4} : & = \langle  Q_2(x) \cdot p(y)_j , Q_2(y) \cdot p(y)_i \rangle - \langle  Q_2(y) \cdot p(y)_j , Q_2(y) \cdot p(y)_i \rangle 
    \end{align*}
    Than it's apparent that
    \begin{align*}
        |\langle  Q_2(x) \cdot p(x)_j , Q_2(x) \cdot p(x)_i \rangle - \langle  Q_2(y) \cdot p(y)_j , Q_2(y) \cdot p(y)_i \rangle| = |G_{1, 1} + G_{1, 2} + G_{1, 3} + G_{1, 4}|
    \end{align*}
    Since $G_{1, i}, i \in [4]$ is similar, we only need to bound $G_{1, 1}$ and $G_{1, 2}$, for $\|G_{1, 1}\|$, we have
    \begin{align*}
        |G_{1,1}|
        = & ~ |\langle  Q_2(x) \cdot p(x)_j , Q_2(x) \cdot (p(x)_i - p(y)_i) \rangle| \\
        \leq & ~ \|Q_2(x) \cdot p(x)_j\|_2 \|Q_2(x) \cdot (p(x)_i - p(y)_i)\|_2 \\
        \leq & ~ \|Q_2(x)\|^2 \|p(x)_j\|_2 \|p(x)_i - p(y)_i\|_2 \\
        \leq & ~ 6 R_h^2 R^4 R_f \beta^{-3} \cdot n^{\frac{3}{2}} \cdot \exp(3R^2) \cdot  \|x - y\|_2
    \end{align*}
    where the first step follows from the definition of $G_{1,1}$,
    the second step follows from Fact~\ref{fac:vec_norm},
    the third step follows from Fact~\ref{fac:matrix_norm},
    the last step follows from {\bf Part 3} and {\bf Part 5} of Lemma~\ref{lem:function_norm_bounds} and {\bf Part 8} of Lemma~\ref{lem:norms}.

    For $G_{1,2}$, we have
    \begin{align*}
        |G_{1,2}|
        = & ~ |\langle  Q_2(x) \cdot p(x)_j , (Q_2(x) - Q_2(y)) \cdot p(y)_i \rangle| \\
        \leq & ~ \|Q_2(x)p(x)_j\|_2 \|Q_2(x) - Q_2(y)\| \|p(y)_i\|_2 \\
        \leq & ~ \|Q_2(x)\| \|p(x)_j\|_2^2 \|Q_2(x) - Q_2(y)\| \\
        \leq & ~ 2 R_h^2 R_f R^5 L_h (R + R_h)  \beta^{-4} \cdot n^2 \cdot \exp(4R^2)  \|x - y\|_2
    \end{align*}
    where the first step follows from the definition of $G_{1,2}$,
    the second step follows from Fact~\ref{fac:vec_norm},
    the third step follows from Fact~\ref{fac:vec_norm},
    the last step follows from {\bf Part 3} and {\bf Part 5} of Lemma~\ref{lem:function_norm_bounds} and {\bf Part 6} of Lemma~\ref{lem:norms}.

    Thus, it follows from combining the upper bounds of $|G_{1,i}|$, we have
    \begin{align*}
        |G_1(x) - G_1(y)|
        = & ~ |G_{1, 1} + G_{1, 2} + G_{1, 3} + G_{1, 4}| \\
        \leq & ~ |G_{1, 1}| + |G_{1, 2}| + |G_{1, 3}| + |G_{1, 4}| \\
        \leq & ~ 8 R_h^2 R_f R^5 L_h (R + R_h)  \beta^{-4} \cdot n^2 \cdot \exp(4R^2)  \|x - y\|_2
    \end{align*}
\end{proof}

\subsection{Step 2: \texorpdfstring{$G_2$}{} is lipschitz continuous}
\label{sub:hessian_is_lip_app:2}

In this section, we show that $G_2$ is Lipschitz continuous.

\begin{lemma}
    Let $G_2(x)$ be defined in Definition~\ref{def:G_i}, then, we have
    \begin{align*}
        |G_2(x) - G_2(y)|
        \leq & ~ 24 R_h R_f R^4 (R + R_h) \beta^{-4} n^2 \exp(4 R^2) \|x - y\|_2
    \end{align*}
\end{lemma}
\begin{proof}
    Note that
    \begin{align*}
        & ~ |G_2(x) - G_2(y)| \\
        = & ~ |\langle c(x) , \diag( \diag( h''(A_2 f(x)) )  \cdot  A_2  \cdot p(x)_j ) \cdot A_2 \cdot p(x)_i \rangle - \langle c(y) , \diag( \diag( h''(A_2 f(y)) )  \cdot  A_2  \cdot p(y)_j ) \cdot A_2 \cdot p(y)_i \rangle|
    \end{align*}

    For simplicity, we define
    \begin{align*}
        G_{2, 1} : & = \langle c(x) , \diag( \diag( h''(A_2 f(x)) )  A_2   p(x)_j ) A_2  p(x)_i \rangle - \langle c(x) , \diag( \diag( h''(A_2 f(x)) )    A_2   p(x)_j )  A_2  p(y)_i \rangle \\
        G_{2, 2} : & = \langle c(x) , \diag( \diag( h''(A_2 f(x)) )    A_2  p(x)_j )  A_2  p(y)_i \rangle - \langle c(x) , \diag( \diag( h''(A_2 f(x)) )   A_2   p(y)_j )  A_2  p(y)_i \rangle \\
        G_{2, 3} : & = \langle c(x) , \diag( \diag( h''(A_2 f(x)) )   A_2  p(y)_j ) A_2  p(y)_i \rangle - \langle c(x) , \diag( \diag( h''(A_2 f(y)) )   A_2  p(y)_j ) A_2  p(y)_i \rangle \\
        G_{2, 4} : & = \langle c(x) , \diag( \diag( h''(A_2 f(y)) )   A_2  p(y)_j ) A_2  p(y)_i \rangle - \langle c(y) , \diag( \diag( h''(A_2 f(y)) )   A_2  p(y)_j )  A_2  p(y)_i \rangle
    \end{align*}

    First, we upper bound $|G_{2, 1}|$:
    \begin{align*}
        |G_{2, 1}|
        = & ~  |\langle c(x) , \diag( \diag( h''(A_2 f(x)) )  A_2   p(x)_j ) A_2  (p(x)_i -  p(y)_i) \rangle| \\
        \leq & ~ \|c(x)\|_2 \|h''(A_2 f(x))\| \|A_2\|^2 |p(x)_j| |p(x)_i -  p(y)_i| \\
        \leq & ~ 6 R_h R_f R^4 (R + R_h) \beta^{-3} n^{\frac{3}{2}} \exp(3 R^2) \|x - y\|_2
    \end{align*}
    where the first step follows from the definition of $G_{2, 1}$,
    the second step follows from Fact~\ref{fac:vec_norm} and Fact~\ref{fac:matrix_norm},
    the last step follows from {\bf Part 2} and {\bf Part 5} of Lemma~\ref{lem:function_norm_bounds}, {\bf Part 9} of Lemma~\ref{lem:norms} and $\|A_2\| \leq R$.

    Since $G_{2, 1}$ and $G_{2, 2}$ are similar, we directly obtained the bound for $|G_{2, 2}|$:
    \begin{align*}
        |G_{2, 2}| \leq 6 R_h R_f R^4 (R + R_h) \beta^{-3} n^{\frac{3}{2}} \exp(3 R^2) \|x - y\|_2
    \end{align*}

    Next, we uppper bound $|G_{2, 3}|$:
    \begin{align*}
        |G_{2, 3}|
        = & ~ |\langle c(x) , (\diag( \diag( h''(A_2 f(x)) ) - \diag( \diag( h''(A_2 f(y)) ) )  A_2  p(y)_j ) A_2  p(y)_i \rangle| \\
        \leq & ~ \|c(x)\|_2 \|h''(A_2(f(x))) - h''(A_2(f(y)))\| \|A_2\|^2 |p(x)_i|^2 \\
        \leq & ~ 4(R + R_h) L_h R^4\beta^{-4} n^2 \exp(4R^2)
    \end{align*}
    where the first step follows from the definition of $G_{2, 1}$,
    the second step follows from Fact~\ref{fac:vec_norm} and Fact~\ref{fac:matrix_norm},
    the last step follows from {\bf Part 2} and {\bf Part 5} of Lemma~\ref{lem:function_norm_bounds}, $\|A_2\| \leq R$ and the assumption that $h(x)$ is $L_h$-lipschitz continuous.

    Last, we upper bound $|G_{2, 4}|$:
    \begin{align*}
        |G_{2, 4}|
        = & ~ |\langle c(x) - c(y) , \diag( \diag( h''(A_2 f(y)) )   A_2  p(y)_j )  A_2  p(y)_i \rangle| \\
        \leq & ~ \|c(x) - c(y)\|_2 \|h''(A_2 f(y))\| \|A_2\|^2 |p(x)_i|^2 \\
        \leq & ~ 4 L_h R_f R^5 R_h \beta^{-4} n^2 \exp(4R^2) \|x - y\|_2
    \end{align*}
    where the first step follows from the definition of $G_{2, 1}$,
    the second step follows from Fact~\ref{fac:vec_norm} and Fact~\ref{fac:matrix_norm},
    the last step follows from {\bf Part 2} and {\bf Part 5} of Lemma~\ref{lem:function_norm_bounds}, $\|A_2\| \leq R$ and {\bf Part 5} of Lemma~\ref{lem:norms}.

    Finally, we have
    \begin{align*}
        |G_2(x) - G_2(y)|
        = & ~ |\sum_{i=1}^4 G_{2, i}| \\
        \leq & ~ 24 R_h R_f R^4 (R + R_h) \beta^{-4} n^2 \exp(4 R^2) \|x - y\|_2
    \end{align*}
\end{proof}

\subsection{Step 3: \texorpdfstring{$G_3$}{} is lipschitz continuous}
\label{sub:hessian_is_lip_app:3}

In this section, we show that $G_3$ is Lipschitz continuous.

\begin{lemma}\label{lem:lipschitz_of_G_3}
    Let $G_3$ be defined in Definition~\ref{def:G_i}, then we have
    \begin{align*}
        |G_3(x) - G_3(y)| \leq 10 (R + R_h) R^4 R_f L_h \beta^{-3} \cdot n^{\frac{3}{2}} \cdot \exp(3R^2) \|x - y\|_2 \cdot  \|x - y\|_2
    \end{align*}
\end{lemma}
\begin{proof}
    Note that
    \begin{align*}
        & ~ |G_3(x) - G_3(y)| \\
        = & ~ |\langle c(x), 2 Q_2(x) \cdot f(x) \rangle \cdot \langle f(x) , A_{1,*,i} \rangle \langle f(x) , A_{1,*,j} \rangle - \langle c(y), 2 Q_2(y) \cdot f(y) \rangle \cdot \langle f(y) , A_{1,*,i} \rangle \langle f(y) , A_{1,*,j} \rangle|
    \end{align*}
    For simplicity, we define
    \begin{align*}
        G_{3, 1} : & = \langle c(x), 2 Q_2(x) \cdot f(x) \rangle \cdot \langle f(x) , A_{1,*,i} \rangle \langle f(x) , A_{1,*,j} \rangle - \langle c(x), 2 Q_2(x) \cdot f(x) \rangle \cdot \langle f(x) , A_{1,*,i} \rangle \langle f(y) , A_{1,*,j} \rangle \\
        G_{3, 2} : & = \langle c(x), 2 Q_2(x) \cdot f(x) \rangle \cdot \langle f(x) , A_{1,*,i} \rangle \langle f(y) , A_{1,*,j} \rangle - \langle c(x), 2 Q_2(x) \cdot f(x) \rangle \cdot \langle f(y) , A_{1,*,i} \rangle \langle f(y) , A_{1,*,j} \rangle \\
        G_{3, 3} : & = \langle c(x), 2 Q_2(x) \cdot f(x) \rangle \cdot \langle f(y) , A_{1,*,i} \rangle \langle f(y) , A_{1,*,j} \rangle - \langle c(x), 2 Q_2(x) \cdot f(y) \rangle \cdot \langle f(y) , A_{1,*,i} \rangle \langle f(y) , A_{1,*,j} \rangle \\
        G_{3, 4} : & = \langle c(x), 2 Q_2(x) \cdot f(y) \rangle \cdot \langle f(y) , A_{1,*,i} \rangle \langle f(y) , A_{1,*,j} \rangle - \langle c(x), 2 Q_2(y) \cdot f(y) \rangle \cdot \langle f(y) , A_{1,*,i} \rangle \langle f(y) , A_{1,*,j} \rangle \\
        G_{3, 5} : & = \langle c(x), 2 Q_2(y) \cdot f(y) \rangle \cdot \langle f(y) , A_{1,*,i} \rangle \langle f(y) , A_{1,*,j} \rangle - \langle c(y), 2 Q_2(y) \cdot f(y) \rangle \cdot \langle f(y) , A_{1,*,i} \rangle \langle f(y) , A_{1,*,j} \rangle
    \end{align*}

    First, we upper bound $|G_{3, 1}|$:
    \begin{align*}
        |G_{3, 1}|
        = & ~ |\langle c(x), 2 Q_2(x) \cdot f(x) \rangle \cdot \langle f(x) , A_{1,*,i} \rangle \langle f(x) - f(y) , A_{1,*,j} \rangle| \\
        \leq & ~ \|c(x)\|_2 2 \|Q(x)\|  \|f(x)\|_2^2 \|A_{1,*,i}\|_2^2 \|f(x) - f(y)\|_2 \\
        \leq & ~ 2(R + R_h) R_f R_h R^3 \beta^{-2} \cdot n \cdot \exp(2R^2) \|x - y\|_2
    \end{align*}
    where the first step follows from the definition of $G_{3, 1}$,
    the second step follows from Fact~\ref{fac:vec_norm} and Fact~\ref{fac:matrix_norm},
    the last step follows from {\bf Part 1}, {\bf Part 2}, and {\bf Part 3} of Lemma~\ref{lem:function_norm_bounds} and {\bf Part 4} of Lemma~\ref{lem:norms}.

    Since $G_{3, 1}$ and $G_{3, 2}$ are symmetry, we directly obtained the bound for $|G_{3, 2}|$:
    \begin{align*}
        |G_{3, 2}| \leq & ~ 2(R + R_h) R_f R_h R^3 \beta^{-2} \cdot n \cdot \exp(2R^2) \|x - y\|_2
    \end{align*}

    Then, we upper bound $|G_{3, 3}|$:
    \begin{align*}
        |G_{3, 3}|
        = & ~ |\langle c(x), 2 Q_2(x) \cdot (f(x) - f(y)) \rangle \cdot \langle f(y) , A_{1,*,i} \rangle \langle f(y) , A_{1,*,j} \rangle| \\
        \leq & ~ \|c(x)\|_2 2 \|Q(x)\|  \|f(x)\|_2^2 \|A_{1,*,i}\|_2^2 \|f(x) - f(y)\|_2 \\
        \leq & ~ 2(R + R_h) R_f R_h R^3 \beta^{-2} \cdot n \cdot \exp(2R^2) \|x - y\|_2
    \end{align*}
    where the first step follows from the definition of $G_{3, 2}$,
    the second step follows from Fact~\ref{fac:vec_norm} and Fact~\ref{fac:matrix_norm},
    the last step follows from {\bf Part 1}, {\bf Part 2}, and {\bf Part 3} of Lemma~\ref{lem:function_norm_bounds} and {\bf Part 4} of Lemma~\ref{lem:norms}.

    Next, we upper bound $|G_{3, 4}|$:
    \begin{align*}
        |G_{3, 4}|
        = & ~ |\langle c(x), 2 (Q_2(x) -  Q_2(y)) \cdot f(y) \rangle \cdot \langle f(y) , A_{1,*,i} \rangle \langle f(y) , A_{1,*,j} \rangle| \\
        \leq & ~ \|c(x)\|_2 2 \|Q_2(x) - Q_2(y)\| \|f(x)\|_2^3 \|A_{1,*,j}\|_2^2 \\
        \leq & ~ 2 (R + R_h) R^4 R_f L_h \beta^{-3} \cdot n^{\frac{3}{2}} \cdot \exp(3R^2) \|x - y\|_2
    \end{align*}
    where the first step follows from the definition of $G_{3, 4}$,
    the second step follows from Fact~\ref{fac:vec_norm} and Fact~\ref{fac:matrix_norm},
    the last step follows from {\bf Part 1}, {\bf Part 2} of Lemma~\ref{lem:function_norm_bounds} and {\bf Part 6} of Lemma~\ref{lem:norms} and $\|A_{1,*,j}\|_2 \leq R$.

    Finally, we upper bound $\|G_{3, 5}\|$:
    \begin{align*}
        |G_{3, 5}|
        = & ~ |\langle c(x) - c(y), 2 Q_2(y) \cdot f(y) \rangle \cdot \langle f(y) , A_{1,*,i} \rangle \langle f(y) , A_{1,*,j} \rangle| \\
        \leq & ~ \|c(x) - c(y)\|_2 2 \|Q_2(y)\| \|f(x)\|_2^3 \|A_{1,*,j}\|_2^2 \\
        \leq & ~ 2 L_h R^4 R_f R_h \beta^{-3} \cdot n^{\frac{3}{2}} \cdot \exp(3R^2) \|x - y\|_2
    \end{align*}
    where the first step follows from the definition of $G_{3, 4}$,
    the second step follows from Fact~\ref{fac:vec_norm} and Fact~\ref{fac:matrix_norm},
    the last step follows from {\bf Part 1}, {\bf Part 3} of Lemma~\ref{lem:function_norm_bounds} and {\bf Part 5} of Lemma~\ref{lem:norms} and $\|A_{1,*,j}\|_2 \leq R$.

    Thus, we obtained the bound for $|G_3(x) - G_3(y)|$:
    \begin{align*}
        |G_3(x) - G_3(y)|
        = & ~ |\sum_{i=1}^5 G_i| \\
        \leq & ~ \sum_{i=1}^5 |G_i| \\
        \leq & ~ 10 (R + R_h) R^4 R_f L_h \beta^{-3} \cdot n^{\frac{3}{2}} \cdot \exp(3R^2) \|x - y\|_2
    \end{align*}
\end{proof}

\subsection{Step 4: \texorpdfstring{$G_4$}{} is lipschitz continuous}
\label{sub:hessian_is_lip_app:4}

In this section, we show that $G_4$ is Lipschitz continuous.

\begin{lemma}
    Let $G_4(x)$ be defined in Definition~\ref{def:G_i}, then, we have
    \begin{align*}
        |G_4(x) - G_4(y)| \leq  4 (R + R_h) R^4 R_f L_h \beta^{-1} \cdot \sqrt{n} \cdot \exp(R^2) \|x - y\|_2
    \end{align*}
\end{lemma}
\begin{proof}
    The proof is similar to Lemma~\ref{lem:lipschitz_of_G_6}.
\end{proof}

\subsection{Step 5: \texorpdfstring{$G_5$}{} is lipschitz continuous}
\label{sub:hessian_is_lip_app:5}

In this section, we show that $G_5$ is Lipschitz continuous.

\begin{lemma}
    Let $G_5(x)$ be defined in Definition~\ref{def:G_i}, then, we have
    \begin{align*}
        |G_5(x) - G_5(y)| \leq 10 (R + R_h) R^4 R_f L_h \beta^{-3} \cdot n^{\frac{3}{2}} \cdot \exp(3R^2) \|x - y\|_2
    \end{align*}
\end{lemma}
\begin{proof}
    The proof is similar to Lemma~\ref{lem:lipschitz_of_G_3}.
\end{proof}

\subsection{Step 6: \texorpdfstring{$G_6$}{} is lipschitz continuous}
\label{sub:hessian_is_lip_app:6}

In this section, we show that $G_6$ is Lipschitz continuous.

\begin{lemma}\label{lem:lipschitz_of_G_6}
    Let $G_6(x)$ be defined in Definition~\ref{def:G_i}, then, we have
    \begin{align*}
        |G_6(x) - G_6(y)| \leq 3 (R + R_h) R^4 R_f L_h \beta^{-1} \cdot \sqrt{n} \cdot \exp(R^2) \|x - y\|_2
    \end{align*}
\end{lemma}
\begin{proof}
    Note that
    \begin{align*}
        |G_6(x) - G_6(y)|
        = & ~ |\langle c(x) , Q_2(x) \cdot ( A_{1,*,i} \circ f(x) \circ A_{1,*,j}) - \langle c(y) , Q_2(y) \cdot ( A_{1,*,i} \circ f(y) \circ A_{1,*,j})|
    \end{align*}
    For simplicity, we define
    \begin{align*}
        G_{6, 1} : & = \langle c(x) , Q_2(x) \cdot ( A_{1,*,i} \circ f(x) \circ A_{1,*,j}) - \langle c(x) , Q_2(x) \cdot ( A_{1,*,i} \circ f(y) \circ A_{1,*,j}) \\
        G_{6, 2} : & = \langle c(x) , Q_2(x) \cdot ( A_{1,*,i} \circ f(y) \circ A_{1,*,j}) - \langle c(x) , Q_2(y) \cdot ( A_{1,*,i} \circ f(y) \circ A_{1,*,j}) \\
        G_{6, 3} : & = \langle c(x) , Q_2(y) \cdot ( A_{1,*,i} \circ f(y) \circ A_{1,*,j}) - \langle c(y) , Q_2(y) \cdot ( A_{1,*,i} \circ f(y) \circ A_{1,*,j})
    \end{align*}
    
    First, we upper bound $|G_{6, 1}|$:
    \begin{align*}
        |G_{6, 1}|
        = & ~ | \langle c(x) , Q_2(x) \cdot ( A_{1,*,i} \circ (f(x) - f(y)) \circ A_{1,*,j}) | \\
        \leq & ~ \|c(x)\|_2 \|Q_2(x)\| \|A_{1,*,i}\|_2^2 \|f(x) - f(y)\|_2 \\
        \leq & ~ (R + R_h) R_h R^3 R_f \|x - y\|_2
    \end{align*}
    where the first step follows from the definition of $G_{6, 1}$,
    the second step follows from Fact~\ref{fac:matrix_norm} and Fact~\ref{fac:vec_norm},
    the last step follows from {\bf Part 2} and {\bf Part 3} of Lemma~\ref{lem:function_norm_bounds}, {\bf Part 4} of Lemma~\ref{lem:norms} and $\|A_{1,*,i}\|_2 \leq R$.

    Next, we upper bound $|G_{6, 2}|$:
    \begin{align*}
        |G_{6, 2}|
        = & ~ |\langle c(x) , (Q_2(x) - Q_2(y)) \cdot ( A_{1,*,i} \circ f(y) \circ A_{1,*,j})| \\
        \leq & ~ \|c(x)\|_2 \|Q_2(x) - Q_2(y)\| \|A_{1,*,i}\|_2^2 \|f(x)\|_2 \\
        \leq & ~ (R + R_h) R^4 R_f L_h \beta^{-1} \cdot \sqrt{n} \cdot \exp(R^2) \|x - y\|_2
    \end{align*}
    where the first step follows from the definition of $G_{6, 1}$,
    the second step follows from Fact~\ref{fac:matrix_norm} and Fact~\ref{fac:vec_norm},
    the last step follows from {\bf Part 1} and {\bf Part 2} of Lemma~\ref{lem:function_norm_bounds}, {\bf Part 6} of Lemma~\ref{lem:norms} and $\|A_{1,*,i}\|_2 \leq R$.

    Then, we upper bound $|G_{6, 3}|$:
    \begin{align*}
        |G_{6, 3}|
        = & ~ |\langle c(x) - c(y) , Q_2(y) \cdot ( A_{1,*,i} \circ f(y) \circ A_{1,*,j})| \\
        \leq & ~ \|c(x) - c(y)\|_2 \|Q_2(x)\| \|A_{1,*,i}\|_2^2 \|f(x)\|_2 \\
        \leq & ~ L_h R^4 R_f R_h \beta^{-1} \cdot \sqrt{n} \cdot \exp(R^2) \|x - y\|_2
    \end{align*}

    Thus, we have
    \begin{align*}
        |G_6(x) - G_6(y)|
        = & ~ |G_{6, 1} + G_{6, 2} + G_{6, 3}| \\
        \leq & ~ 3 (R + R_h) R^4 R_f L_h \beta^{-1} \cdot \sqrt{n} \cdot \exp(R^2) \|x - y\|_2
    \end{align*}
\end{proof}

%\iffalse

\ifdefined\isarxiv
\bibliographystyle{alpha}
\bibliography{ref}
\else

\fi

%\fi

%%%% Cut-line between first 10 pages and appendix

%%% some writing rules

%% Writing rule for creating tags.
%% Tags :
%% Theorem    \ref{thm:bla_bla}
%% Lemma      \ref{lem:bla_bla}
%% Claim      \ref{cla:bla_bla}
%% Corollary  \ref{cor:bla_bla}
%% Fact       \ref{fac:bla_bla}
%% Definition \ref{def:bla_bla}
%% Section    \ref{sec:bla_bla}
%% Subsection \ref{sub:bla_bla}
%% Equation   \ref{eq:bla_bla}

\end{document}